\documentclass[runningheads]{llncs}
\usepackage{graphicx}

\usepackage{tikz}
\usepackage{comment}
\usepackage{amsmath,amssymb} 

\usepackage[accsupp]{axessibility}  

\usepackage{graphicx}
\usepackage{amsmath}
\usepackage{amssymb}
\usepackage{booktabs}
\usepackage{xfrac}
\usepackage{colortbl}
\usepackage{sidecap}
\usepackage{enumitem}
\usepackage{xcolor}
\usepackage{adjustbox}
\usepackage{multirow}
\usepackage{caption}
\usepackage{subcaption}
\usepackage{xspace}
\usepackage{wrapfig,lipsum,booktabs}
\usepackage{multicol}
\usepackage{hyperref}
\hypersetup{
    colorlinks=true,
    linkcolor=blue,
    filecolor=blue,      
    urlcolor=blue,
    pdftitle={Overleaf Example},
    pdfpagemode=FullScreen,
    }
\usepackage{cleveref}

\def\eg{\emph{e.g.,}\xspace} 
\def\ie{\emph{i.e.,}\xspace}

\def\etal{\emph{et al.}\xspace}

\usepackage{etoolbox}
\BeforeBeginEnvironment{wraptable}{\setlength{\intextsep}{0.1cm}}

\newcommand\blfootnote[1]{%
  \begingroup
  \renewcommand\thefootnote{}\footnote{#1}%
  \addtocounter{footnote}{-1}%
  \endgroup
}

\newcommand{\txt}[1]{{\texttt{#1}}}

\definecolor{Gray}{gray}{0.90}
\definecolor{LightCyan}{rgb}{0.82,0.82,1}
\definecolor{LightGray}{HTML}{E3E4E4}
\definecolor{Pearl}{HTML}{F1EAE3}
\definecolor{yellow-small}{HTML}{FCF3CF}  
\definecolor{purple-med}{HTML}{E8DAEF}
\definecolor{green-large}{HTML}{A9DFBF}
\definecolor{tableblue}{HTML}{98AFC7}

\begin{document}
\pagestyle{headings}
\mainmatter
\def\ECCVSubNumber{6491}  

\title{Class-agnostic Object Detection with Multi-modal Transformer} 

\titlerunning{Class-agnostic Object Detection with Multi-modal Transformer}
%
\author{
Muhammad Maaz\inst{1}$^{*}$ \and
Hanoona Rasheed\inst{1}$^{*}$ \and
Salman Khan\inst{1,2} \and
Fahad Shahbaz Khan\inst{1,3} \and
Rao Muhammad Anwer\inst{1,4} \and
Ming-Hsuan Yang\inst{5,6,7}
}
\institute{$^\text{1} $Mohamed bin Zayed University of AI \qquad $^\text{2} $Australian National University \qquad \\ $^\text{3} $ Linköping University \qquad $^\text{4} $  Aalto University \qquad $^\text{5} $ University of California, Merced \qquad $^\text{6} $ Yonsei University \qquad $^\text{7} $  Google Research}
\authorrunning{Maaz et al.}
\maketitle

\begin{abstract}
\blfootnote{\textsuperscript{*}Equal contribution}
What constitutes an object? This has been a long-standing question in computer vision. Towards this goal, numerous learning-free and learning-based approaches have been developed to score \emph{objectness}. However, they generally do not scale well across new domains and novel objects. In this paper, we advocate that existing methods lack a top-down supervision signal governed by human-understandable semantics. For the first time in literature, we demonstrate that Multi-modal Vision Transformers (MViT) trained with aligned image-text pairs can effectively bridge this gap. Our extensive experiments across various domains and novel objects show the state-of-the-art performance of MViTs to localize generic objects in images. Based on the observation that existing MViTs do not include multi-scale feature processing and usually require longer training schedules, we develop an efficient MViT architecture using multi-scale deformable attention and late vision-language fusion. We show the significance of MViT proposals in a diverse range of applications including open-world object detection, salient and camouflage object detection, supervised and self-supervised detection tasks. Further, MViTs can adaptively generate proposals given a  specific language query and thus offer enhanced interactability. Code: \url{https://git.io/J1HPY}.

\end{abstract}

\begin{figure}[ht]
\centering
\includegraphics[width=0.98\linewidth]{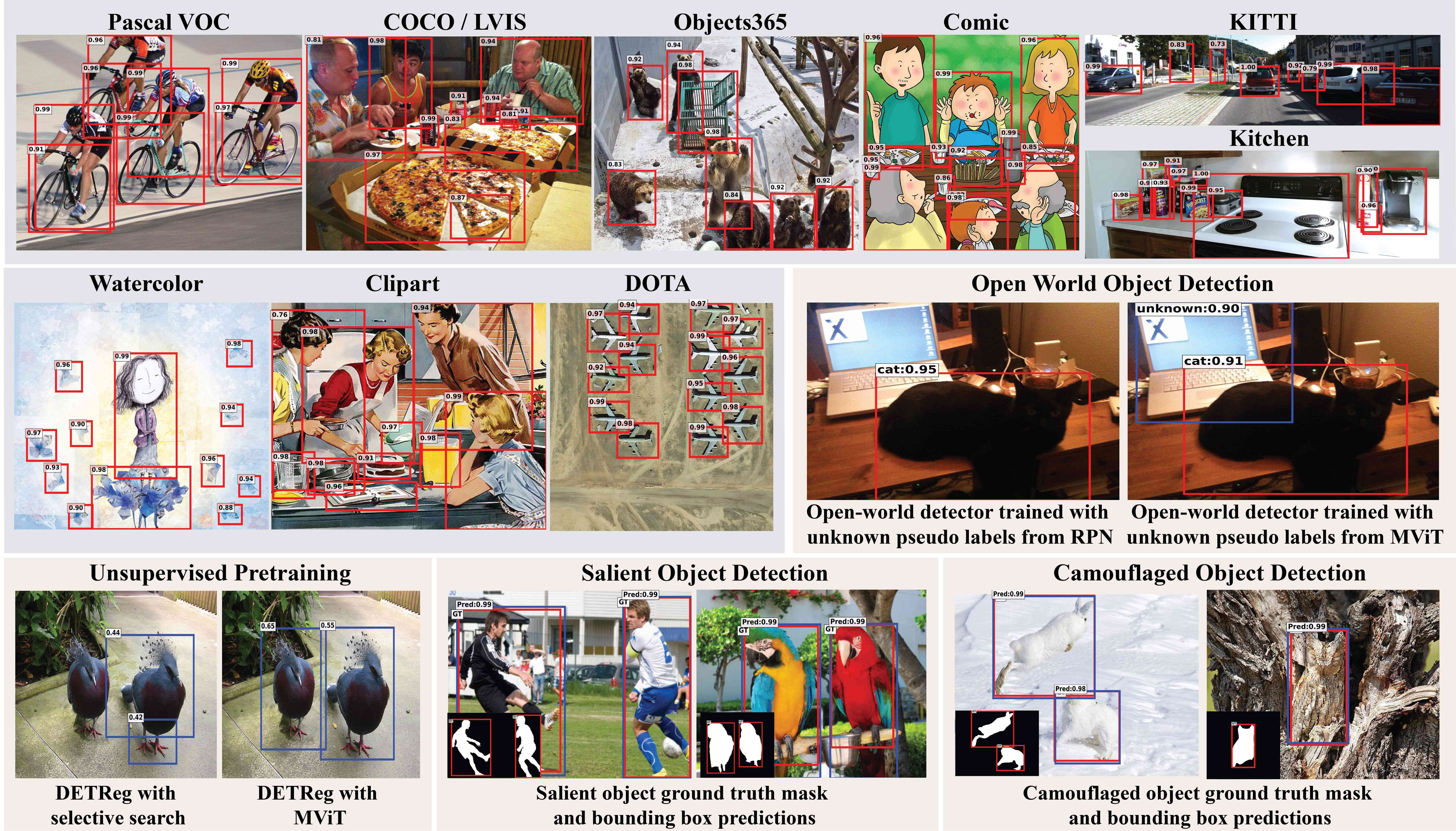}
\caption{\small We show that Multi-modal Vision Transformers (MViTs) excel at Class-agnostic
OD across multiple domains: natural images \cite{voc,coco,kitti,kitchen}, satellite images \cite{dota}, sketches, cartoons  and paintings \cite{clipart-comic-water}
(\colorbox{LightGray}{gray background}).
The MViTs perform well on diverse datasets (with many classes \eg\ LVIS, Object365) using intuitive natural language text queries
(\eg\ \txt{all objects}).
Further, class-agnostic detectors (MViTs) can be applied to several downstream applications
(\colorbox{Pearl}{pearl background}).
In Open-world OD \cite{joseph2021towards}, unknown pseudo-labels generated using MDETR \cite{mdetr} can improve novelty detection. For unsupervised object localization, replacing Selective Search proposals \cite{uijlings2013selective} in DETReg \cite{detreg} pretraining with only top-30 MViT proposals leads to improved localization. For Salient and Camouflaged OD, task specific text queries
can help perform competitively against fully supervised models without any task specific tuning. Overall, MViTs achieve the state-of-the-art results on various downstream applications.}
\label{fig:fig_1}
\end{figure}

\section{Introduction}

\label{sec:intro}

The recent years have witnessed significant advances in object detection (OD) \cite{liu2020deep} based on developments of large-scale annotated datasets and carefully designed deep learning models. Notably, efforts have been made to tackle more difficult cases such as universal OD \cite{universal}, long-tailed object distribution modeling \cite{gupta2019lvis}, open-vocabulary \cite{zareian2021open} and open-world OD \cite{joseph2021towards}. In contrast, little progress has been made towards a seemingly simpler task of class-agnostic OD \cite{alexe2010object} in recent years. In the era of fully trainable pipelines, class-agnostic OD is still often approached using typical bottom-up approaches such as Selective Search \cite{uijlings2013selective}, EdgeBox \cite{zitnick2014edge}, DeepMask \cite{pinheiro2015learning} and MCG \cite{pont2016multiscale}.

Despite being an apparently simpler problem in terms of the two-way classification space, the class-agnostic OD task is indeed challenging from the representation learning perspective. The main challenge is to model the vast diversity of \emph{all} valid object classes and delineate such a diverse group from the \emph{background} class which itself has vague semantic definition \cite{alexe2012measuring}. Our experiments indicate that this intrinsic complexity of the task makes it difficult to design fully trainable class-agnostic OD models that can work across domains and for novel unseen objects. Although the bottom-up approaches offer proposals for generic objects, they come at the cost of a prohibitively large number of candidate boxes, low-precision, lack of semantic understanding and slow processing, making them less scalable to generic operation in the wild. More recently, self-supervised learning frameworks -- based on both ViTs \cite{detreg,dai2021up} and CNNs \cite{xie2021detco,xiao2021region} -- have 
focused on promoting better localization of generic objects, however they still show modest performance on class-agnostic OD \cite{detreg}. Our intuition is that \emph{top-down supervisory signals} are necessary to resolve the ambiguous nature of class-agnostic OD task, which is precisely what is missing from the aforementioned approaches. 

In this paper, we bring out the capacity of recent Multi-modal Vision Transformers (MViTs) to propose generic class-agnostic OD across different domains. The high-level information provided by the language descriptions helps learn fairly generalizable properties of universal object categories. In turn, the MViTs perform exceptionally well compared to uni-modal object detectors trained for generic object detection as well as the typical bottom-up object proposal generation schemes. Due to the multi-modal nature of these models, we design language-driven queries to discover valid objects in a human-understandable format that can be adapted to explore varied aspects of the object semantic space. With the state-of-the-art performance, an ensuing question is to explore the root cause of such generalization for the `\emph{concept of objects}' embedded in MViTs. Through a series of systematic experiments, we find that it is the language skeleton/structure (rather than the lexicon itself) that defines this strong understanding of generic object definition within MViT models. As an interesting example, when the MViT is trained without actual captions, but just the bounding boxes corresponding to a natural language description, the model still demonstrates strong class-agnostic OD generalization. These insights on the interactive class-agnostic OD mechanism can be deployed in several \emph{downstream} tasks such as novel object discovery, saliency detection, self-supervised learning and open-world detection. The main highlights of this work include:

\begin{itemize}
    \item We demonstrate the state-of-the-art performance of pre-trained MViTs \cite{mdetr,gpv1} towards class-agnostic OD via a set of human-understandable natural language queries. We also develop an efficient and flexible MViT model, \emph{Multiscale Attention ViT with Late fusion} (MAVL), which performs better in locating generic objects as compared to existing MViTs (Secs.~\ref{sec:mdef} and \ref{sec:multi-modal}).
    \item We benchmark
    generalization of MViT based 
    OD models on diverse domains \eg natural images, sketches, cartoons, satellite images, paintings and show their favorable performance compared to existing class-agnostic OD models (bottom-up approaches, CNN and ViT based uni-modal 
    pipelines) (Sec.~\ref{sec:multi-modal}).
    \item  Our class-agnostic detectors can benefit various down-stream applications: Open-world OD, Salient OD, Camouflaged OD and Self-supervised learning. Furthermore, when these proposals are combined with RPN proposals in two-stage detectors, it can lead to overall performance improvements due to their rich top-down semantic understanding of the image content (Sec.~\ref{sec:app}).
    \item Through an extensive set of systematic experiments, we analyze the factors that majorly contribute to the improved performance of MViTs (Sec.~\ref{sec:what}).
\end{itemize}

\section{Multi-modal ViTs}\label{sec:mdef}

\begin{figure*}[!t]
\centering
{\includegraphics[height=3.3cm, width=1.0\textwidth]{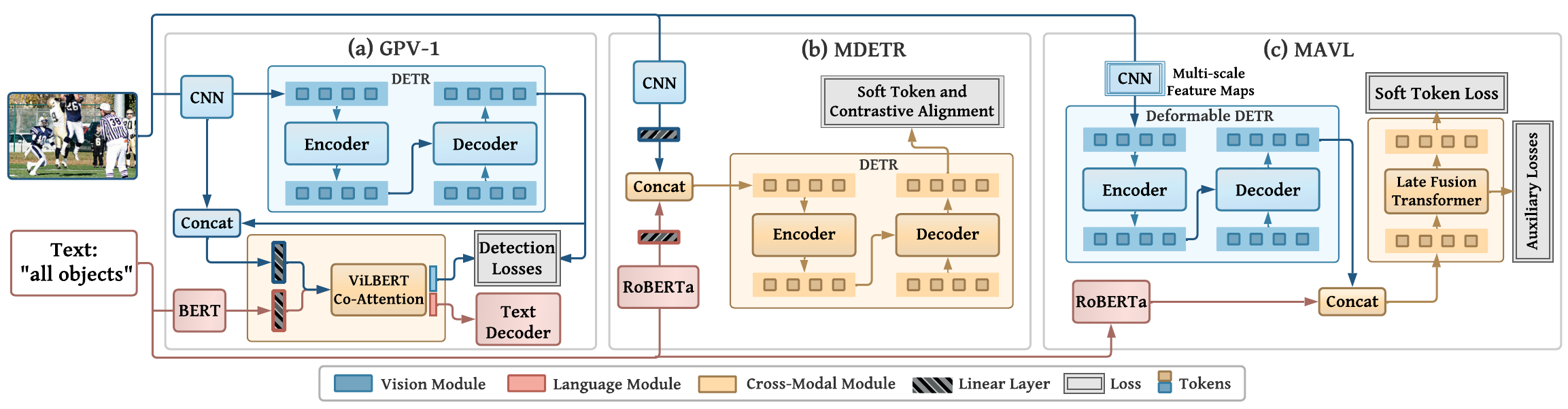}}
\caption{\small Architecture overview of MViTs used in this work -- GPV-1 \cite{gpv1}, MDETR \cite{mdetr} and MAVL (ours). GPV-1 takes image along with a task description as input and outputs relevant region boxes and text. MDETR uses soft token prediction and contrastive alignment in latent space for cross-conceptualization using aligned image-text pairs. MAVL utilizes multi-scale image features with multi-scale deformable attention module (MSDA), and uses late-fusion strategy for vision-language fusion.}
\label{mvit_block_diagram}
\end{figure*}

In this work, we bring out the generalization capacity of Multi-modal ViTs (MViT) to tackle generic OD. The capability of relating natural language with visual features helps MViTs to generalize to novel concepts, achieving state-of-the-art results on class-agnostic OD using human-understandable text queries (e.g., ‘\txt{all objects/entities}’). Before a detailed analysis, we provide background on MViTs and propose Multiscale Attention ViT with Late fusion (MAVL).

\noindent \textbf{\txt{(a)} GPV:}  Gupta \etal proposed GPV-I \cite{gpv1}, a unified architecture for multi-task learning, where the task is inferred from the text prompt. It takes an image and a task description as input and outputs text with the corresponding bounding boxes. This model uses pretrained BERT \cite{BERT} to encode the text, concatenates it with the region descriptors from DETR \cite{DETR} and passes it to ViLBERT \cite{ViLBERT} co-attention layers for cross-modal conceptualization. It predicts relevance scores for each predicted bounding box indicating the importance of the region for the prompted task. An output text decoder conditioned on the relevance scores is used for better cross-modal understanding (Fig.~\ref{mvit_block_diagram}~(a)). 
GPV is trained on data from five different vision-language tasks.

\noindent \textbf{\txt{(b)} MDETR:} Kamath~\etal~\cite{mdetr} proposed a modulated transformer  trained to detect objects in an image conditioned on a text query. In MDETR, visual and text features are extracted from a convolutional backbone (\eg ResNet-101 \cite{he2016deep} or EfficientNet \cite{tan2019efficientnet}) and a language model (RoBERTa \cite{roberta}) respectively. These features are then concatenated and passed to the DETR~\cite{DETR} model for detection (Fig.~\ref{mvit_block_diagram}~(b)). MDETR uses soft token prediction and contrastive alignment in latent space for addressing text-conditioned object detection. In soft token prediction, a uniform probability distribution is predicted over all text tokens for each detected object. In contrastive alignment, the embedded object queries from decoder are aligned with the text representation from encoder. This multi-modal alignment makes the object embeddings closer to the corresponding text embeddings in feature space. The model is pre-trained with 1.3M image-text pairs and achieves the state-of-the-art results on various vision-language downstream tasks including VQA, referring expression and phrase grounding.

\noindent \textbf{\txt{(c)} MAVL:}  We develop a new multimodal architecture called Multi-scale Attention ViT with Late fusion (MAVL) that improves the class-agnostic OD performance of MDETR {using multi-scale spatial context and deformable attention} making it efficient to train. Fig.~\ref{mvit_block_diagram}~(c) shows our overall design. Below, we highlight the main features of MAVL: \\
\noindent
$-$\textit{Multi-scale Deformable Attention (MSDA).} MDETR \cite{mdetr} finds it challenging to scale to high-resolution feature maps due to a fixed self-attention design. 
Further, it operates on a specified spatial scale which can be sub-optimal for small objects. Our design calculates attention at multiple scales to incorporate better contextual information. However, multiple scales can increase the computational cost, therefore we use Deformable Attention proposed in \cite{zhu2020deformable} that employs multi-scale feature processing and dynamically attends to relevant pixel locations for context aggregation. Specifically, it samples a small set of keys around a reference (query) image location. The sparse key sampling in MSDA achieves linear complexity with respect to the size of the image feature maps.\\
\noindent
$-$\textit{Late Multi-modal Fusion.} MSDA module
utilizes the spatial structure of an image to sparsely sample keys for each query point. Following the MDETR strategy of concatenating text embeddings with flattened 
features would destroy the spatial structure of an image. Hence, we fuse text in MAVL model after the images are processed through the Def-DETR encoder-decoder architecture using a \emph{late fusion} mechanism. Specifically, the object query representations from the deformable decoder are concatenated with the text embeddings, and passed through a series of {six} transformer self-attention (SA) blocks. This design choice is inspired by the recent vision-language fusion works \cite{ViLBERT,VL-BERT,LXMERT,sun2019videobert}. Using the training procedure of \cite{DETR}, the output head is applied after each SA block and the total loss is calculated by adding all auxiliary losses. We note that no explicit contrastive alignment of object query representation and encoded text is required in our approach. Our experiments show fast convergence (only \emph{half} iterations) and competitive performance of MAVL against MDETR (Tables \ref{table1:class_agnostic}, \ref{table2:class_agnostic_generalization}).

\noindent
$-$\textit{Implementation Details.}
Similar to MDETR \cite{mdetr}, we train MAVL on approx. 1.3M aligned image-text pairs, using
images from Flickr30k \cite{plummer2015flickr30k}, MS-COCO (2014) \cite{coco} and Visual Genome (VG) \cite{krishna2017visual}. The corresponding annotations are taken from Flickr entities, RefCOCO/+/g referring expression 
\cite{kazemzadeh2014referitgame}, VG regions and GQA \cite{hudson2019gqa}. In the onward discussion, we refer to this dataset as \emph{Large-scale Modulated Detection} ({LMDet})dataset.
All MDETR and MAVL models are trained with ImageNet-1K \cite{russakovsky2015imagenet} pretrained ResNet-101 \cite{he2016deep}.
Our MAVL \emph{converges in 20 epochs} (MDETR requires 40 epochs) on LMDet using the same hyper-parameters as in MDETR. See Appendix~\ref{app:imp_detail_mdef} for more details.

\section{Multi-modal ViTs as Generic Detectors}\label{sec:multi-modal}
The class-agnostic OD seeks to differentiate between generic objects and background in images. This task involves learning the notion of \emph{objectness}. Existing approaches typically explore low-level visual cues
(i.e. superpixels, edges, etc.) or directly learn the mapping between images and generic object locations using fully trainable pipelines learned with bounding box annotations \cite{uijlings2013selective,zitnick2014edge,jaiswal2021class,detreg}. We note that these procedures lack high-level semantic information necessary to relate objects across diverse scenes to derive a comprehensive and general notion of universal objects. In this work, 
We explore the class-agnostic OD capacity of MViTs trained using aligned image-text pairs (Sec.~\ref{sec:mdef}).  We observe these models can produce high quality object proposals by using intuitive text queries like ‘\txt{all objects}’ and ‘\txt{all entities}’. This demonstrates  their capability to relate natural language with visual concepts to model generic objectness, enabling them to discover novel categories and generalize across different domains while offering human interaction with intelligible text queries.

\begin{table}[!t]
\caption{\small Class-agnostic OD results of \colorbox{orange!6}{MViTs} in comparison with bottom-up approaches (row 3-5) and uni-modal detectors (row 6-8) trained to localize generic objects. {Bottom row shows gain of MAVL over the best uni-modal method.}
In general, MViTs achieve state-of-the-art performance 
using intuitive text queries (details in Sec.~\ref{subsec:language_queries}).}
\begin{center}
\setlength{\tabcolsep}{6pt}
\resizebox{0.95\linewidth}{!}{
\begin{tabular}{l *{10}{c}}
  \toprule
  \rowcolor{Gray}
  Dataset $\rightarrow$ & \multicolumn{2}{c}{Pascal-VOC} & \multicolumn{2}{c}{COCO} & \multicolumn{2}{c}{KITTI} & \multicolumn{2}{c}{Objects365} & \multicolumn{2}{c}{LVIS} \\
  \rowcolor{Gray} Model $\downarrow$ & AP50 & R50 & AP50 & R50 & AP50 & R50 & AP50 & R50 & AP50 & R50\\
  \midrule
  \midrule
  Edge Boxes & 0.08	& 7.14 & 0.09 & 5.16 & 0.09	& 6.58 &  0.07 &   3.27 &   0.05 &  3.00 \\
  Selective Search & 0.32 & 21.4 & 0.27 & 12.7 & 0.03 & 4.85 &   0.38 &  10.7 &  0.24 &  9.31 \\
  Deep Mask	& 5.92 & 40.4 & 2.16 & 19.2 & 1.33 & 15.5 & 1.31 & 14.5 & 0.51 & 8.17 \\
  \midrule
  Faster-RCNN & 42.9 & 85.8 & 26.4 & 58.7 & 23.5 & 53.2 & 24.8 & 54.6 & 8.91 & 35.6 \\
  RetinaNet	& 43.2	& 86.6	& 24.6	& 59.1	& 30.4	& 57.6 & 24.3 & 54.8 & 8.57 & 35.7 \\
  Def-DETR & 30.1 & 81.0 & 20.0 & 53.5 & 23.7 & 55.0 & 17.0 & 45.9 & 6.60 & 30.7 \\
  \midrule
  \rowcolor{orange!6}
  GPV-I	& 61.9 & 91.1 & 38.0 & 64.4 & 43.0 & 64.4 & 25.6 & 50.2 & 9.18 & 27.5 \\
  \rowcolor{orange!6}
  MDETR	& 66.0	& 90.1	& 40.7	& 62.2	& 46.7	& \textbf{67.2} & 30.4 & 54.0 & 10.7 & 32.8 \\
  \rowcolor{orange!6}
  MAVL (Ours)	& \textbf{68.6} & \textbf{91.3} & \textbf{43.6} & \textbf{65.0} & \textbf{48.2} & 63.5 & \textbf{33.2} & \textbf{57.9} & \textbf{11.7} & \textbf{37.0} \\ 
  \midrule
   	& \color{blue} +25.4 & \color{blue} +4.7 & \color{blue} +19.0 & \color{blue} +5.9 & \color{blue} +17.8 & \color{blue} +5.9 & \color{blue} +8.4 & \color{blue} +3.1  & \color{blue} +2.8 & \color{blue} +1.3 \\
  \bottomrule
\end{tabular}}
\end{center}
\label{table1:class_agnostic}
\end{table}

\subsection{Class-agnostic Object Detection}\label{sec:class-agnostic}
\textbf{Settings:}
Table \ref{table1:class_agnostic} shows the object proposal generation performance of MViTs with the typical bottom-up approaches and the end-to-end supervised deep learning methods on five  challenging natural image OD datasets (Pascal VOC \cite{voc}, MS COCO \cite{coco}, KITTI \cite{kitti}, {Objects365 \cite{shao2019objects365} and LVIS \cite{gupta2019lvis}}). The bottom-up approaches considered for comparison include EdgeBoxes \cite{zitnick2014edge}, Selective Search \cite{uijlings2013selective} and DeepMask \cite{pinheiro2015learning} while Faster-RCNN \cite{ren2015faster}, RetinaNet \cite{lin2017focal} and Deformable-DETR \cite{zhu2020deformable} are selected from the deep-learning based methods due to the state-of-the-art performance in class-aware OD. The MViTs considered are GPV-I \cite{gpv1} and MDETR \cite{mdetr} alongside our proposed MAVL (see Sec.~\ref{sec:mdef} for details). 

For fairness, all the uni-modal detectors considered for evaluation are trained with ResNet-101 backbone using box-level supervision on {LMDet} dataset. Faster-RCNN and RetinaNet follow the standard Detectron2 \cite{wu2019detectron2} training setting with FPN at 1$\times$ schedule. {The combined detections from the text queries in Table~\ref{table6:combined} are used for evaluating MViTs (see Sec.~\ref{subsec:language_queries} and Appendix~\ref{appendix:class_agnostic_evaluation} for details).} Moreover, images used in the evaluation \emph{do not} have any overlap with LMDet.

\noindent\textbf{Results:} We report both average precision (AP) and Recall at IoU threshold of 0.5 using the top-50 boxes from each method. Overall, the detectors trained in class-agnostic fashion perform reasonably well on all datasets, surpassing the bottom-up methods by a large margin. Furthermore, the MViTs perform better than the uni-modal approaches with the use of simple human understandable natural language text queries. This performance shows MViTs' strong understanding of language content obtained from the pretrained NLP model (BERT \cite{BERT}, RoBERTa \cite{roberta}) along with the aligned image-text pairs used in pretraining.

For MViTs, interestingly a relatively small number of boxes match the quality achieved by a much larger proposal set
from competing methods. Fig.~\ref{graph_1} shows the recall
obtained by varying the number of top object proposals for all methods on two datasets. MViTs achieve competitive recall
with only top-10 proposals.

\begin{table}[!t]
\caption{\small Class-agnostic OD performance of \colorbox{orange!6}{MViTs} in comparison with RetinaNet \cite{lin2017focal} on several out-of-domain datasets. MViTs show consistently good results on all datasets.  $^{\dagger}$Proposals on DOTA \cite{dota} are generated by multi-scale inference (see Sec.~\ref{appendix:class_agnostic_evaluation}).}
\begin{center}
\setlength{\tabcolsep}{6pt}
\resizebox{0.95\linewidth}{!}{
\begin{tabular}{l cc cc cc cc cc}
  \toprule
  \rowcolor{Gray}
  Dataset $\rightarrow$ & \multicolumn{2}{c}{Kitchen} & \multicolumn{2}{c}{Clipart} & \multicolumn{2}{c}{Comic} & \multicolumn{2}{c}{Watercolor} & \multicolumn{2}{c}{DOTA$^{\dagger}$}
  \\ 
  \rowcolor{Gray} Model $\downarrow$ & AP50 & R50 & AP50 & R50 & AP50 & R50 & AP50 & R50 & AP50 & R50 \\
  \midrule
  \midrule
  RetinaNet & 35.3	& 89.5	& 27.0	& 90.0	& 33.1	& 86.1	& 47.8	& 91.9	& 0.72 & 15.6 \\
  \rowcolor{orange!6}
  GPV-1 & 24.5	& 84.8	& 35.1	& 86.1	& 42.3	& 83.6	& 50.3	& 89.5	& 0.55 & 9.33 \\
  \rowcolor{orange!6}
  MDETR & 38.4	& \textbf{91.4}	& 44.9	& 90.7	& 55.8	& \textbf{89.5}	& 63.6	& 94.3	& 1.94 & 21.8 \\
  \rowcolor{orange!6}
  MAVL (Ours) & \textbf{45.4} & 91.0 & \textbf{50.6} & \textbf{92.9} & \textbf{57.7} & 89.2 & \textbf{63.8} & \textbf{95.6} & \textbf{2.86} & \textbf{24.2} \\ 
  \bottomrule                             
\end{tabular}}
\end{center}
\label{table2:class_agnostic_generalization}
\end{table}

\subsection{How well MViTs generalize?}

\noindent \textbf{Generalization to New Domains:}
We extend our analysis from natural image datasets (Sec.~\ref{sec:class-agnostic}) to rule out if MViT representations are biased towards natural images, for which these models are originally trained on. To this end, we evaluate on universal OD datasets \cite{universal} belonging to five different domains (Table \ref{table2:class_agnostic_generalization}). The studied domains include indoor kitchen scenes \cite{kitchen}, cartoon images, watercolor drawings, clipart, comics \cite{clipart-comic-water} and satellite/aerial images (DOTA dataset) \cite{dota}. The experiments follow the same setting as in Sec.~\ref{sec:class-agnostic}. These results indicate the generalization capability of MViTs in comparison to the best proposal generation methods earlier evaluated in Table~\ref{table1:class_agnostic} (RetinaNet trained for class-agnostic OD).

\begin{figure}[!t]
  \centering
  \begin{subfigure}{.51\textwidth}
    \centering
    \includegraphics[height=3.0cm, width=1.0\linewidth]{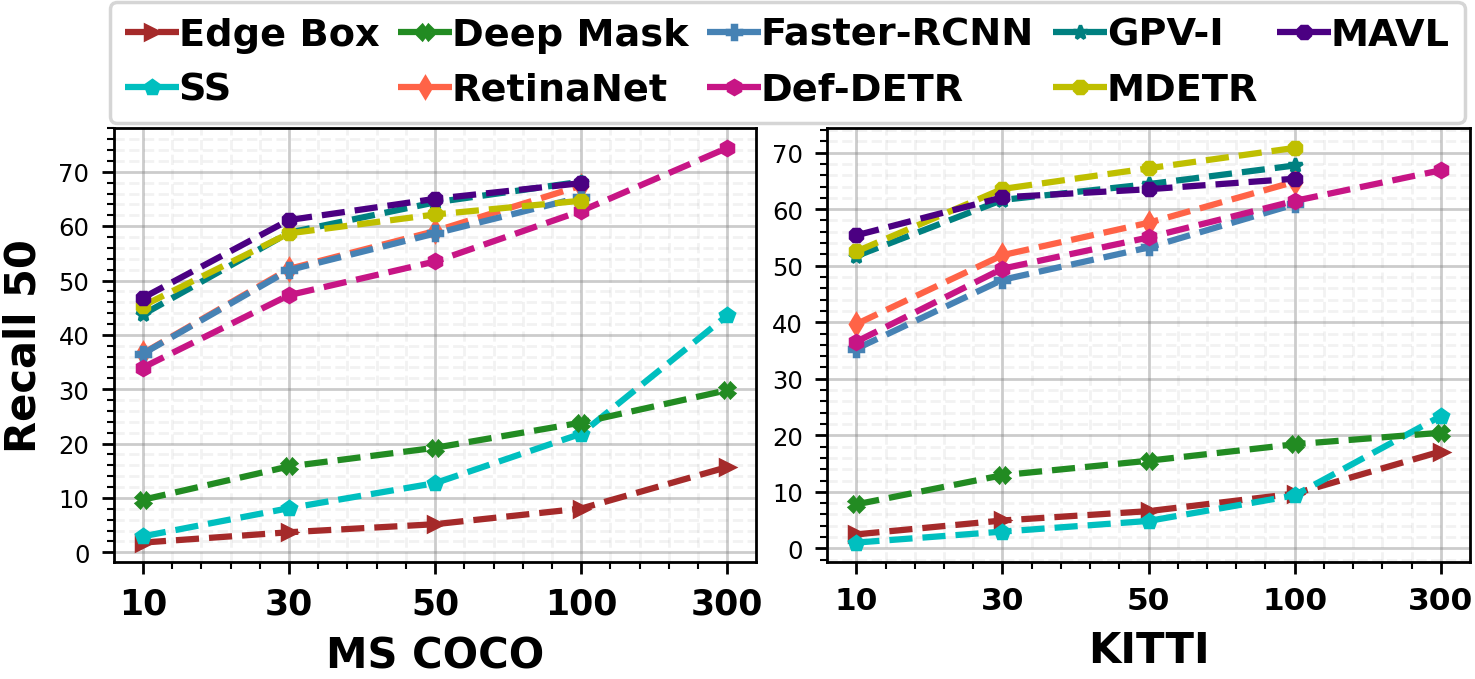}
    \caption{\small}
    \label{graph_1}
  \end{subfigure}%
  \hfill
  \begin{subfigure}{.48\textwidth}
    \centering
    \includegraphics[height=3.0cm, width=1.0\linewidth]{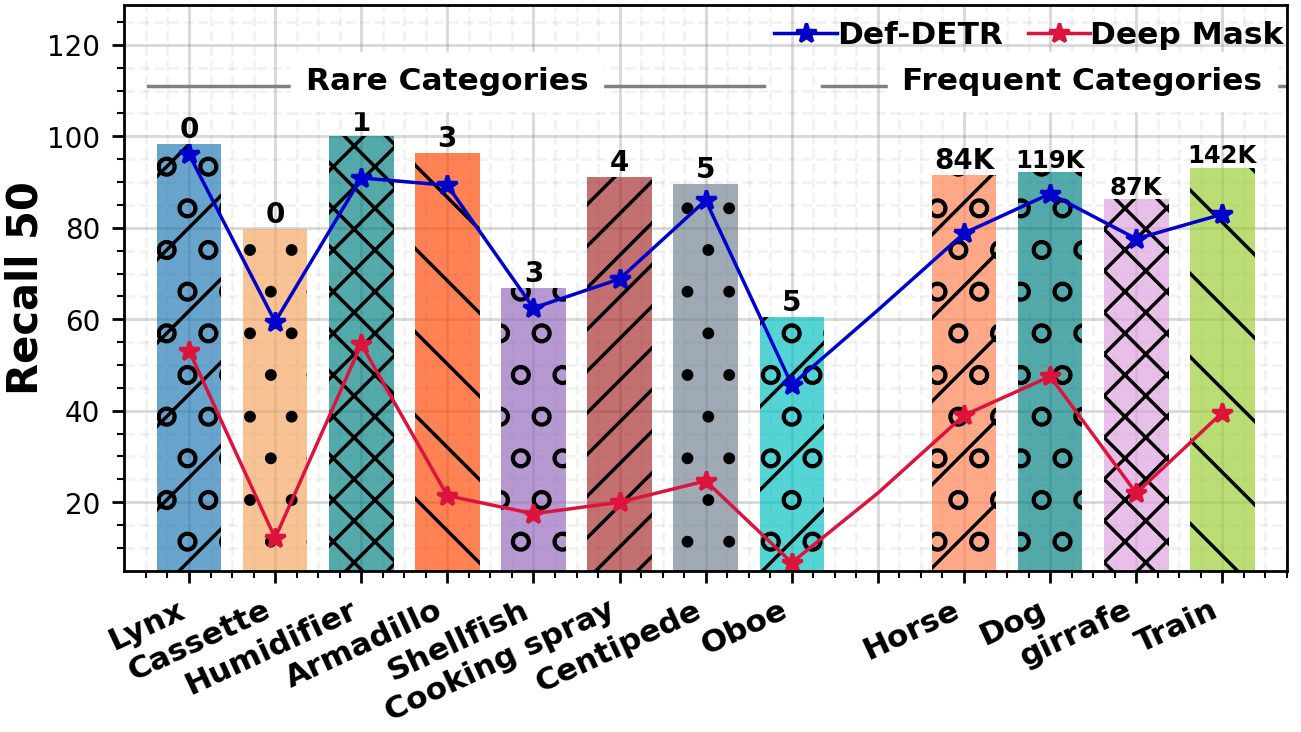}
    \caption{}
    \label{fig:unseen_classes}
    
  \end{subfigure}%
  \caption{\small \textbf{\color{blue}(a)} Effect of using different number of top-ranked boxes on multiple class-agnostic OD methods. The MViTs exhibits good recall even with only top-10 proposals. \textbf{\color{blue}(b)} MAVL class-agnostic OD performance on rarely and frequently occurring categories in LMDet. Rare categories are selected from Open Images \cite{OpenImages}. The MAVL recall rates (represented by the bars) are compared with those of Def-DETR \cite{zhu2020deformable} and DeepMask \cite{pinheiro2015learning} (represented by the lines). The numbers on top of the bars indicate the total occurrences of the category in LMDet captions. The MViT achieves good recall even for the classes with no or very few occurrences in the training dataset.}
\label{fig:1-3}
\end{figure}

\noindent \textbf{Generalization to Rare/Novel Classes:}
With the notion of objectness, humans are capable of identifying novel and rare objects, although they may not recognize their specific category. Similarly, scalabiltiy to rare and novel classes is a desired quality of an object detector. 
To analyze this, the class-agnostic OD mechanism of MAVL is evaluated on rare categories from Open-Images \cite{OpenImages} versus frequent categories and compared with Deformable DETR and Deep Mask trained for class agnostic OD. 
Fig. \ref{fig:unseen_classes} indicate 
state-of-the-art recall 
on rare categories such as \emph{lynx}, \emph{humidifier}, and \emph{armadillo} with as few as zero training instance. Overall, we note the model generalizes well to rare/unseen categories.

\section{Applications and Use-cases}\label{sec:app}
The high-quality class-agnostic object proposals obtained from MViTs can be helpful towards several downstream applications, as we demonstrate next.

\subsection{Enhanced Interactability} \label{subsec:language_queries}
We have observed that MViTs can generate high quality object proposals with intuitive human understandable  queries such as ‘\txt{all objects}’. This motivates us to explore the language semantic space of such models to construct a set of queries
\begin{wraptable}[11]{r}{7cm}
\caption{\small Using different intuitive text queries with MAVL. Combining detections from multiple queries captures varying aspects of objectness.}
\setlength{\tabcolsep}{3pt}
\resizebox{1.0\linewidth}{!}{
\begin{tabular}{l *{6}{c}}
  \toprule
  \rowcolor{Gray} Dataset $\rightarrow$ & \multicolumn{2}{c}{Pascal-VOC} & \multicolumn{2}{c}{COCO} & \multicolumn{2}{c}{KITTI} \\
  \rowcolor{Gray} Text Query $\downarrow$ & AP50 & R50 & AP50 & R50 & AP50 & R50 \\
  \midrule
  \midrule
  \txt{all objects} & 51.3 & 85.5 & 33.3 & 58.4 & 40.2 & 64.0 \\
  \txt{all entities}  & 65.2 & 88.4 & 34.6 & 54.6 & 41.9 & 59.5 \\
  \txt{all visible entities \& objects} & 63.3 & 89.0 & 37.9 & 61.6 & 42.0 & 63.0 \\
  \txt{all obscure entities \& objects} & 59.5 & 86.6 & 35.2 & 59.1 & 42.4 & 63.5 \\
  \txt{all small objects} & 40.0 & 83.9 & 28.9 & 58.9 & 40.4 & 65.2  \\
  \midrule
  \rowcolor{orange!6}
  combined detections (CD) & 63.7 & 91.0 & 42.0 & \textbf{65.0} & \textbf{48.2} & \textbf{63.5} \\
  \rowcolor{orange!6}
  CD w/o ‘\text{all small objects}' & \textbf{68.6} & \textbf{91.3} & \textbf{43.6} & \textbf{65.0} & 45.8 & 61.6 \\
  \bottomrule                             
\end{tabular}
}
\label{table6:combined}
\end{wraptable}
that can well capture the generic concept of objectness. We filter words from captions in LMDet that are semantically close to the word ‘\txt{object}’ in the linguistic feature space. We then utilize these words to construct intuitive text queries such as ‘\txt{all objects}’, ‘\txt{all entities}’, ‘\txt{all visible entities and objects}’, and ‘\txt{all obscure entities and objects}’, for exploiting the class-agnostic OD performance of MViTs. The detections from the individual text queries are combined, filtered with class-agnostic non-maximum suppression (NMS) to remove duplicate detections, and top-N boxes are selected for evaluation. We use N=$50$ in all of our experiments.
\begin{figure}[t]
  \centering
  \begin{subfigure}{.495\textwidth}
    \centering
    \includegraphics[height=3.4cm, width=1.0\linewidth]{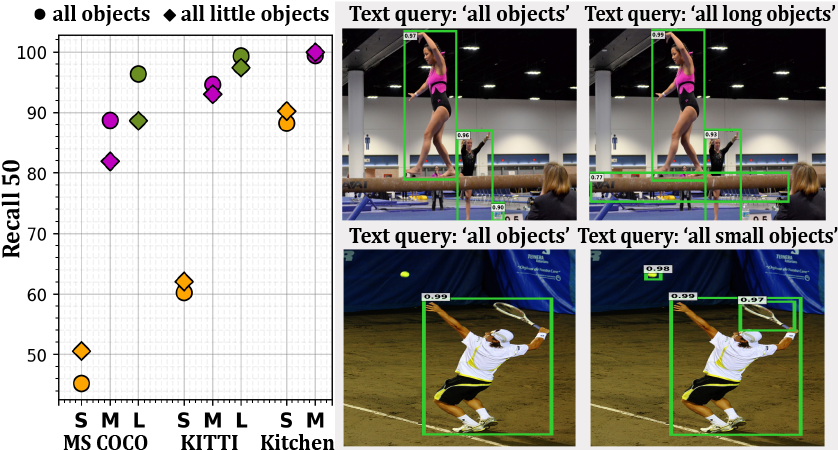}
    \caption{}
    \label{fig:graph_2}
  \end{subfigure}%
  \hfill
  \begin{subfigure}{.495\textwidth}
    \centering
    \includegraphics[height=3.4cm, width=1.0\linewidth]{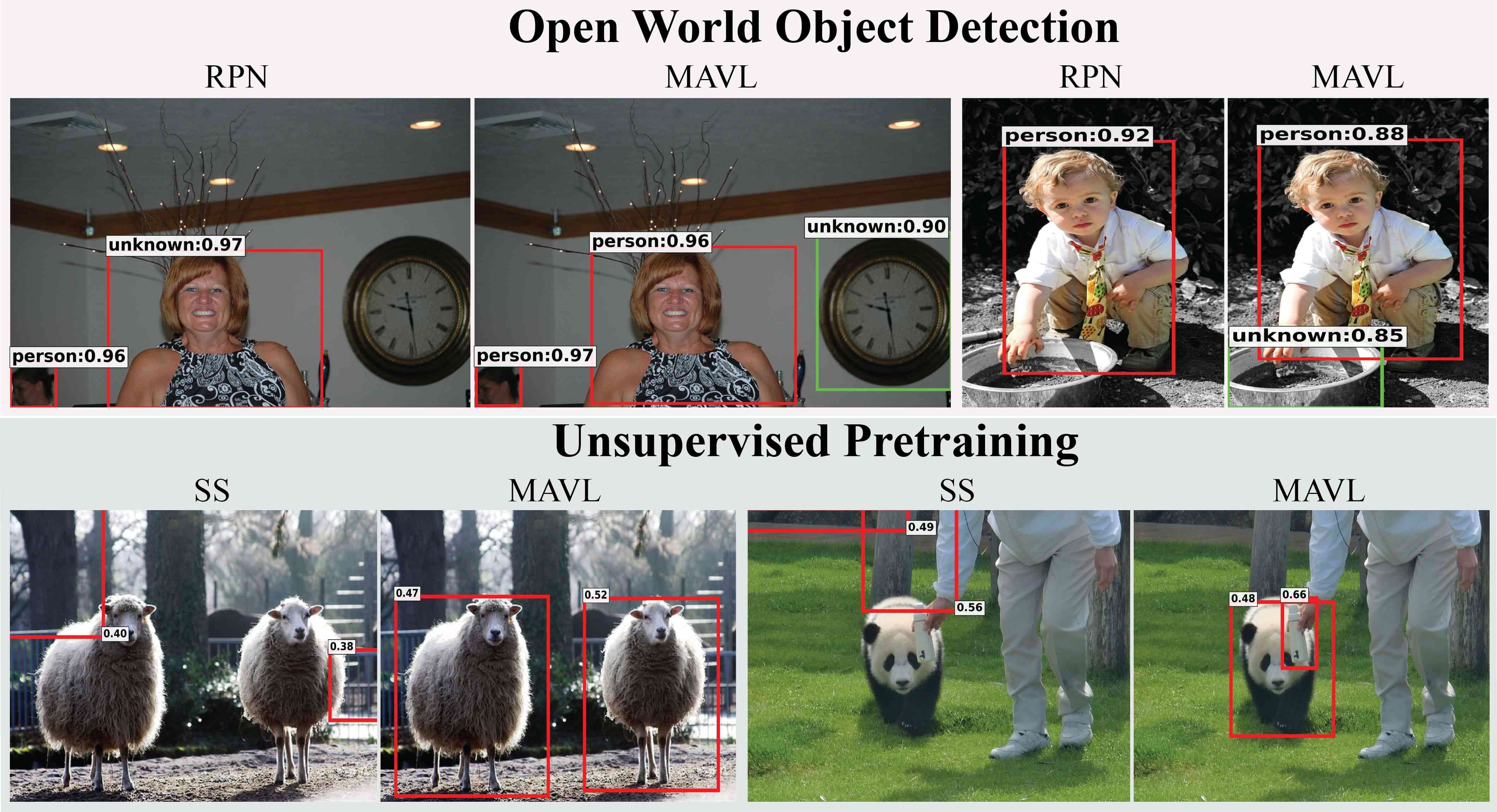}
    \caption{}
    \label{fig:ore_detreg}
  \end{subfigure}%
 \caption{\textbf{\color{blue}(a)} \small MAVL recall for small (S), medium (M) and large (L) objects across three datasets. The use of specific query (‘all little objects’) increases recall of small objects across different datasets (\emph{left}). Targeted detections by the relevant text queries (\emph{right}). \textbf{\color{blue}(b)} \small Visualizations of ORE \cite{joseph2021towards} unknown detections when trained with RPN versus MAVL  unknown pseudo-labels (\emph{top}). Class-agnostic OD of DETReg \cite{detreg} when trained using Selective Search (SS) \cite{uijlings2013selective} 
 versus MAVL proposals (\emph{bottom}).}
\label{fig:4-5_comb}
\end{figure}

\noindent \textbf{Task specific queries:} 
The detection of small and irregular sized objects has remained a long-standing challenge. 
In our case, the flexible nature of MViTs facilitates using a range of human-understandable text queries. The queries can be chosen that best describe the special requirements needed in a given detection task. We demonstrate certain scenarios of how this feature can be exploited for better predictions. Fig.~\ref{fig:graph_2}~(left) shows an interesting case of how the text query ‘\txt{all little objects}’ improves recall for small objects as compared to a rather general text query. Similarly, Fig.~\ref{fig:graph_2}~(right) indicates how the use of special queries like ‘\txt{all long objects}’ helps improve the detection of irregular shaped objects (without any dataset specific fine-tuning!). 

\subsection{Open-world Object Detection}
The open-world setting assumes a realistic paradigm where a model can experience \emph{unknown objects} during training and inference \cite{bendale2015towards,dhamija2020overlooked,wang2021unidentified,joseph2021towards}. The goal is to identify unknowns and incrementally learn about them as and when new annotations are provided about a subset of unknowns. This stands in contrast to generic OD where models are trained to label unknown objects as background and only focus on the known objects. Here, we explore how a generic class-agnostic OD model can help with the open-world task to identify unknowns. As a case study, we apply our approach to a recent open-world detector (ORE) \cite{joseph2021towards}.

\noindent
$-$\textit{ORE Setting:} The authors distribute the 80 COCO \cite{coco} classes in four incremental learning tasks where 20 classes have been added to the known categories in each subsequent task. At each stage, the model must learn from the given subset of 20 newly introduced known classes, should not forget the previous known classes and must be able to detect unknown classes whose labelled examples have not been provided so far as the unknowns. ORE uses Faster-RCNN \cite{ren2015faster} as the base detector, with contrastive clustering in latent space and an energy-based classification head for unknown detection. It utilizes example-replay strategy \cite{wang2020frustratingly} for alleviating forgetting, when progressively learning the unknown categories once their labels become available. 

\noindent
$-$\textit{Unknown Pseudo-labels with MViTs:} ORE exploits the two-stage mechanism of Faster-RCNN \cite{ren2015faster} and uses proposals from the class-agnostic region proposal network (RPN) for pseudo-labelling of unknowns. The foreground object proposals with high objectness score which do not overlap with any ground-truth are labelled as unknowns. We note that since RPN is only trained on the objects of interest, its detections are overly sparse and lead to a low recall for unknowns. The pipeline therefore lacks a good proposal set that generalizes to \emph{novel objects}. We propose a variant of ORE, by using class-agnostic proposals for unknown object categories obtained from MAVL. For a fair comparison, the MViT is trained on a filtered dataset, generated by explicitly removing all captions from LMDet that contain any unknown category, leaving 0.76M image-text pairs {(see Appendix~\ref{appendix:ore} for further details)}. The results in Table~\ref{table7:owod} and Fig.~\ref{fig:ore_detreg} indicate significant improvements in unknown detection. {See Fig.~\ref{figure:owod} in Appendix~\ref{app:qual_results} for more qualitative results}.

\begin{table*}[!t]
\caption{\small MViT proposals are used for pseudo-labelling of unknowns in ORE \cite{joseph2021towards}. MAVL represents the model trained on a filtered dataset generated by \emph{removing} all captions from LMDet listing any of the 60 unknown categories evaluated in ORE. The results indicate a notable improvement in unknown detection. }
\begin{center}
\arrayrulecolor{black}
\resizebox{\linewidth}{!}{
\begin{tabular}{l|c|c|ccc|c|ccc|c|ccc} 
\toprule
\rowcolor{Gray} Task ID & \multicolumn{2}{c|}{Task 1} & \multicolumn{4}{c|}{Task 2} & \multicolumn{4}{c|}{Task 3} & \multicolumn{3}{c}{Task~ 4}
\\ 
\hline
 & mAP & {\cellcolor{orange!6}} & \multicolumn{3}{c|}{mAP} & {\cellcolor{orange!6}} & \multicolumn{3}{c|}{mAP} & {\cellcolor{orange!6}} & \multicolumn{3}{c}{mAP} \\
\cline{2-2}\cline{4-6}\cline{8-10}\cline{12-14}
 \multirow{-2}{*}{\begin{tabular}[c]{@{}c@{}}Pseudo-label \\ for Unknown\end{tabular}} & \begin{tabular}[c]{@{}c@{}}Current\\Known\end{tabular} & \multirow{-2}{*}{{\cellcolor{orange!6}}\begin{tabular}[c]{@{}>{\cellcolor{orange!6}}c@{}}R50 \\Unknown\end{tabular}} & \begin{tabular}[c]{@{}c@{}}Previous\\Known\end{tabular} & \begin{tabular}[c]{@{}c@{}}Current\\Known\end{tabular} & Both & \multirow{-2}{*}{{\cellcolor{orange!6}}\begin{tabular}[c]{@{}>{\cellcolor{orange!6}}c@{}}R50 \\Unknown\end{tabular}} & \begin{tabular}[c]{@{}c@{}}Previous\\Known\end{tabular} & \begin{tabular}[c]{@{}c@{}}Current\\Known\end{tabular} & Both & \multirow{-2}{*}{{\cellcolor{orange!6}}\begin{tabular}[c]{@{}>{\cellcolor{orange!6}}c@{}}R50 \\Unknown\end{tabular}} & \begin{tabular}[c]{@{}c@{}}Previous\\Known\end{tabular} & \begin{tabular}[c]{@{}c@{}}Current\\Known\end{tabular} & Both \\
\hline\hline
RPN & 63.4 & {\cellcolor{orange!6}} 14.4 & 58.3 & 30.8 & 45.1 & {\cellcolor{orange!6}} 11.3 & 43.3 & 23.4 & 36.7 & {\cellcolor{orange!6}} 14.8 & 37.2 & 20.7 & 33.1 \\
MAVL$^*$  & 64.0 & {\cellcolor{orange!6}} \textbf{50.1} & 61.6 & 30.8 & 46.2 & {\cellcolor{orange!6}} \textbf{49.5} & 43.8 & 22.7 & 36.8 & {\cellcolor{orange!6}} \textbf{50.9} & 36.2 & 20.6 & 32.3 \\
\bottomrule
\end{tabular}
}
\end{center}
\label{table7:owod}
\end{table*}

\subsection{Pretraining for Class-aware Object Detection}

\begin{wraptable}[8]{r}{6.5cm}
\caption{\small Effect of using MAVL proposals for pre-training of DETReg \cite{detreg} instead of Selective Search \cite{uijlings2013selective} proposals.}
\setlength{\tabcolsep}{4pt}
\resizebox{1.0\linewidth}{!}{
\begin{tabular}{l *{6}{c}}
  \toprule
  \rowcolor{Gray}
  Dataset$\rightarrow$ & \multicolumn{3}{c}{Pascal-VOC 10\%} & \multicolumn{3}{c}{Pascal-VOC 100\%} \\
  \rowcolor{Gray} Model $\downarrow$ & AP & AP50 & AP75 & AP & AP50 & AP75\\
  \midrule
  \midrule
  DETReg - SS & 51.4	& 72.2	& 56.6	& 63.5	& 83.3	& 70.3 \\
  DETReg - MAVL & \textbf{58.8} & \textbf{80.5} & \textbf{65.7} & \textbf{64.5} & \textbf{84.2} & \textbf{71.3} \\
  \bottomrule                             
\end{tabular}}
\label{table10:voc_results}
\end{wraptable}
The recent progress in self-supervised learning (SSL) \cite{PIRL,MoCo,SwAV,barlow_twins} has minimized the need for large \emph{labelled} datasets to achieve good performance on downstream tasks. These techniques encode the global image representation and achieve competitive generalization on various downstream tasks. However, these methods are suboptimal for class-aware OD, where the classification needs to be performed at local image patches (i.e. bounding boxes). Several recent efforts have been reported to address this challenge. ReSim \cite{xiao2021region} and DetCo \cite{xie2021detco} only pretrain the backbone to encode local and global representations. Whereas, DETReg \cite{detreg} pretrains both the backbone and detection network using off-the-shelf proposals from selective search \cite{uijlings2013selective} and achieves improvement over the previous methods. 

However, the proposals from heuristic selective search method, used in DETReg pretraining, are overly noisy and contain redundant boxes. We show that replacing these noisy pseudo-labels with MViT proposals can improve the downstream performance on OD task (Table \ref{table10:voc_results}). Following DETReg,  we select top-30 proposals from MAVL and pretrain the model for 50 epochs on ImageNet \cite{russakovsky2015imagenet} dataset, followed by fine-tuning on $10\%$ and $100\%$ data from Pascal VOC \cite{voc} for $150$ and $100$ epochs respectively. The results show an absolute gain of  $\sim7$ and $\sim1$ in AP in the two respective cases.

\begin{table}[!t]
\caption{\small
Proposals from MAVL are evaluated against state-of-the-art SOD and COD approaches. The \txt{general$^{\dagger}$} represents \lq \txt{all objects}\rq~text query. }
\setlength\tabcolsep{4pt}
  \begin{subtable}[t]{0.43\linewidth}
    \resizebox{\linewidth}{!}{%
      \begin{tabular}{l *{5}{c}}
          \toprule
         \rowcolor{Gray} Dataset $\rightarrow$ & &\multicolumn{2}{c}{DUT-OMRON} & \multicolumn{2}{c}{ECSSD} \\
          \rowcolor{Gray} Model $\downarrow$ & Text Query & AP50 & R50 & AP50 & R50\\
          \midrule
          CPD \cite{wu2019cascaded} & - & 64.5 & 77.4 & 87.1 & 92.7 \\
          PoolNet \cite{liu2019simple} & - & 66.5 & 78.8 & 87.4 & 93.1 \\
          \rowcolor{orange!6}
          MAVL & \txt{General$^{\dagger}$} & 67.0	 & 89.1  & 84.5  & 95.7 \\
          \rowcolor{orange!6}
          MAVL & \txt{Task specific$^{\dagger \dagger}$} & 75.5 & 93.3 & 85.7 & 96.1\\
          \bottomrule                             
        \end{tabular}
   }
   \caption{Salient OD (SOD). Here \txt{task specific$^{\dagger \dagger}$} query combines proposals from ‘\txt{all salient objects}' and ‘\txt{all foreground objects}' text queries.}
   \label{table7:saliency}
  \end{subtable}
  \begin{subtable}[t]{0.55\linewidth}
  \setlength{\tabcolsep}{2pt}
  \resizebox*{\linewidth}{!}{%
      \begin{tabular}{l *{7}{c}}
          \toprule
          \rowcolor{Gray} Dataset $\rightarrow$ & & \multicolumn{2}{c}{CHAMELEON} & \multicolumn{2}{c}{CAMO} & \multicolumn{2}{c}{COD10K} \\
          \rowcolor{Gray} Model $\downarrow$ & Text Query & AP50 & R50 & AP50 & R50 & AP50 & R50 \\
          \midrule
          SINET-V2 \cite{SINET-V2} & - & 67.3 & 76.7 & 56.5 & 77.2 & 44.4 & 66.6 \\
          \rowcolor{orange!6}
          MAVL & \txt{General$^{\dagger}$} & 30.2	& 53.3	& 46.5	& 75.4 & 39.6 & 67.8 \\
          \rowcolor{orange!6}
          MAVL & \txt{Task specific$^{\dagger \dagger}$} & 36.2 & 61.1 & 48.0 & 78.3 & 42.0 & 69.1 \\
          \bottomrule                             
        \end{tabular}
    }
    \caption{Camouflaged OD (COD) on three datasets. Here \txt{task specific$^{\dagger \dagger}$} query combines proposals from ‘\txt{all camouflaged objects}' and ‘\txt{all disguised objects}' text queries.}
    \label{table12:cod}
  \end{subtable}
\end{table}

\subsection{Salient Object Detection}
Given the generalized class-agnostic performance of MViTs on multiple domains, we evaluate their ability to distinguish between salient and non-salient parts of an image. We exploit the interactive nature of MViTs by passing specific queries to detect the salient objects. 
To this end, MAVL proposals generated with queries like ‘\txt{all salient objects}' are compared with PoolNet \cite{liu2019simple} and CPD \cite{wu2019cascaded} models that are specifically trained for predicting saliency maps. We evaluate the models on the DUT-OMRON \cite{yang2013saliency} and ECSSD \cite{shi2015hierarchical} datasets. These datasets are only used for MViT evaluation and are not used during training. Since MViTs generate bounding boxes, we convert the saliency ground-truths and the saliency maps predicted by CPD and PoolNet to bounding boxes using connected components labelling \cite{wu2005optimizing}. In the case of DUT-OMRON, the provided ground-truth bounding boxes are used by computing an average across the five human annotations. 

Table \ref{table7:saliency} indicates the effectiveness of MAVL in detecting the foreground salient objects. It is also interesting to note how the task specific$^{\dagger \dagger}$ query (e.g., ‘\txt{all salient/foreground objects}') provides better prediction of salient parts of the image in comparison to a more generic$^{\dagger}$ query like ‘\txt{all objects}' (Fig.~\ref{fig:appl_2}). See Appendix~\ref{app:add_results_sod} and Fig.~\ref{figure:sod-cod} in Appendix~\ref{app:qual_results} for additional details.
\begin{figure}[!t]
  \centering
  \begin{subfigure}{.5\textwidth}
    \centering
    \includegraphics[width=1.0\linewidth]{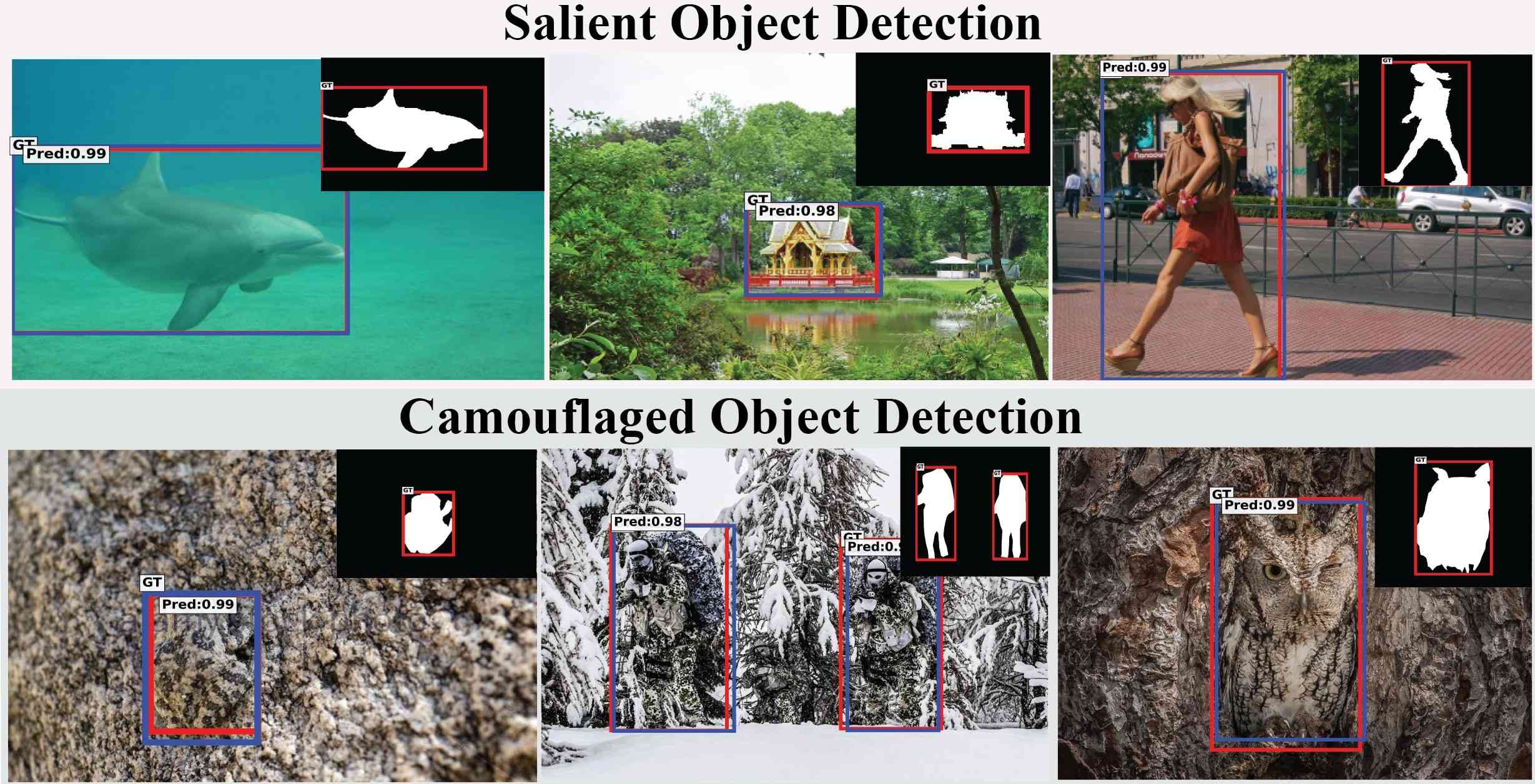}
    \caption{}
    \label{fig:appl_2}
  \end{subfigure}%
  \hfill
  \begin{subfigure}{.48\textwidth}
    \centering
    \includegraphics[height=2.7cm,width=1.0\linewidth]{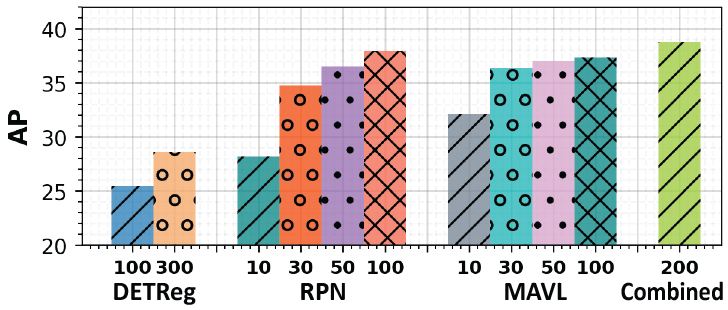}
    \caption{}
    \label{fig:graph_4_rpn}
  \end{subfigure}
\caption{\small
\textbf{\color{blue}(a)} Qualitative results of Salient (Top) and Camouflaged OD (Bottom). The ground-truth masks and boxes are shown on top right of the images. 
\textbf{\color{blue}(b)} 
Complimentary effect of using off-the-shelf proposals from MAVL in Faster RCNN \cite{ren2015faster} trained on COCO \cite{coco}, indicated as \lq combined\rq~(\ie RPN + MAVL). The x-axis shows the number of proposals. MAVL generates good quality proposals, which perform well even with  small proposal set sizes and demonstrate complimentary advantage to RPN.}
\label{fig:6-7_comb}
\end{figure}
\subsection{Camouflaged Object Detection}
Camouflaged object detection (COD) involves identifying objects that are \emph{seamlessly} embedded in their background. The objects have a similar texture to their surroundings and are difficult to locate as compared to salient or generic objects. Here, we explore the interactive OD capacity of MViTs on COD task by evaluating the performance of MAVL against the state-of-the-art model (SINET-V2 \cite{SINET-V2}) on CHAMELEON \cite{CHAMELEON}, CAMO \cite{CAMO} and COD10K \cite{fan2020camouflaged} datasets (Table \ref{table12:cod}). Similar to salient OD setting, we convert camouflage ground-truth masks and masks predicted by SINET-V2 to bounding boxes using connected components labelling \cite{wu2005optimizing}. However, the available bounding box ground-truths have been used for COD10K dataset. 
We note favorable performance of MAVL proposals, although the model is not specifically trained on camouflaged objects (Fig.~\ref{fig:appl_2}). This affirms the generality of MAVL proposals. See Appendix~\ref{app:add_results_cod} and Fig.~\ref{figure:sod-cod}.
\subsection{Improving Two-stage Object Detection}
The class-agnostic object proposals from MViTs have strong understanding of semantics and can be deployed along with the region proposal network (RPN) \cite{ren2015faster}. We observe an improvement in accuracy when off-the-shelf MAVL proposals are combined with RPN proposals in Faster RCNN \cite{ren2015faster} during inference (Fig.~\ref{fig:graph_4_rpn}). This indicates the complimentary nature of these proposals that is based on a rich top-down perception of the image content. 

Fig.~\ref{fig:graph_4_rpn} shows the results of replacing RPN proposals in Faster RCNN with DETReg \cite{detreg} and MAVL proposals. The results indicate that the supervised proposal generation methods (RPN and MAVL) perform well compared to the unsupervised method (DETReg). However, off-the-shelf MAVL proposals show better performance than RPN when using a small proposal set (\eg 10 proposals). Combining RPN and MAVL proposals improves the overall detection accuracy. 

\section{What makes MViTs a Generic Detector?}\label{sec:what}
Our empirical analysis shows the state-of-the-art performance of MViTs towards class-agnostic OD across different domains (Sec.~\ref{sec:multi-modal}) which positively impacts a number of downstream applications (Sec.~\ref{sec:app}). Having established this, we conduct a series of systematic experiments to explore the  contributing factors for representational learning of the general ‘\emph{objectness measure}’ in MViTs. Specifically, we identify the role of supervision and multi-modal learning as crucial factors.

\subsection{On the importance of supervision}
We consider two recent unsupervised learning models, DETReg \cite{detreg} and UP-DETR \cite{dai2021up}. DETReg trains Deformable DETR \cite{zhu2020deformable} to localize objects in class-agnostic fashion, with bounding box pseudo-labels from an off-the-shelf region
\begin{wraptable}[10]{r}{6.5cm}
\caption{\small MAVL proposals perform well compared to unsupervised methods (UP-DETR \cite{dai2021up} and DETReg \cite{detreg}) and supervised uni-modal method (Def-DETR \cite{zhu2020deformable}).} 
\setlength{\tabcolsep}{4pt}
\resizebox{1.0\linewidth}{!}{
\begin{tabular}{l *{7}{c}}
      \toprule
      \rowcolor{Gray} Dataset $\rightarrow$ &  & \multicolumn{2}{c}{Pascal-VOC} & \multicolumn{2}{c}{COCO} & \multicolumn{2}{c}{KITTI} \\
      \rowcolor{Gray} Model $\downarrow$ & Supervision & AP50 & R50 & AP50 & R50 & AP50 & R50\\
      \midrule
      \midrule
      UP-DETR & {unsupervised} & 0.56 & 16.6 & 0.19 & 6.56 & 0.01 & 0.65 \\
      DETReg  & {self-supervised} & 2.58 & 45.7 & 2.04 & 26.0 & 0.01 & 2.48 \\
      \midrule
      Def-DETR & {box-level} & 30.1 & 81.0 & 20.0 & 53.5 & 23.7 & 55.0 \\
      MAVL & {box + text} & 68.6 & 91.3 & 43.6 & 65.0 & 48.2 & 63.5 \\
      \bottomrule                             
    \end{tabular}}
\label{table4:supervision}
\end{wraptable}
proposal method (Selective Search \cite{uijlings2013selective}). Meanwhile, UP-DETR performs unsupervised pretraining on random query patches in an image for class-agnostic OD. Both the unsupervised models, DETReg and UP-DETR, are trained on uni-modal (Deformable DETR \cite{zhu2020deformable}) trained on LMDet in class-agnostic fashion, to evaluate the performance contributed by language supervision. We note that the image-level supervision with only box labels improves the performance in comparison with unsupervised methods. However, the use of caption texts aligned with input images proves to be vital and improves the performance approximately by \emph{two} times, highlighting the importance of multi-modal supervision. 

\subsection{How much does language contribute?}
\begin{wraptable}[12]{r}{6.2cm}
\caption{\small Effect of removing language branch from MViTs keeping the data loader structure intact. 
The performance is not affected largely as the language structure is still intact (boxes from caption are seen together).}
\setlength{\tabcolsep}{4pt}
\resizebox{1.0\linewidth}{!}{
\begin{tabular}{l *{7}{c}}
      \toprule
     \rowcolor{Gray} Dataset$\rightarrow$ &  & \multicolumn{2}{c}{Pascal-VOC} & \multicolumn{2}{c}{COCO} & \multicolumn{2}{c}{KITTI} \\
     \rowcolor{Gray} Model $\downarrow$ & Lang. & AP50 & R50 & AP50 & R50 & AP50 & R50\\
      \midrule
      \midrule
      MDETR & \checkmark & 63.9 & 88.0 & 38.1 & 58.5 & 42.5 & 60.9 \\ 
      MAVL &  \checkmark & 65.0 & 89.1 & 39.3 & 62.0 & 39.0 & 61.0 \\ 
      \midrule
      MDETR &  $\times$ & 59.7 & 86.4 & 33.4 & 57.9 & 36.9 & 55.0 \\
      MAVL & $\times$ & 61.6 & 86.7 & 34.4 & 58.3 & 36.5 & 58.9 \\
      \bottomrule                             
    \end{tabular}
    }                      
\label{table5:minus_lang}
\end{wraptable}
Given the importance of multi-modal supervision towards better performance, we find it pertinent to explore the benefit solely
from the language supervision. We conduct an ablation study on MDETR and MAVL,
by removing all textual inputs corresponding to captions, but keeping intact the structure introduced by language \ie learning to localize boxes corresponding to a 
caption for each image in an iteration (without any language branch). Both MDETR and MAVL are trained on LMDet containing
aligned image-text pairs.
Here, the structure in which the information is fed during training is of high importance to us. Each image may have multiple captions, and hence it may be seen multiple times in the same iteration, but with varying contexts. The experimental setup removes all captions during training and evaluations, however keeps the described data loader structure intact, thus having approximately 1.3M iterations in an epoch. All models use ResNet-101 backbone and are evaluated after 10 epochs for ablation (instead of total 20 epochs).
Table \ref{table5:minus_lang} indicate that visual branch plays a vital role, however the importance of language cannot be ruled out since the boxes related to a caption are still seen together. We analyze the importance of this implicit language structure next. 
 
\noindent\textbf{Ablation on language structure:} 
The above experimental results reveal that removal of textual information does not significantly affect model performance. However, a further ablation on the structure introduced by language
is required for the completeness of this evaluation. As such, we conduct ablations at five levels using Deformable DETR \cite{zhu2020deformable}, as shown in Table \ref{table6:ablation_lang_struct}. First, all the annotations in LMDet are combined at image level by concatenating the bounding boxes of all captions corresponding to an image (Setting-1). This removes any prior information introduced by the language structure. Then, class-agnostic NMS is applied at a threshold of $0.9$ to filter boxes that have high overlaps (Setting-2). To imitate the repetitive pattern introduced during training, bounding box annotations corresponding to an image are randomly sampled and grouped (Setting-3).
\begin{wraptable}[12]{r}{6.5cm}
\caption{\small Experimental analysis to explore the contribution of language by removing all textual inputs, but maintaining the structure introduced by captions. Experiments are performed on Def-DETR \cite{zhu2020deformable} using LMDet.}
\setlength{\tabcolsep}{3pt}
\resizebox{1.0\linewidth}{!}{
\begin{tabular}{cccccccc}
\toprule
\rowcolor{Gray}  &  & \multicolumn{2}{c}{Pascal-VOC} & \multicolumn{2}{c}{MSCOCO} & \multicolumn{2}{c}{KITTI} \\
\rowcolor{Gray} Experiment & \multirow{-2}{*}{\begin{tabular}[c]{@{}c@{}}Language\\Structure\end{tabular}} & AP50 & R50 & AP50 & R50 & AP50 & R50 \\
\midrule
\midrule
 Setting-1&  $\times$& 16.2 & 74.5 & 10.7 & 47.0 & 19.4 & 57.3   \\
 Setting-2&  $\times$& 30.1 & 81.0 & 20.0 & 53.5 & 23.7 & 55.0 \\
 Setting-3&  $\times$& 33.8 & 82.5 & 19.3 & 55.8 & 21.2 & 52.7 \\
 Setting-4&  $\times$& 35.1 & 82.7 & 21.2 & 56.3 & 21.5 & 58.5 \\
 \midrule
 Setting-5&  $\checkmark$ & \textbf{61.6} & \textbf{86.7} & \textbf{34.4} & \textbf{58.3} & \textbf{36.5} & \textbf{58.9} \\
\bottomrule
\end{tabular}
}
\label{table6:ablation_lang_struct}
\end{wraptable}
The number of samples in a combination is kept close to the average number of boxes in image-text pairs
in original MAVL training ($\sim$6 boxes). Finally, a longer training schedule is used in the same setting to replicate a scenario closer to the original MAVL training (Setting-4). These four settings are then compared with a model that is trained without any captions, but maintains the structure introduced by language (Setting-5, same as Table~\ref{table5:minus_lang} last row). This analysis indicates that language structure has significant impact in learning a general notion of objectness. With the use of aligned image-text pairs, additional contextual information is provided to the model. As objects generally tend to co-occur with other objects and certain scenes, such contexual association can be exploited for visual understanding \cite{OLIVA2007520}. Use of captions that describe a scene conveys such a notion of co-occurring objects and their mutual relationships, indicating that the structure introduced by language provides rich semantic and spatial context. Consistent with our findings, other recent efforts also indicate strong generalization achieved using the context encoded within natural language \cite{zhang2020putting,radford2021learning,zareian2021open,zhou2021uc2}.

\section{Conclusion}
This paper demonstrates intriguing performance of MViTs, trained only on natural images, for generic OD across a diverse set of domains. We systematically study the main reasons for this generalization, and note that the language structure available in  image-caption pairs used to train MViTs plays a key role. Based on these insights, we develop a more flexible and efficient MViT for off-the-shelf class-agnostic OD, that can be instantiated with different text queries to generate desired proposal sets. Furthermore, we show various use-cases where class-agnostic proposals can be used to improve performance \eg open-world OD, camouflaged and salient OD, supervised and self-supervised OD.\\

\noindent \textbf{Acknowledgements.} Ming-Hsuan Yang is supported by the NSF CAREER grant 1149783. Fahad Shahbaz Khan is supported by the VR starting grant (2016-05543).

\clearpage
\bibliographystyle{splncs04}
\bibliography{egbib}

\begin{thebibliography}{10}
\providecommand{\url}[1]{\texttt{#1}}
\providecommand{\urlprefix}{URL }
\providecommand{\doi}[1]{https://doi.org/#1}

\bibitem{alexe2010object}
Alexe, B., Deselaers, T., Ferrari, V.: {What is an object?} In: Proceedings of
  the IEEE/CVF Conference on Computer Vision and Pattern Recognition. pp.
  73--80. IEEE (2010)

\bibitem{alexe2012measuring}
Alexe, B., Deselaers, T., Ferrari, V.: {Measuring the Objectness of Image
  Windows}. IEEE Transactions on Pattern Analysis and Machine Intelligence
  \textbf{34}(11),  2189--2202 (2012)

\bibitem{detreg}
Bar, A., Wang, X., Kantorov, V., Reed, C.J., Herzig, R., Chechik, G., Rohrbach,
  A., Darrell, T., Globerson, A.: {DETReg: Unsupervised Pretraining with Region
  Priors for Object Detection}. In: Proceedings of the IEEE/CVF Conference on
  Computer Vision and Pattern Recognition (2022)

\bibitem{bendale2015towards}
Bendale, A., Boult, T.: {Towards Open World Recognition}. In: Proceedings of
  the IEEE/CVF Conference on Computer Vision and Pattern Recognition. pp.
  1893--1902 (2015)

\bibitem{DETR}
Carion, N., Massa, F., Synnaeve, G., Usunier, N., Kirillov, A., Zagoruyko, S.:
  {End-to-End Object Detection with Transformers}. In: The European Conference
  on Computer Vision. pp. 213--229. Springer (2020)

\bibitem{SwAV}
Caron, M., Misra, I., Mairal, J., Goyal, P., Bojanowski, P., Joulin, A.:
  {Unsupervised Learning of Visual Features by Contrasting Cluster
  Assignments}. In: Advances in Neural Information Processing Systems (2020)

\bibitem{caron2021emerging}
Caron, M., Touvron, H., Misra, I., J{\'e}gou, H., Mairal, J., Bojanowski, P.,
  Joulin, A.: {Emerging Properties in Self-Supervised Vision Transformers}.
  arXiv preprint arXiv:2104.14294  (2021)

\bibitem{pmlr-v119-chen20j}
Chen, T., Kornblith, S., Norouzi, M., Hinton, G.: {A Simple Framework for
  Contrastive Learning of Visual Representations}. In: International Conference
  on Machine Learning. pp. 1597--1607. PMLR (2020)

\bibitem{UNITER}
Chen, Y.C., Li, L., Yu, L., El~Kholy, A., Ahmed, F., Gan, Z., Cheng, Y., Liu,
  J.: {UNITER: UNiversal Image-TExt Representation Learning}. In: The European
  Conference on Computer Vision. pp. 104--120. Springer (2020)

\bibitem{cheng2014bing}
Cheng, M.M., Zhang, Z., Lin, W.Y., Torr, P.: {BING: Binarized Normed Gradients
  for Objectness Estimation at 300fps}. In: Proceedings of the IEEE/CVF
  Conference on Computer Vision and Pattern Recognition. pp. 3286--3293 (2014)

\bibitem{dai2021up}
Dai, Z., Cai, B., Lin, Y., Chen, J.: {UP-DETR: Unsupervised Pre-training for
  Object Detection with Transformers}. In: Proceedings of the IEEE/CVF
  Conference on Computer Vision and Pattern Recognition. pp. 1601--1610 (2021)

\bibitem{BERT}
Devlin, J., Chang, M.W., Lee, K., Toutanova, K.: {BERT: Pre-training of Deep
  Bidirectional Transformers for Language Understanding}. In: NAACL (2019)

\bibitem{dhamija2020overlooked}
Dhamija, A., Gunther, M., Ventura, J., Boult, T.: {The Overlooked Elephant of
  Object Detection: Open Set}. In: Proceedings of the IEEE/CVF Winter
  Conference on Applications of Computer Vision. pp. 1021--1030 (2020)

\bibitem{voc}
Everingham, M., Van~Gool, L., Williams, C.K., Winn, J., Zisserman, A.: {The
  Pascal Visual Object Classes (VOC) Challenge}. International Journal of
  Computer Vision  \textbf{88}(2),  303--338 (2010)

\bibitem{SINET-V2}
Fan, D.P., Ji, G.P., Cheng, M.M., Shao, L.: {Concealed Object Detection}. IEEE
  Transactions on Pattern Analysis and Machine Intelligence  (2021)

\bibitem{fan2020camouflaged}
Fan, D.P., Ji, G.P., Sun, G., Cheng, M.M., Shen, J., Shao, L.: {Camouflaged
  Object Detection}. In: Proceedings of the IEEE/CVF Conference on Computer
  Vision and Pattern Recognition. pp. 2777--2787 (2020)

\bibitem{kitti}
Geiger, A., Lenz, P., Urtasun, R.: {Are we ready for autonomous driving? The
  KITTI vision benchmark suite}. In: Proceedings of the IEEE/CVF Conference on
  Computer Vision and Pattern Recognition. pp. 3354--3361. IEEE (2012)

\bibitem{kitchen}
Georgakis, G., Reza, M.A., Mousavian, A., Le, P.H., Ko{\v{s}}eck{\'a}, J.:
  {Multiview RGB-D Dataset for Object Instance Detection}. In: CoRR. pp.
  426--434. IEEE (2016)

\bibitem{gupta2019lvis}
Gupta, A., Dollar, P., Girshick, R.: {LVIS: A Dataset for Large Vocabulary
  Instance Segmentation}. In: Proceedings of the IEEE/CVF Conference on
  Computer Vision and Pattern Recognition. pp. 5356--5364 (2019)

\bibitem{gpv1}
Gupta, T., Kamath, A., Kembhavi, A., Hoiem, D.: {Towards General Purpose Vision
  Systems}. In: Proceedings of the IEEE/CVF Conference on Computer Vision and
  Pattern Recognition. pp. 16399--16409 (2022)

\bibitem{MoCo}
He, K., Fan, H., Wu, Y., Xie, S., Girshick, R.: {Momentum Contrast for
  Unsupervised Visual Representation Learning}. In: Proceedings of the IEEE/CVF
  Conference on Computer Vision and Pattern Recognition. pp. 9729--9738 (2020)

\bibitem{he2017mask}
He, K., Gkioxari, G., Doll{\'a}r, P., Girshick, R.: {Mask R-CNN}. In:
  Proceedings of the IEEE/CVF Conference on Computer Vision and Pattern
  Recognition. pp. 2961--2969 (2017)

\bibitem{he2016deep}
He, K., Zhang, X., Ren, S., Sun, J.: {Deep Residual Learning for Image
  Recognition}. In: Proceedings of the IEEE/CVF Conference on Computer Vision
  and Pattern Recognition. pp. 770--778 (2016)

\bibitem{spacy}
Honnibal, M., Montani, I.: {spaCy: Industrial-strength Natural Language
  Processing in Python}  (2020)

\bibitem{hudson2019gqa}
Hudson, D.A., Manning, C.D.: Gqa: A new dataset for real-world visual reasoning
  and compositional question answering. In: Proceedings of the IEEE/CVF
  Conference on Computer Vision and Pattern Recognition. pp. 6700--6709 (2019)

\bibitem{clipart-comic-water}
Inoue, N., Furuta, R., Yamasaki, T., Aizawa, K.: {Cross-Domain
  Weakly-Supervised Object Detection Through Progressive Domain Adaptation}.
  In: Proceedings of the IEEE/CVF Conference on Computer Vision and Pattern
  Recognition. pp. 5001--5009 (2018)

\bibitem{jaiswal2021class}
Jaiswal, A., Wu, Y., Natarajan, P., Natarajan, P.: {Class-agnostic Object
  Detection}. In: Proceedings of the IEEE/CVF Winter Conference on Applications
  of Computer Vision. pp. 919--928 (2021)

\bibitem{joseph2021towards}
Joseph, K., Khan, S., Khan, F.S., Balasubramanian, V.N.: {Towards Open World
  Object Detection}. In: Proceedings of the IEEE/CVF Conference on Computer
  Vision and Pattern Recognition. pp. 5830--5840 (2021)

\bibitem{mdetr}
Kamath, A., Singh, M., LeCun, Y., Synnaeve, G., Misra, I., Carion, N.:
  {MDETR--Modulated Detection for End-to-End Multi-Modal Understanding}. In:
  Proceedings of the IEEE/CVF International Conference on Computer Vision. pp.
  1780--1790 (2021)

\bibitem{kazemzadeh2014referitgame}
Kazemzadeh, S., Ordonez, V., Matten, M., Berg, T.: Referitgame: Referring to
  objects in photographs of natural scenes. In: Conference on Empirical Methods
  in Natural Language Processing

\bibitem{kim2021learning}
Kim, D., Lin, T.Y., Angelova, A., Kweon, I.S., Kuo, W.: {Learning Open-World
  Object Proposals without Learning to Classify}. arXiv preprint
  arXiv:2108.06753  (2021)

\bibitem{krishna2017visual}
Krishna, R., Zhu, Y., Groth, O., Johnson, J., Hata, K., Kravitz, J., Chen, S.,
  Kalantidis, Y., Li, L.J., Shamma, D.A., et~al.: Visual genome: Connecting
  language and vision using crowdsourced dense image annotations. International
  Journal of Computer Vision  \textbf{123}(1),  32--73 (2017)

\bibitem{kuo2015deepbox}
Kuo, W., Hariharan, B., Malik, J.: {DeepBox: Learning Objectness with
  Convolutional Networks}. In: Proceedings of the IEEE/CVF Conference on
  Computer Vision and Pattern Recognition. pp. 2479--2487 (2015)

\bibitem{OpenImages}
Kuznetsova, A., Rom, H., Alldrin, N., Uijlings, J., Krasin, I., Pont-Tuset, J.,
  Kamali, S., Popov, S., Malloci, M., Kolesnikov, A., et~al.: {The open images
  dataset v4}. IJCV  \textbf{128}(7),  1956--1981 (2020)

\bibitem{CAMO}
Le, T.N., Nguyen, T.V., Nie, Z., Tran, M.T., Sugimoto, A.: {Anabranch network
  for camouflaged object segmentation}. Computer Vision and Image Understanding
   \textbf{184},  45--56 (2019)

\bibitem{VisualBERT}
Li, L.H., Yatskar, M., Yin, D., Hsieh, C.J., Chang, K.W.: {VisualBERT: A Simple
  and Performant Baseline for Vision and Language}. arXiv preprint
  arXiv:1908.03557  (2019)

\bibitem{OSCAR}
Li, X., Yin, X., Li, C., Zhang, P., Hu, X., Zhang, L., Wang, L., Hu, H., Dong,
  L., Wei, F., et~al.: {Oscar: Object-Semantics Aligned Pre-training for
  Vision-Language Tasks}. In: The European Conference on Computer Vision. pp.
  121--137. Springer (2020)

\bibitem{lin2017feature}
Lin, T.Y., Doll{\'a}r, P., Girshick, R., He, K., Hariharan, B., Belongie, S.:
  {Feature Pyramid Networks for Object Detection}. In: Proceedings of the
  IEEE/CVF Conference on Computer Vision and Pattern Recognition. pp.
  2117--2125 (2017)

\bibitem{lin2017focal}
Lin, T.Y., Goyal, P., Girshick, R., He, K., Doll{\'a}r, P.: {Focal Loss for
  Dense Object Detection}. In: Proceedings of the IEEE/CVF Conference on
  Computer Vision and Pattern Recognition. pp. 2980--2988 (2017)

\bibitem{coco}
Lin, T.Y., Maire, M., Belongie, S., Hays, J., Perona, P., Ramanan, D.,
  Doll{\'a}r, P., Zitnick, C.L.: {Microsoft COCO: Common Objects in Context}.
  In: The European Conference on Computer Vision. pp. 740--755. Springer (2014)

\bibitem{liu2019simple}
Liu, J.J., Hou, Q., Cheng, M.M., Feng, J., Jiang, J.: {A Simple Pooling-Based
  Design for Real-Time Salient Object Detection}. In: Proceedings of the
  IEEE/CVF Conference on Computer Vision and Pattern Recognition. pp.
  3917--3926 (2019)

\bibitem{liu2020deep}
Liu, L., Ouyang, W., Wang, X., Fieguth, P., Chen, J., Liu, X., Pietik{\"a}inen,
  M.: {Deep Learning for Generic Object Detection: A Survey}. International
  Journal of Computer Vision  \textbf{128}(2),  261--318 (2020)

\bibitem{roberta}
Liu, Y., Ott, M., Goyal, N., Du, J., Joshi, M., Chen, D., Levy, O., Lewis, M.,
  Zettlemoyer, L., Stoyanov, V.: {RoBERTa: A Robustly Optimized BERT
  Pretraining Approach}. arXiv preprint arXiv:1907.11692  (2019)

\bibitem{ViLBERT}
Lu, J., Batra, D., Parikh, D., Lee, S.: {ViLBERT: Pretraining Task-Agnostic
  Visiolinguistic Representations for Vision-and-Language Tasks }. In: Advances
  in Neural Information Processing Systems (2019)

\bibitem{12-in-1}
Lu, J., Goswami, V., Rohrbach, M., Parikh, D., Lee, S.: {12-in-1: Multi-Task
  Vision and Language Representation Learning}. In: Proceedings of the IEEE/CVF
  Conference on Computer Vision and Pattern Recognition. pp. 10437--10446
  (2020)

\bibitem{PIRL}
Misra, I., Maaten, L.v.d.: {Self-Supervised Learning of Pretext-Invariant
  Representations}. In: Proceedings of the IEEE/CVF Conference on Computer
  Vision and Pattern Recognition. pp. 6707--6717 (2020)

\bibitem{OLIVA2007520}
Oliva, A., Torralba, A.: The role of context in object recognition. Trends in
  Cognitive Sciences  \textbf{11}(12),  520--527 (2007)

\bibitem{OT}
Peyré, G., Cuturi, M.: {Computational Optimal Transport} (2020)

\bibitem{pinheiro2015learning}
Pinheiro, P.O., Collobert, R., Dollár, P.: {Learning to Segment Object
  Candidates}. In: Advances in Neural Information Processing Systems (2015)

\bibitem{pinheiro2016learning}
Pinheiro, P.O., Lin, T.Y., Collobert, R., Dollár, P.: {Learning to Refine
  Object Segments}. In: The European Conference on Computer Vision (2016)

\bibitem{plummer2015flickr30k}
Plummer, B.A., Wang, L., Cervantes, C.M., Caicedo, J.C., Hockenmaier, J.,
  Lazebnik, S.: Flickr30k entities: Collecting region-to-phrase correspondences
  for richer image-to-sentence models. In: Proceedings of the IEEE/CVF
  Conference on Computer Vision and Pattern Recognition. pp. 2641--2649 (2015)

\bibitem{pont2016multiscale}
Pont-Tuset, J., Arbelaez, P., Barron, J.T., Marques, F., Malik, J.: {Multiscale
  Combinatorial Grouping for Image Segmentation and Object Proposal
  Generation}. IEEE Transactions on Pattern Analysis and Machine Intelligence
  \textbf{39}(1),  128--140 (2016)

\bibitem{radford2021learning}
Radford, A., Kim, J.W., Hallacy, C., Ramesh, A., Goh, G., Agarwal, S., Sastry,
  G., Askell, A., Mishkin, P., Clark, J., Krueger, G., Sutskever, I.: {Learning
  Transferable Visual Models From Natural Language Supervision}. In:
  International Conference on Machine Learning (2021)

\bibitem{ren2015faster}
Ren, S., He, K., Girshick, R., Sun, J.: {Faster R-CNN: Towards Real-Time Object
  Detection with Region Proposal Networks}. Advances in Neural Information
  Processing Systems  \textbf{28},  91--99 (2015)

\bibitem{russakovsky2015imagenet}
Russakovsky, O., Deng, J., Su, H., Krause, J., Satheesh, S., Ma, S., Huang, Z.,
  Karpathy, A., Khosla, A., Bernstein, M., et~al.: {ImageNet Large Scale Visual
  Recognition Challenge}. International Journal of Computer Vision
  \textbf{115}(3),  211--252 (2015)

\bibitem{shao2019objects365}
Shao, S., Li, Z., Zhang, T., Peng, C., Yu, G., Zhang, X., Li, J., Sun, J.:
  Objects365: A large-scale, high-quality dataset for object detection. In:
  Proceedings of the IEEE/CVF International Conference on Computer Vision. pp.
  8430--8439 (2019)

\bibitem{shi2015hierarchical}
Shi, J., Yan, Q., Xu, L., Jia, J.: {Hierarchical Image Saliency Detection on
  Extended CSSD}. IEEE Transactions on Pattern Analysis and Machine
  Intelligence  \textbf{38}(4),  717--729 (2015)

\bibitem{simeoni2021localizing}
Sim{\'e}oni, O., Puy, G., Vo, H.V., Roburin, S., Gidaris, S., Bursuc, A.,
  P{\'e}rez, P., Marlet, R., Ponce, J.: {Localizing Objects with
  Self-Supervised Transformers and no Labels}. In: British Machine Vision
  Conference (2021)

\bibitem{CHAMELEON}
Skurowski, P., Abdulameer, H., B{\l}aszczyk, J., Depta, T., Kornacki, A.,
  Kozie{\l}, P.: {Animal Camouflage Analysis: CHAMELEON Database}. Unpublished
  Manuscript  \textbf{2}(6), ~7 (2018)

\bibitem{VL-BERT}
Su, W., Zhu, X., Cao, Y., Li, B., Lu, L., Wei, F., Dai, J.: {VL-BERT:
  Pre-training of Generic Visual-Linguistic Representations}. In: International
  Conference on Learning Representations (2019)

\bibitem{sun2019videobert}
Sun, C., Myers, A., Vondrick, C., Murphy, K., Schmid, C.: {VideoBERT: A Joint
  Model for Video and Language Representation Learning}. In: Proceedings of the
  IEEE/CVF Conference on Computer Vision and Pattern Recognition. pp.
  7464--7473 (2019)

\bibitem{LXMERT}
Tan, H., Bansal, M.: {LXMERT: Learning Cross-Modality Encoder Representations
  from Transformers}. In: Conference on Empirical Methods in Natural Language
  Processing (2019)

\bibitem{tan2019efficientnet}
Tan, M., Le, Q.: {EfficientNet: Rethinking Model Scaling for Convolutional
  Neural Networks}. In: International Conference on Machine Learning. pp.
  6105--6114. PMLR (2019)

\bibitem{uijlings2013selective}
Uijlings, J.R., Van De~Sande, K.E., Gevers, T., Smeulders, A.W.: {Selective
  Search for Object Recognition}. International Journal of Computer Vision
  \textbf{104}(2),  154--171 (2013)

\bibitem{wang2021unidentified}
Wang, W., Feiszli, M., Wang, H., Tran, D.: {Unidentified Video Objects: A
  Benchmark for Dense, Open-World Segmentation}. arXiv preprint
  arXiv:2104.04691  (2021)

\bibitem{wang2020frustratingly}
Wang, X., Huang, T.E., Darrell, T., Gonzalez, J.E., Yu, F.: {Frustratingly
  Simple Few-Shot Object Detection}. arXiv preprint arXiv:2003.06957  (2020)

\bibitem{universal}
Wang, X., Cai, Z., Gao, D., Vasconcelos, N.: {Towards Universal Object
  Detection by Domain Attention}. In: Proceedings of the IEEE/CVF Conference on
  Computer Vision and Pattern Recognition. pp. 7289--7298 (2019)

\bibitem{rw2019timm}
Wightman, R.: {PyTorch Image Models}.
  \url{https://github.com/rwightman/pytorch-image-models} (2019).
  \doi{10.5281/zenodo.4414861}

\bibitem{wu2005optimizing}
Wu, K., Otoo, E., Shoshani, A.: {Optimizing connected component labeling
  algorithms}. In: Medical Imaging 2005: Image Processing. vol.~5747, pp.
  1965--1976. International Society for Optics and Photonics (2005)

\bibitem{wu2019detectron2}
Wu, Y., Kirillov, A., Massa, F., Lo, W.Y., Girshick, R.: {Detectron2}.
  \url{https://github.com/facebookresearch/detectron2} (2019)

\bibitem{wu2019cascaded}
Wu, Z., Su, L., Huang, Q.: {Cascaded Partial Decoder for Fast and Accurate
  Salient Object Detection}. In: Proceedings of the IEEE/CVF Conference on
  Computer Vision and Pattern Recognition. pp. 3907--3916 (2019)

\bibitem{dota}
Xia, G.S., Bai, X., Ding, J., Zhu, Z., Belongie, S., Luo, J., Datcu, M.,
  Pelillo, M., Zhang, L.: {DOTA: A Large-scale Dataset for Object Detection in
  Aerial Images}. In: Proceedings of the IEEE/CVF Conference on Computer Vision
  and Pattern Recognition. pp. 3974--3983 (2018)

\bibitem{xiao2021region}
Xiao, T., Reed, C.J., Wang, X., Keutzer, K., Darrell, T.: {Region Similarity
  Representation Learning}. In: Proceedings of the IEEE/CVF International
  Conference on Computer Vision (2021)

\bibitem{xie2021detco}
Xie, E., Ding, J., Wang, W., Zhan, X., Xu, H., Sun, P., Li, Z., Luo, P.:
  {DetCo: Unsupervised Contrastive Learning for Object Detection}. In:
  Proceedings of the IEEE/CVF Conference on Computer Vision and Pattern
  Recognition. pp. 8392--8401 (2021)

\bibitem{noisy_student}
Xie, Q., Luong, M.T., Hovy, E., Le, Q.V.: {Self-training with Noisy Student
  improves ImageNet classification}. In: Proceedings of the IEEE/CVF Conference
  on Computer Vision and Pattern Recognition. pp. 10687--10698 (2020)

\bibitem{yan2018deeplesion}
Yan, K., Wang, X., Lu, L., Summers, R.M.: {DeepLesion: automated mining of
  large-scale lesion annotations and universal lesion detection with deep
  learning}. Journal of medical imaging  \textbf{5}(3),  036501 (2018)

\bibitem{yang2013saliency}
Yang, C., Zhang, L., Lu, H., Ruan, X., Yang, M.H.: {Saliency Detection via
  Graph-Based Manifold Ranking}. In: Proceedings of the IEEE/CVF Conference on
  Computer Vision and Pattern Recognition. pp. 3166--3173 (2013)

\bibitem{zareian2021open}
Zareian, A., Rosa, K.D., Hu, D.H., Chang, S.F.: {Open-Vocabulary Object
  Detection Using Captions}. In: Proceedings of the IEEE/CVF Conference on
  Computer Vision and Pattern Recognition. pp. 14393--14402 (2021)

\bibitem{barlow_twins}
Zbontar, J., Jing, L., Misra, I., LeCun, Y., Deny, S.: {Barlow Twins:
  Self-Supervised Learning via Redundancy Reduction}. In: International
  Conference on Machine Learning (2021)

\bibitem{zhang2020putting}
Zhang, M., Tseng, C., Kreiman, G.: {Putting visual object recognition in
  context}. In: Proceedings of the IEEE/CVF Conference on Computer Vision and
  Pattern Recognition. pp. 12985--12994 (2020)

\bibitem{zhang2015bing++}
Zhang, Z., Liu, Y., Chen, X., Zhu, Y., Cheng, M.M., Saligrama, V., Torr, P.H.:
  {BING++: A Fast High Quality Object Proposal Generator at 100fps}. In: IEEE
  Transactions on Pattern Analysis and Machine Intelligence. vol.~40, pp.
  1209--1223 (2018)

\bibitem{zhou2021uc2}
Zhou, M., Zhou, L., Wang, S., Cheng, Y., Li, L., Yu, Z., Liu, J.: {UC2:
  Universal Cross-lingual Cross-modal Vision-and-Language Pre-training}. In:
  Proceedings of the IEEE/CVF Conference on Computer Vision and Pattern
  Recognition. pp. 4155--4165 (2021)

\bibitem{zhu2020deformable}
Zhu, X., Su, W., Lu, L., Li, B., Wang, X., Dai, J.: {Deformable DETR:
  Deformable Transformers for End-to-End Object Detection}. In: International
  Conference on Learning Representations (2021)

\bibitem{zitnick2014edge}
Zitnick, C.L., Doll{\'a}r, P.: {Edge Boxes: Locating Object Proposals from
  Edges}. In: The European Conference on Computer Vision. pp. 391--405.
  Springer (2014)

\end{thebibliography}
\clearpage

\appendix
\begin{center}
\textbf{\Large Supplemental Material}
\end{center}

In this section, we provide additional information regarding,
\begin{itemize}
    \item Implementation details (Appendix~\ref{app:imp_detail})
    \item Limitations (Appendix~\ref{app:limitations})
    \item Qualitative results (Appendix~\ref{app:qual_results})
    \item Additional results (Appendix~\ref{app:add_results})
    \item Related works (Appendix~\ref{app:related_work})
\end{itemize}

\section{Implementation Details}\label{app:imp_detail}
\subsection{MAVL}\label{app:imp_detail_mdef}
Similar to MDETR \cite{mdetr}, we train MAVL on LMDet dataset containing approximately 1.3M aligned image-text pairs. Unlike MDETR which converges in 40 epochs, our MAVL converges only in 20 epochs with overall better class-agnostic object detection (OD) accuracy. However, the inference for MAVL is approximately 30\% slower (see Table~\ref{table:mdetr_vs_mdef_detr}).

MAVL is trained using a learning rate of $1e^{-3}$ which decays by a factor of 10 after 16 epochs. The vision backbone (ResNet-101~\cite{he2016deep}) and language backbone (RoBERTa~\cite{roberta}) use learning rates of $1e^{-4}$ and $1e^{-5}$ respectively. The number of object queries is set to 300. In the late-fusion transformer, a series of six self-attention blocks are used, where a detection head is applied after each block for calculating the individual auxiliary losses which are then summed up (see Fig.~\ref{mvit_block_diagram} in the main paper).

\begin{table}[!ht]
\caption{Comparison of MDETR \cite{mdetr} and MAVL (ours) in terms of convergence epochs, parameters, inference speed and class-agnostic OD performance on COCO \cite{coco} dataset. MAVL converges in half epochs with better accuracy at the cost of slightly slower inference. The frames per second (FPS) are measured on a Quadro RTX 6000 GPU by averaging the time for $1K$ inference passes.}
\begin{center}
\resizebox{0.8\linewidth}{!}{\setlength{\tabcolsep}{4pt}
\begin{tabular}{l*{4}{c}}
  \toprule
  Model & Epochs & Parameters & Inference FPS & COCO AP50 \\
  \midrule
  MDETR & 40 & 185M & 13.0 & 40.7 \\
  MAVL & 20 & 188M & 8.95 & 43.6 \\
  \bottomrule                             
\end{tabular}}
\end{center}
\label{table:mdetr_vs_mdef_detr}
\end{table}
\subsection{MViTs as Class Agnostic Object Detectors}
\label{appendix:class_agnostic_evaluation}
We explore the interactive nature of multi-modal vision transformers (MViTs) for class-agnostic OD task. We construct intuitive natural language text queries by exploring the semantic space of MViTs using an open-source natural language processing (NLP) library, spacy \cite{spacy}. Specifically, we find words closer to the keyword ‘\txt{object}’ in the semantic space and construct multiple text queries for the class-agnostic OD task. The detected boxes from the multiple text queries are combined, a class-agnostic non-maximum suppression (NMS) at IoU threshold of 0.5 is applied and top-N boxes are selected. We use N=50 and report average precision and recall at IoU threshold of 0.5 in all experiments. For the salient and camouflaged object detection (SOD and COD) tasks, we only consider boxes with objectness scores greater than 0.7.

For Pascal VOC \cite{voc}, COCO \cite{coco}, Objects365 \cite{shao2019objects365}, LVIS \cite{gupta2019lvis}, Clipart, Comic and Watercolor \cite{clipart-comic-water}, we use combined detections from queries ‘\txt{all objects}’, ‘\txt{all entities}’, ‘\txt{all visible entities and objects}’, and ‘\txt{all obscure entities and objects}’. Additionally, ‘\txt{all small objects}’ text query is included for the evaluation on KITTI \cite{kitti}, Kitchen \cite{kitchen} and DOTA \cite{dota} as these datasets have a larger number of small sized objects. Moreover, multi-scale evaluation is used for DOTA dataset due to the significant scale variations in the satellite imagery. Here the original image is split into 8 equal crops and the detections from the individual crops are combined. We observe the multi-scale inference improves the performance on DOTA as it contains more tiny objects as compared to other datasets.

\subsection{Detection of Small Objects}
We observe that the targeted queries like ‘\txt{all small objects}’ and ‘\txt{all little objects}’ can improve the detection accuracy of small objects as compared to a rather general text query ‘\txt{all objects}’. Quantitative and qualitative results are presented in Fig.~\ref{fig:graph_2} (main paper). For quantitative comparison, all objects covering less than 5\% of the image area are considered small, between 5\% and 20\% are considered medium and greater than 20\% are considered large. 
\subsection{Open-world Object Detection}
\label{appendix:ore}
The proposals from MAVL are used to generate the pseudo-labels for unknown categories in Open-world Object Detector (ORE) \cite{joseph2021towards} training. To avoid any data leakage, MAVL is trained on a subset of LMDet dataset, removing all the captions that contain any of the 60 unknown categories in ORE task-1. This filtering leaves us with a dataset having approximately 0.76M (out of 1.3M) image-text pairs. MAVL is trained from scratch on this filtered dataset for 20 epochs and then used to produce unknown pseudo-labels using class-agnostic object proposal generation.

To do so, firstly, proposals with objectness score less than 0.7 are discarded. Secondly, all proposals having an IoU greater than 0.5 with any ground-truth bounding box of a known category are removed. Rest of the proposals potentially belong to unknown categories and are used as pseudo-labels of unknowns in ORE training. All relevant scripts and annotations will be publicly released.

\section{Limitations}\label{app:limitations}
Although MViTs (GPV-1 \cite{gpv1}, MDETR \cite{mdetr} and MAVL) show state-of-the-art class-agnostic OD performance across various dataset domains, they cannot be directly adapted to specialized out-of-domain detection tasks such as in medical imaging.
\begin{wrapfigure}{r}{0.5\textwidth}
  \begin{center}
    \begin{subfigure}{0.24\columnwidth}
    \centering
    \includegraphics[height=2.5cm,width=\linewidth]{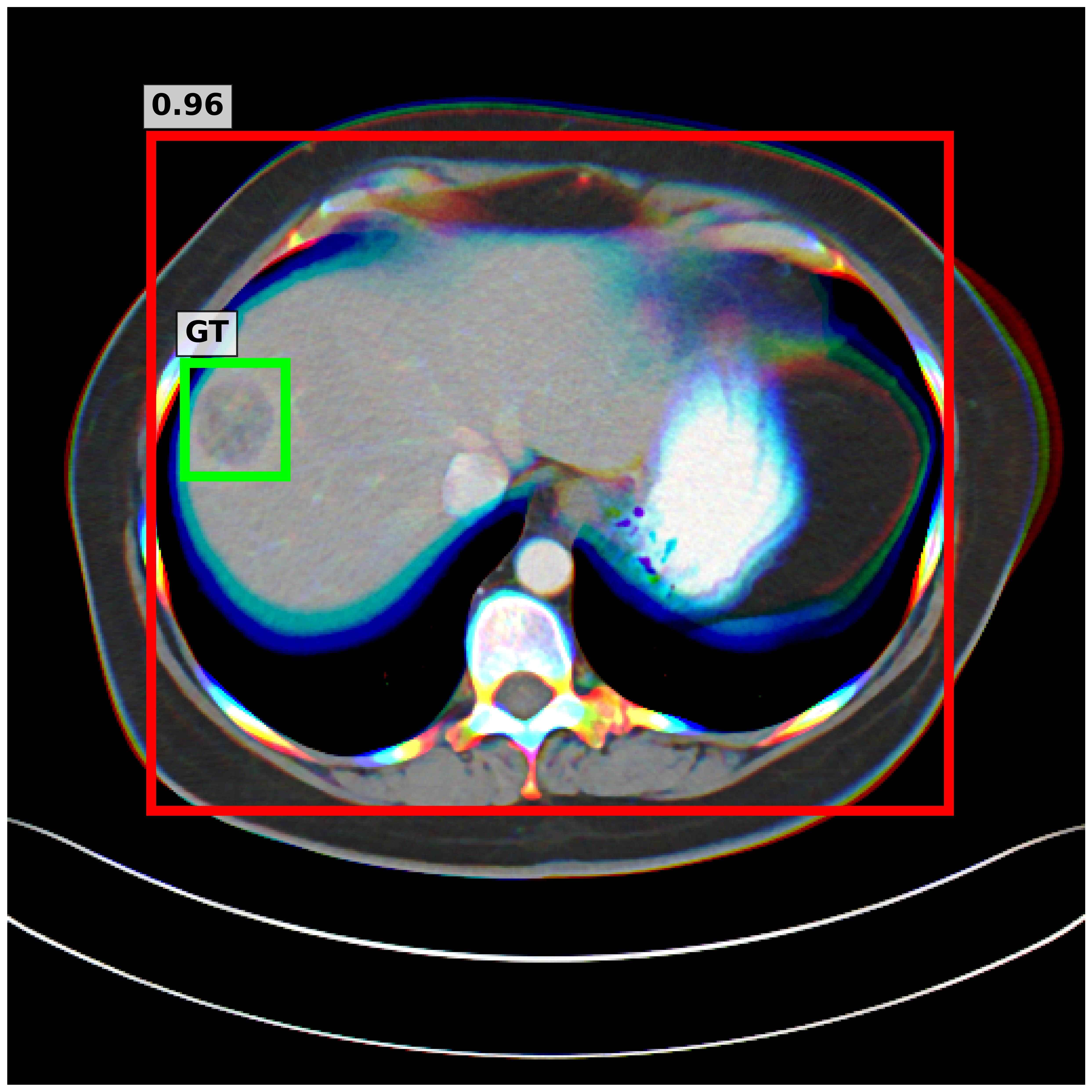}
  \end{subfigure}%
  \begin{subfigure}{0.24\columnwidth}
    \centering
    \includegraphics[height=2.5cm,width=\linewidth]{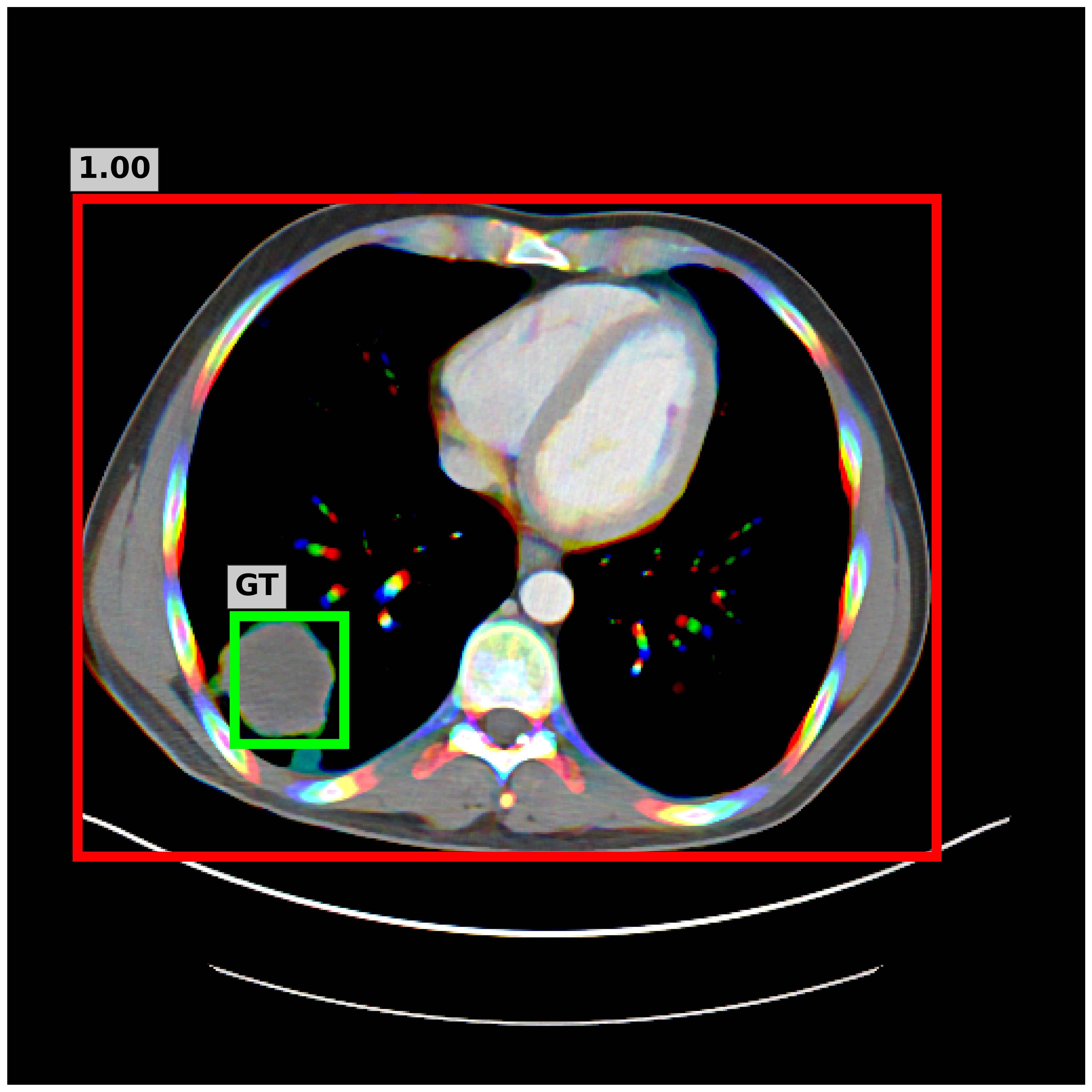}
  \end{subfigure}%
  \end{center}
  \caption{Illustration of MAVL detections on the DeepLesion \cite{yan2018deeplesion} dataset. The green boxes indicate the ground-truth bounding box enclosing the lesion on the CT images and the red boxes are the class-agnostic predictions. The samples indicate a failure case of class-agnostic detection of MViT's on lesion detection.}
  \label{fig: deeplesion}
\end{wrapfigure}
We evaluate the class-agnostic OD performance of MAVL on DeepLesion \cite{yan2018deeplesion} dataset (Fig.~\ref{fig: deeplesion}). The ground-truth annotations represented by the green boxes in Fig.~\ref{fig: deeplesion}, indicate that the target lesions do not well represent the concept of an object, and require expert based supervision to identify the abnormalities. In medical domain, lesion detection task involves locating the congenital malformations in different types of medical images including X-rays, CT scans, MRI scans and Ultrasoud. These applications require specialized data along with expert supervision (obtained from well-trained domain specialists) to perform well. Hence, in most cases, the general class-agnostic OD methods (\eg MViTs) cannot be direclty used. We observe that the generic class-agnostic detection mechanism of MViTs trained on out-of-domain natural images is not well-suited for generating proposals that can cater the need of specific medical applications.

\section{Qualitative Results}\label{app:qual_results}
\begin{figure*}[!ht]
  \centering
  \begin{subfigure}{.5\columnwidth}
    \centering
    \includegraphics[height=2.8cm,width=0.99\linewidth]{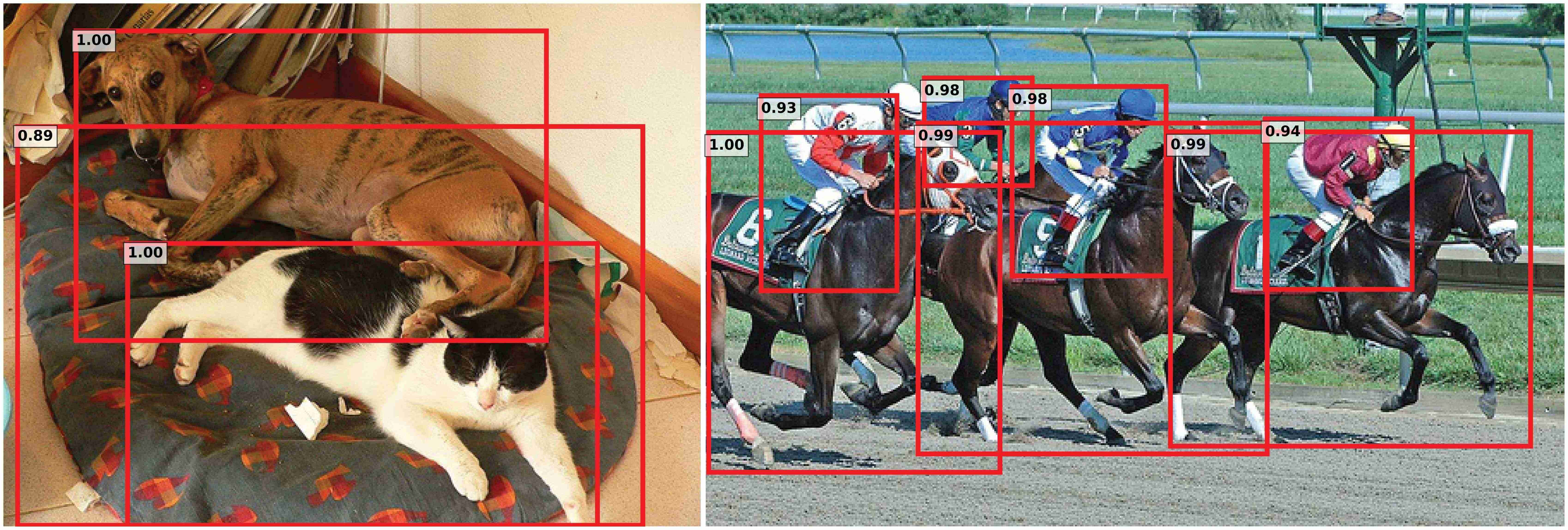}
    \caption{Pascal VOC \cite{voc}}
    \label{a}
  \end{subfigure}%
  \begin{subfigure}{.5\columnwidth}
    \centering
    \includegraphics[height=2.8cm,width=0.99\linewidth]{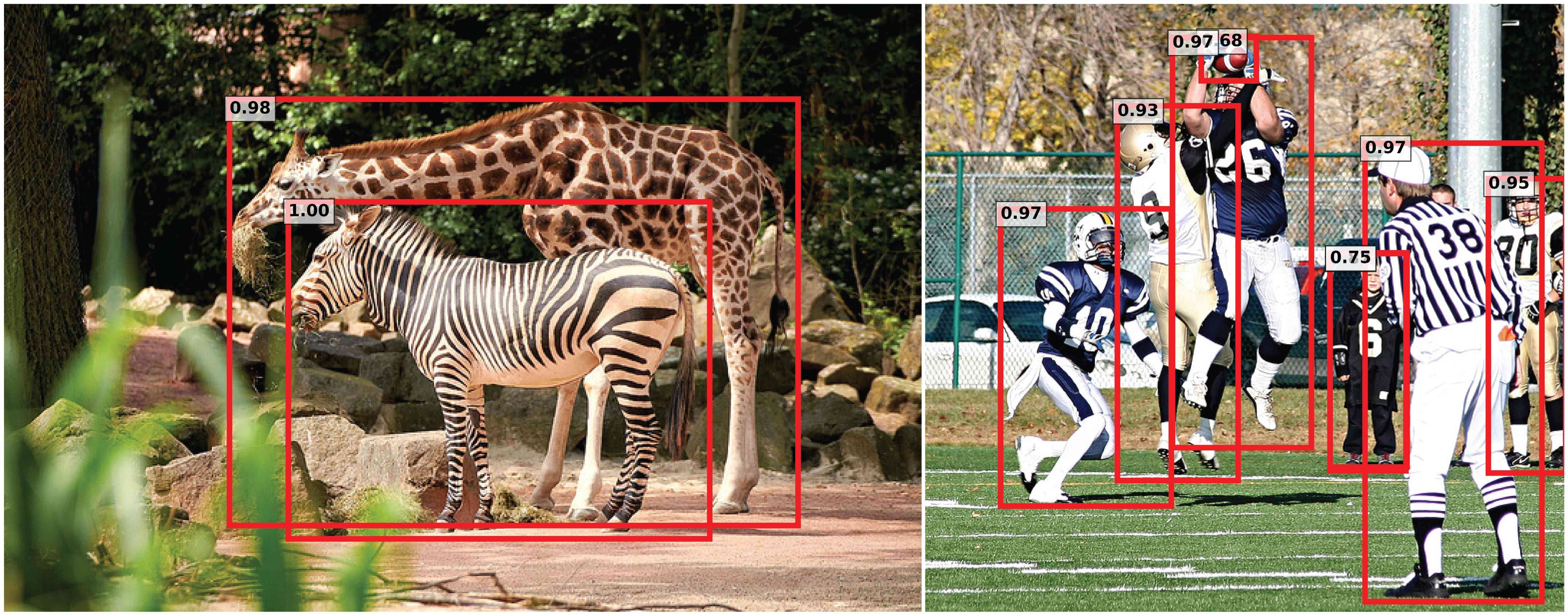}
    \caption{MS COCO \cite{coco} / LVIS \cite{gupta2019lvis}}
    \label{b}
  \end{subfigure}%
  \hfill
  \begin{subfigure}{\columnwidth}
    \centering
    \includegraphics[height=2.5cm,width=0.99\linewidth]{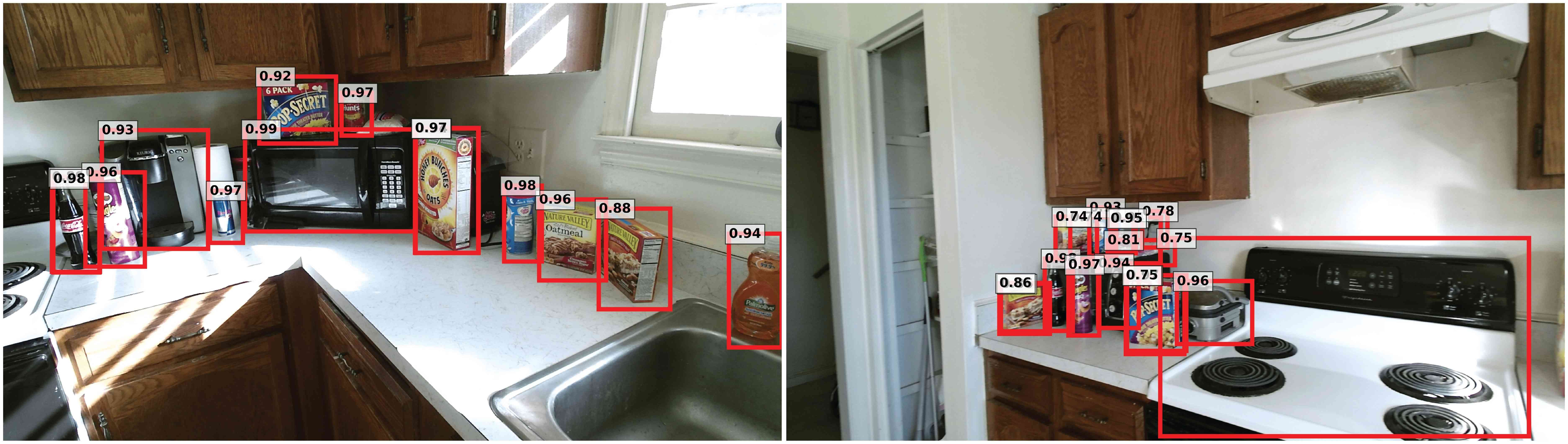}
    \caption{Kitchen \cite{kitchen}}
    \label{g}
  \end{subfigure}%
  \hfill
  \begin{subfigure}{\columnwidth}
    \centering
    \includegraphics[height=2cm,width=0.99\linewidth]{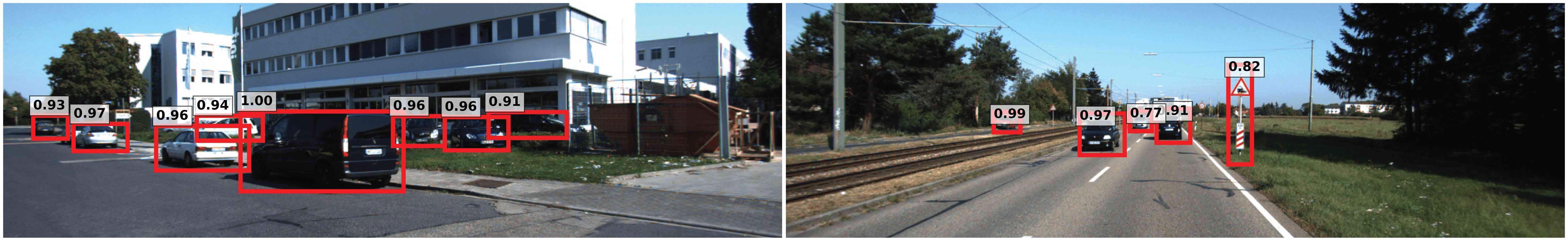}
    \caption{KITTI \cite{kitti}}
    \label{h}
  \end{subfigure}
  \caption{Class-agnostic detections of MViTs (MDETR \cite{mdetr} and MAVL) on natural image datasets, Pascal VOC, MS COCO/LVIS, Kitchen and KITTI.}
  \label{figure:class_agnostic1}
\end{figure*}

\begin{figure*}[!ht]
  \centering
  \begin{subfigure}{.5\columnwidth}
    \centering
    \includegraphics[height=3cm,width=0.99\linewidth]{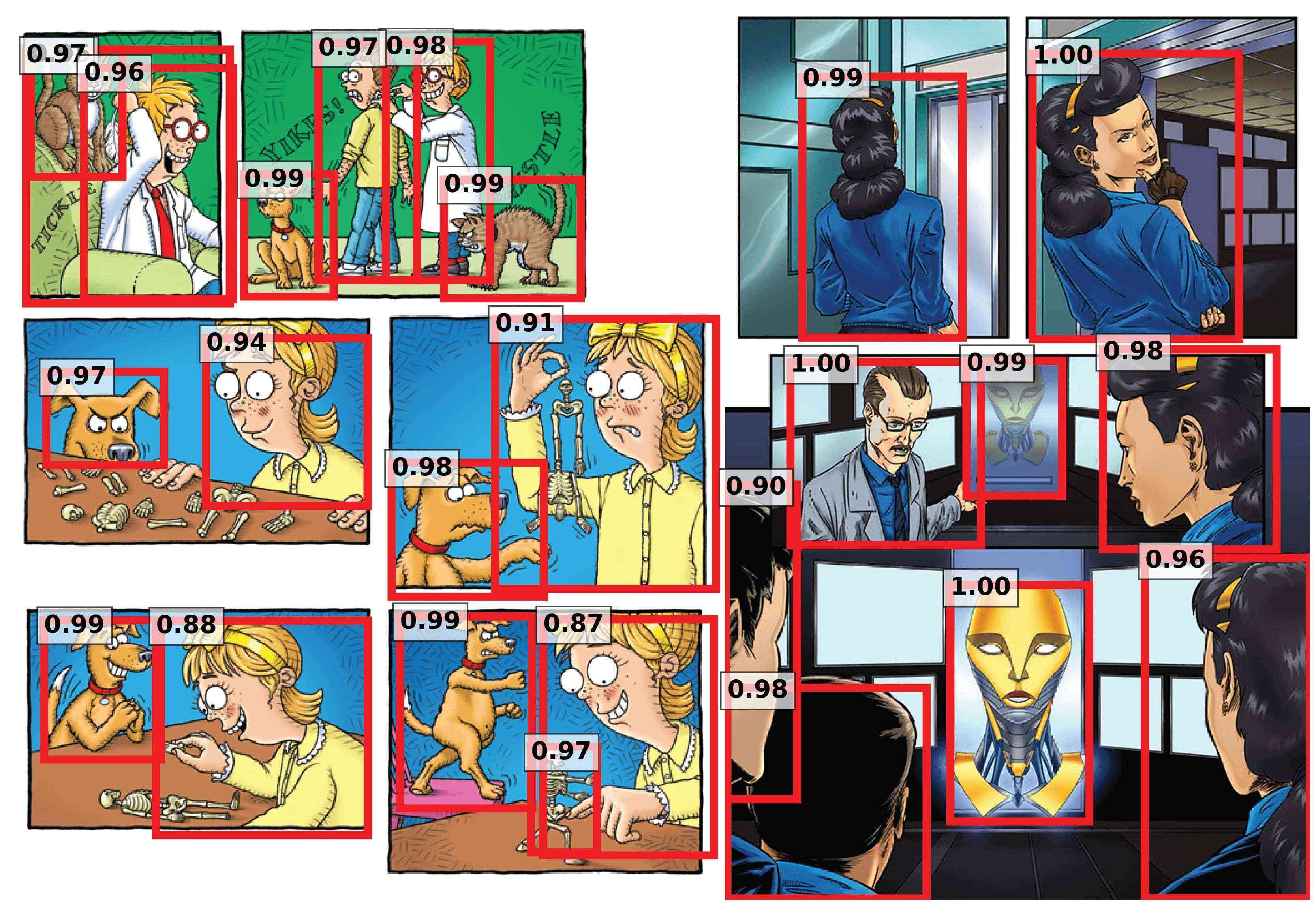}
    \caption{Comic \cite{clipart-comic-water}}
    \label{c}
  \end{subfigure}%
  \begin{subfigure}{.5\columnwidth}
    \centering
    \includegraphics[height=3cm,width=0.99\linewidth]{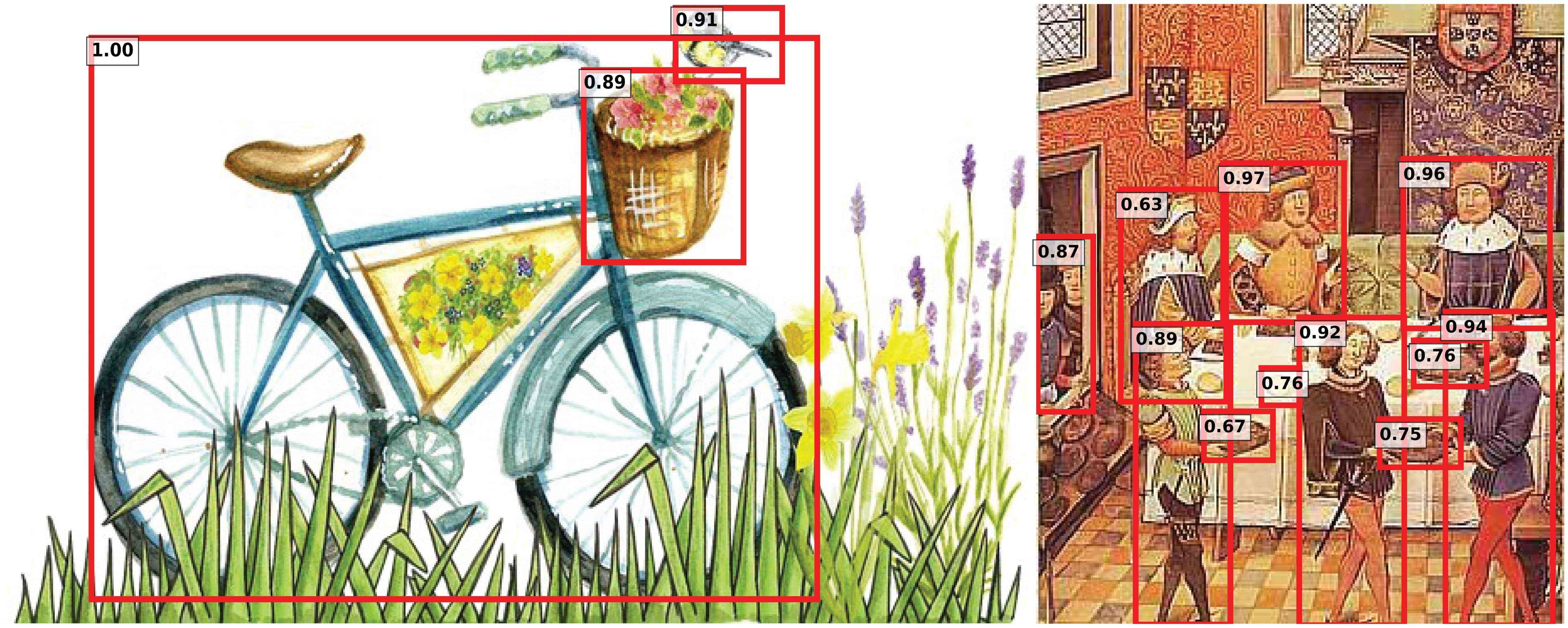}
    \caption{Clipart \cite{clipart-comic-water}}
    \label{d}
  \end{subfigure}%
  \hfill
 \begin{subfigure}{.5\columnwidth}
    \centering
    \includegraphics[height=2.8cm,width=0.99\linewidth]{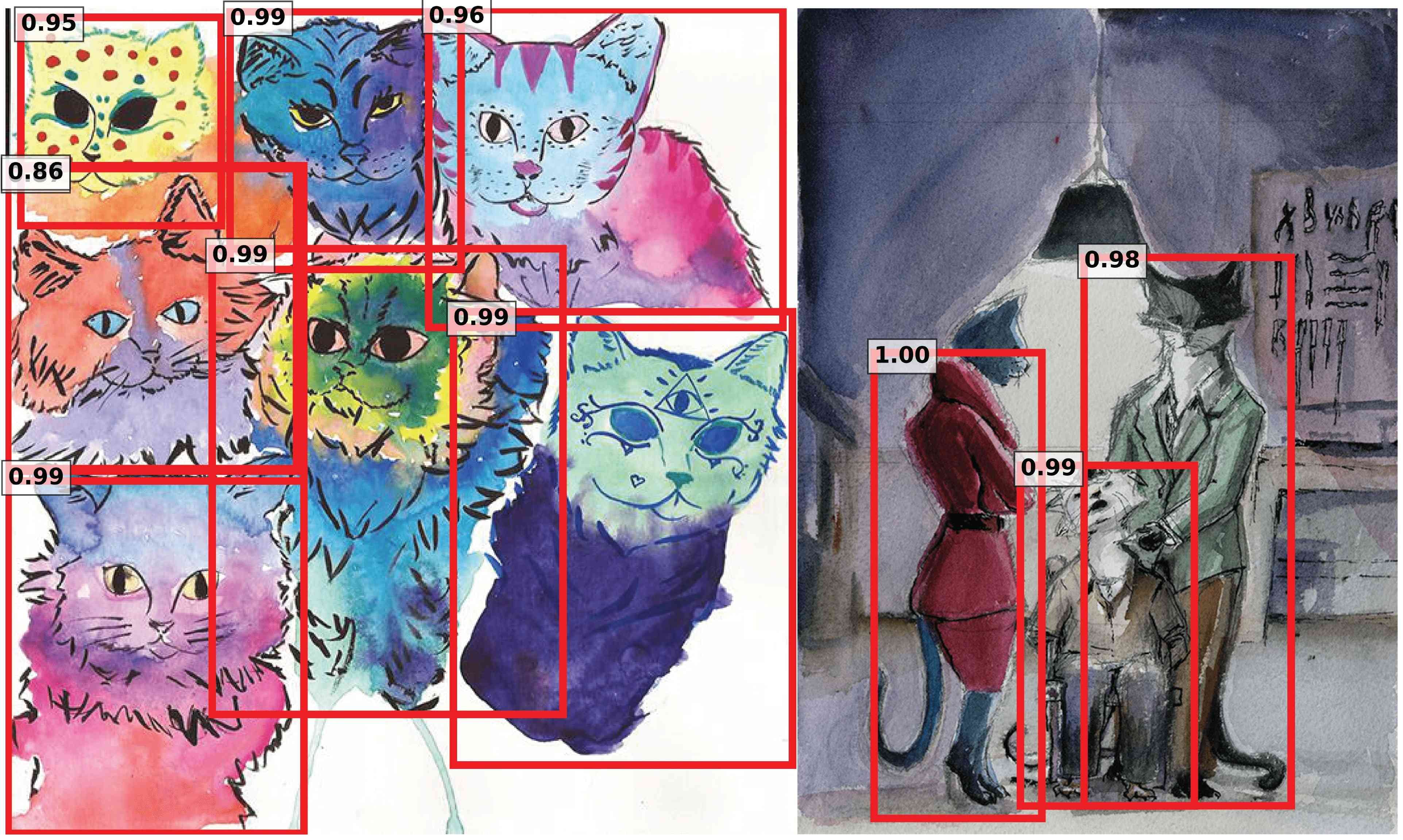}
    \caption{Watercolor \cite{clipart-comic-water}}
    \label{e}
  \end{subfigure}%
  \begin{subfigure}{.5\columnwidth}
    \centering
    \includegraphics[height=2.8cm,width=0.99\linewidth]{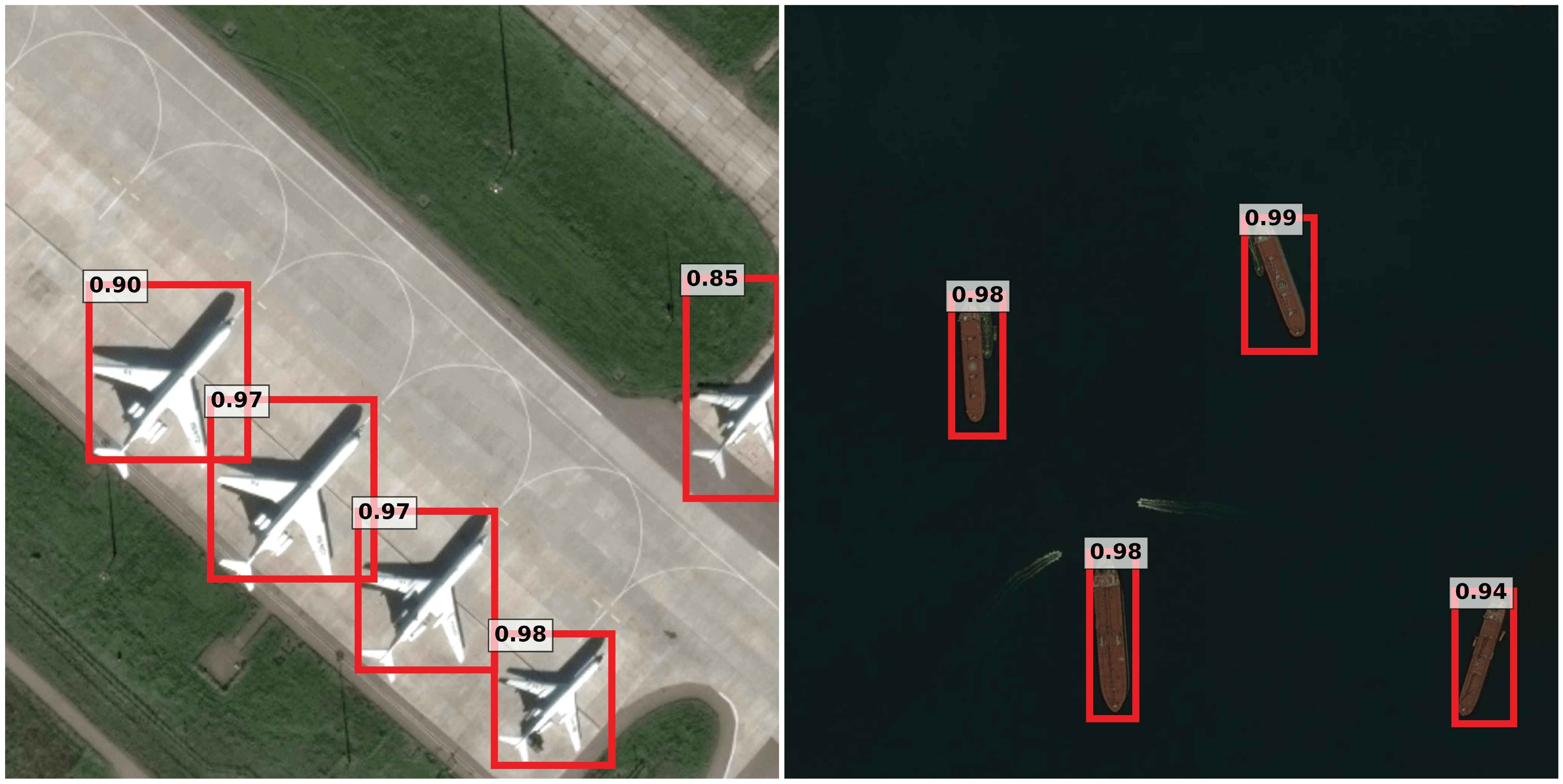}
    \caption{DOTA \cite{dota}}
    \label{f}
  \end{subfigure}%
  \hfill
  \caption{Class-agnostic detections of MViTs (MDETR \cite{mdetr} and MAVL) on out-of-domian datasets, Comic, Clipart, Watercolor and DOTA.}
  \label{figure:class_agnostic2}
\end{figure*}

\begin{figure*}[!h]
  \centering
    \begin{subfigure}{.245\columnwidth}
    \centering
    \includegraphics[height=2.5cm,width=\linewidth]{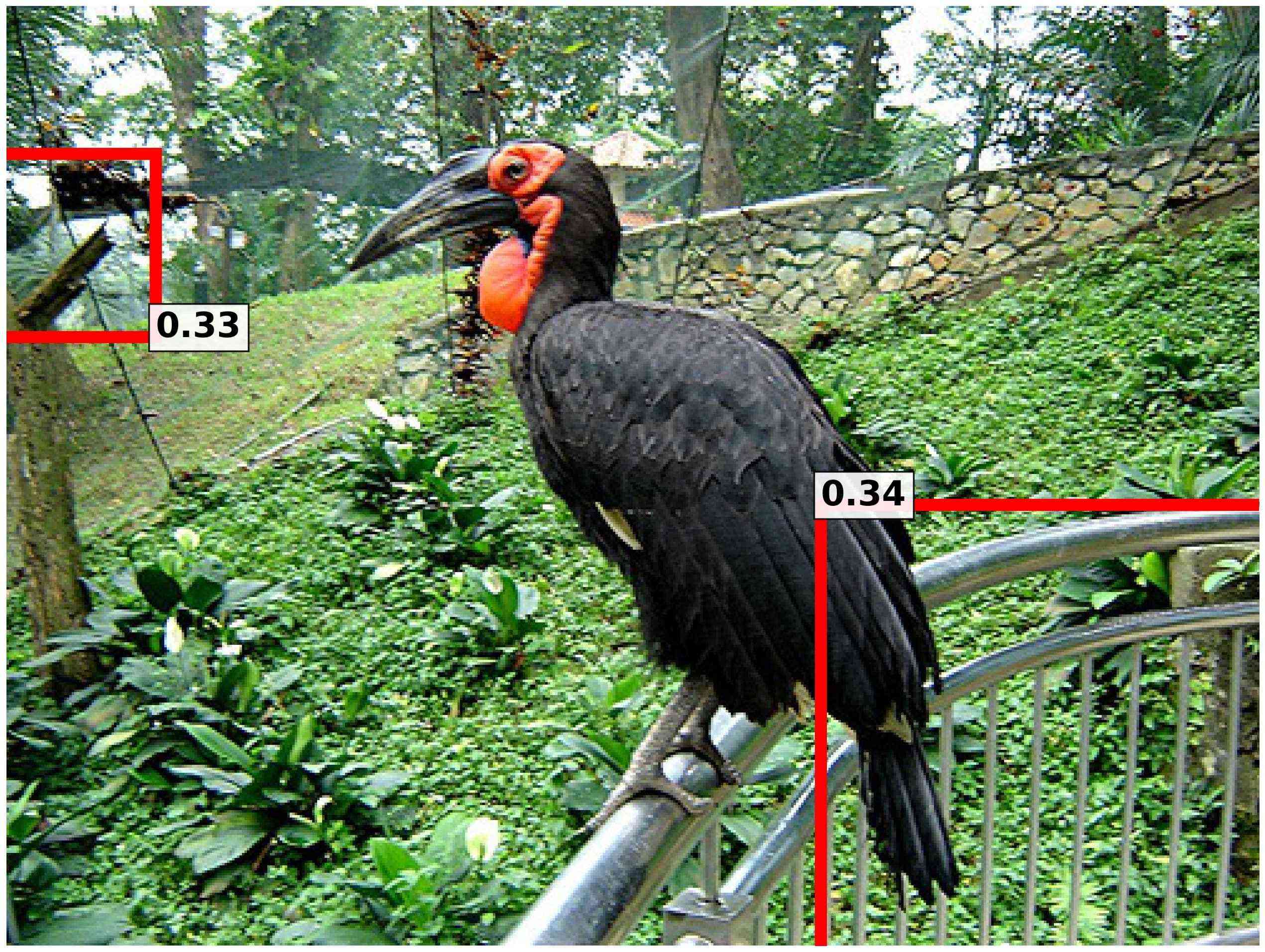}
  \end{subfigure}%
  \begin{subfigure}{.245\columnwidth}
    \centering
    \includegraphics[height=2.5cm,width=\linewidth]{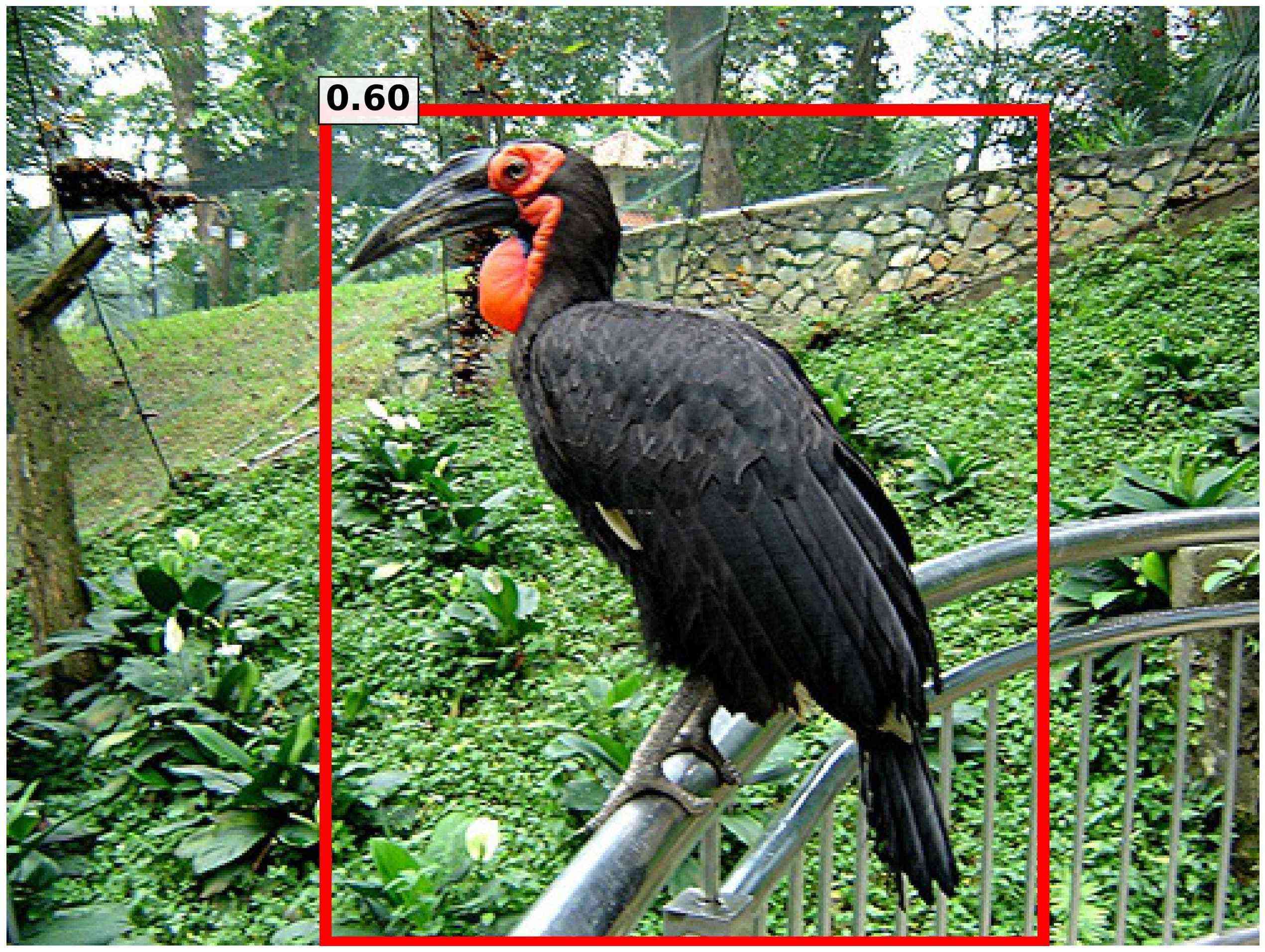}
  \end{subfigure}
  \begin{subfigure}{.245\columnwidth}
    \centering
    \includegraphics[height=2.5cm,width=\linewidth]{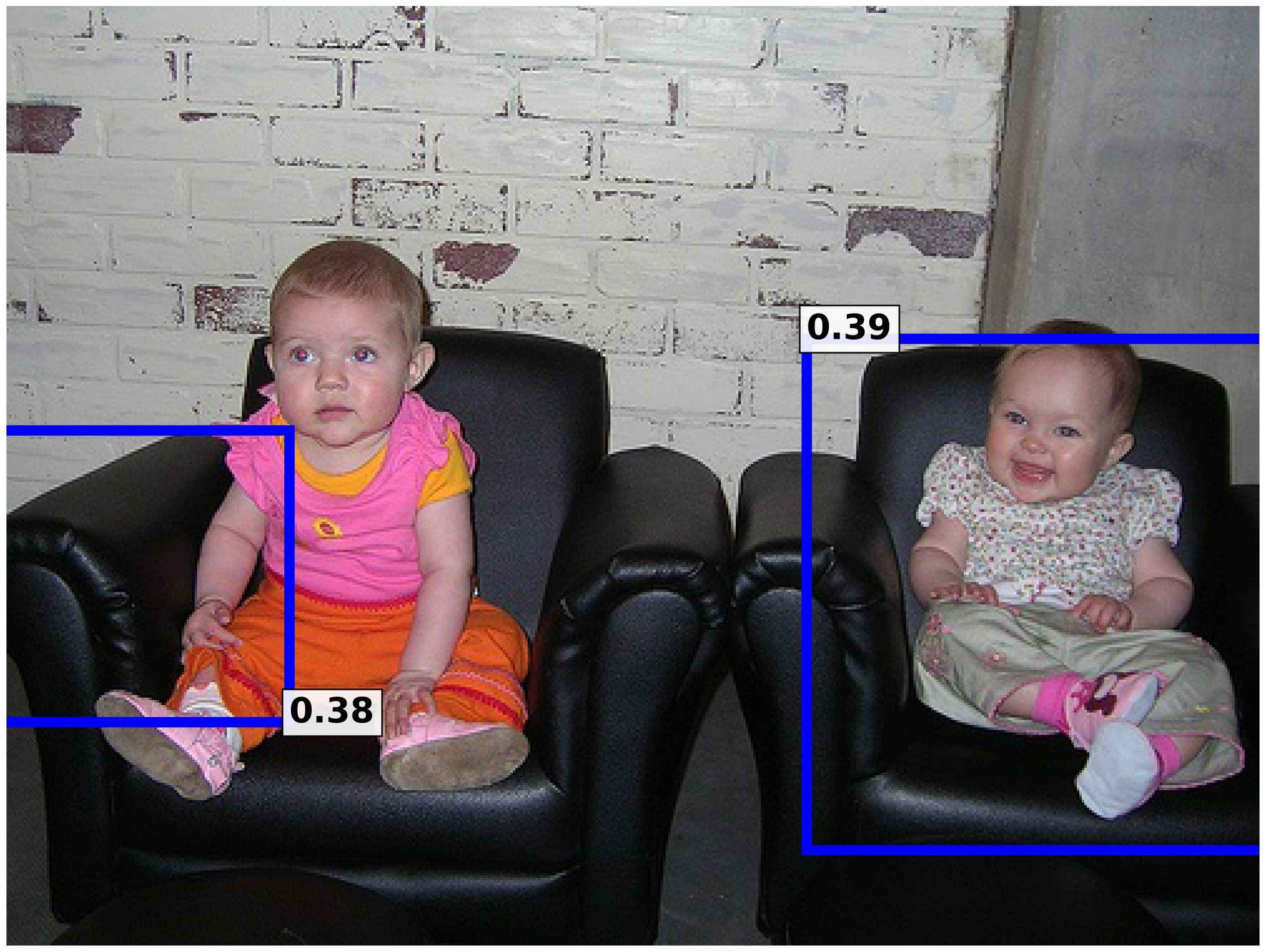}
  \end{subfigure}%
  \begin{subfigure}{.245\columnwidth}
    \centering
    \includegraphics[height=2.5cm,width=\linewidth]{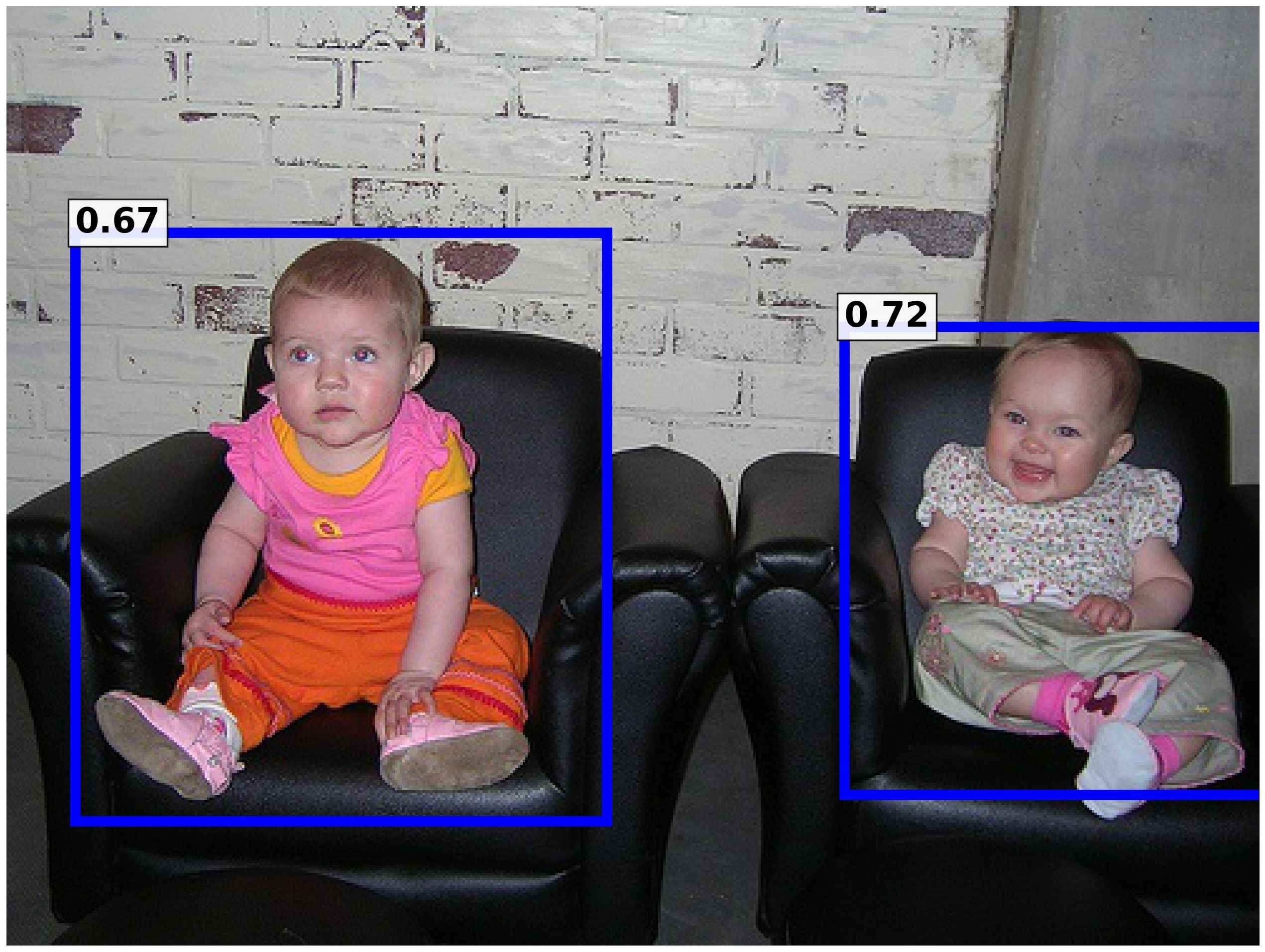}
  \end{subfigure}%
  \\
  \begin{subfigure}{.245\columnwidth}
    \centering
    \includegraphics[height=2.5cm,width=\linewidth]{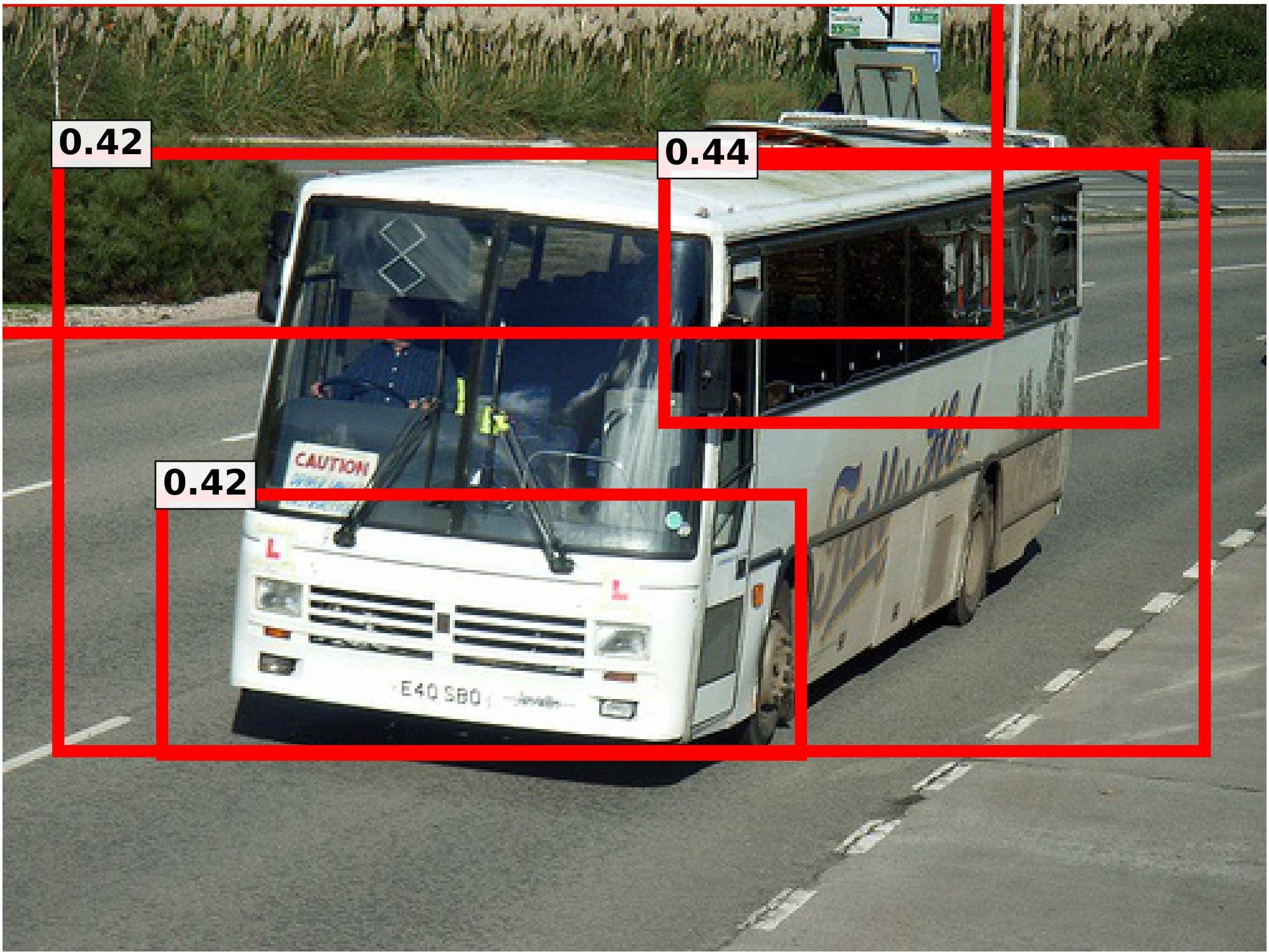}
  \end{subfigure}%
  \begin{subfigure}{.245\columnwidth}
    \centering
    \includegraphics[height=2.5cm,width=\linewidth]{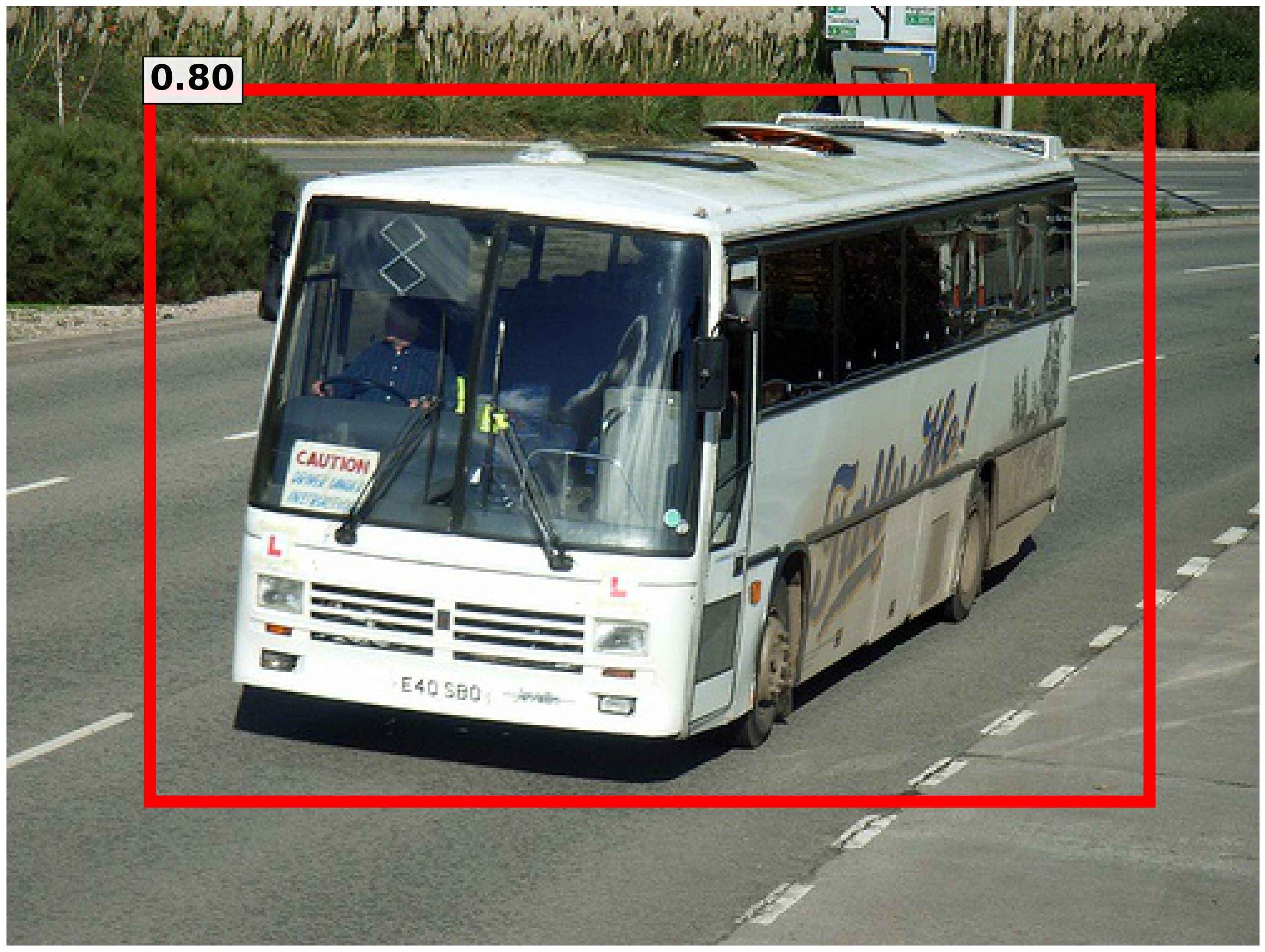}
  \end{subfigure}%
  \begin{subfigure}{.245\columnwidth}
    \centering
    \includegraphics[height=2.5cm,width=\linewidth]{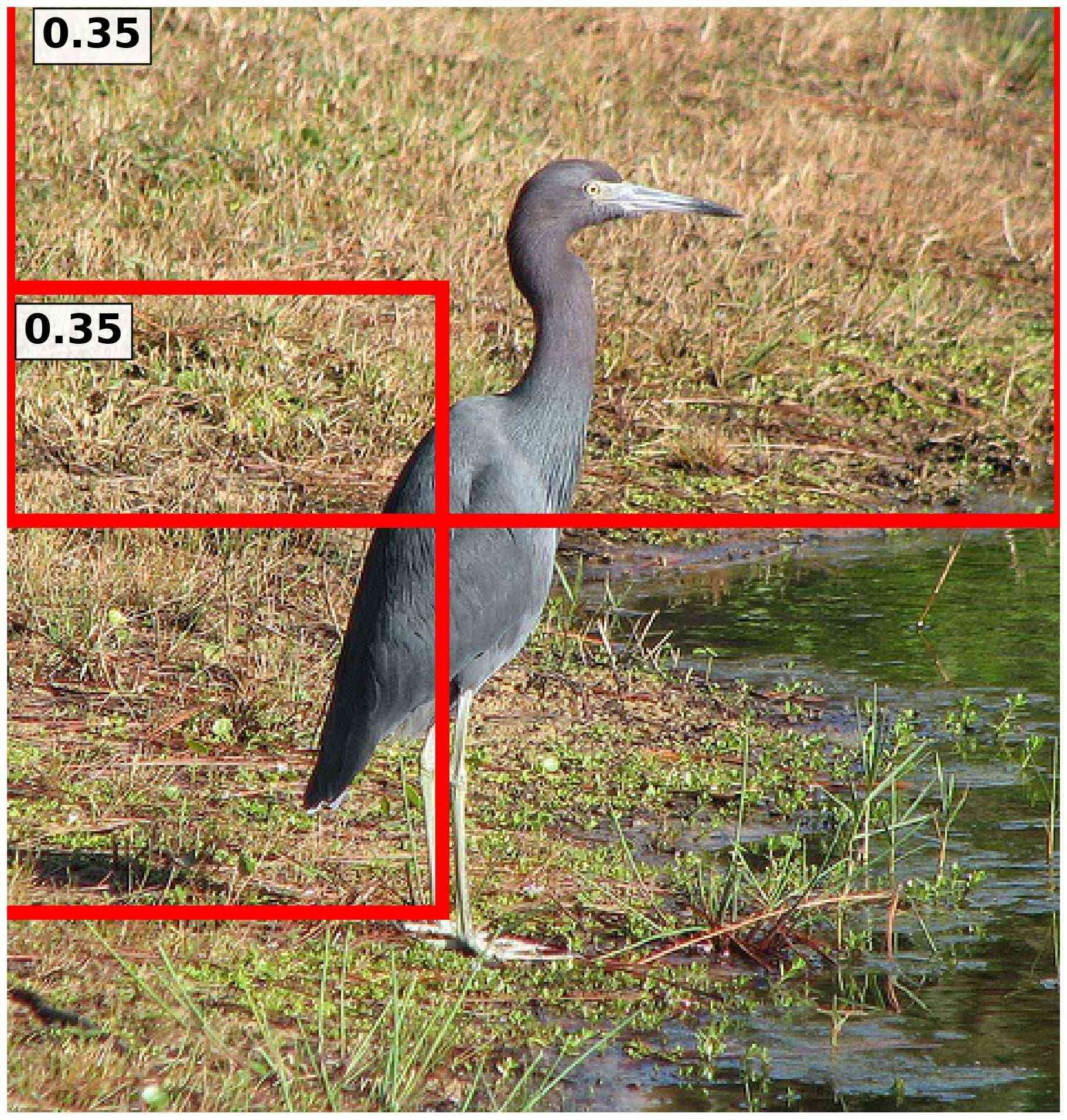}
  \end{subfigure}%
  \begin{subfigure}{.245\columnwidth}
    \centering
    \includegraphics[height=2.5cm,width=\linewidth]{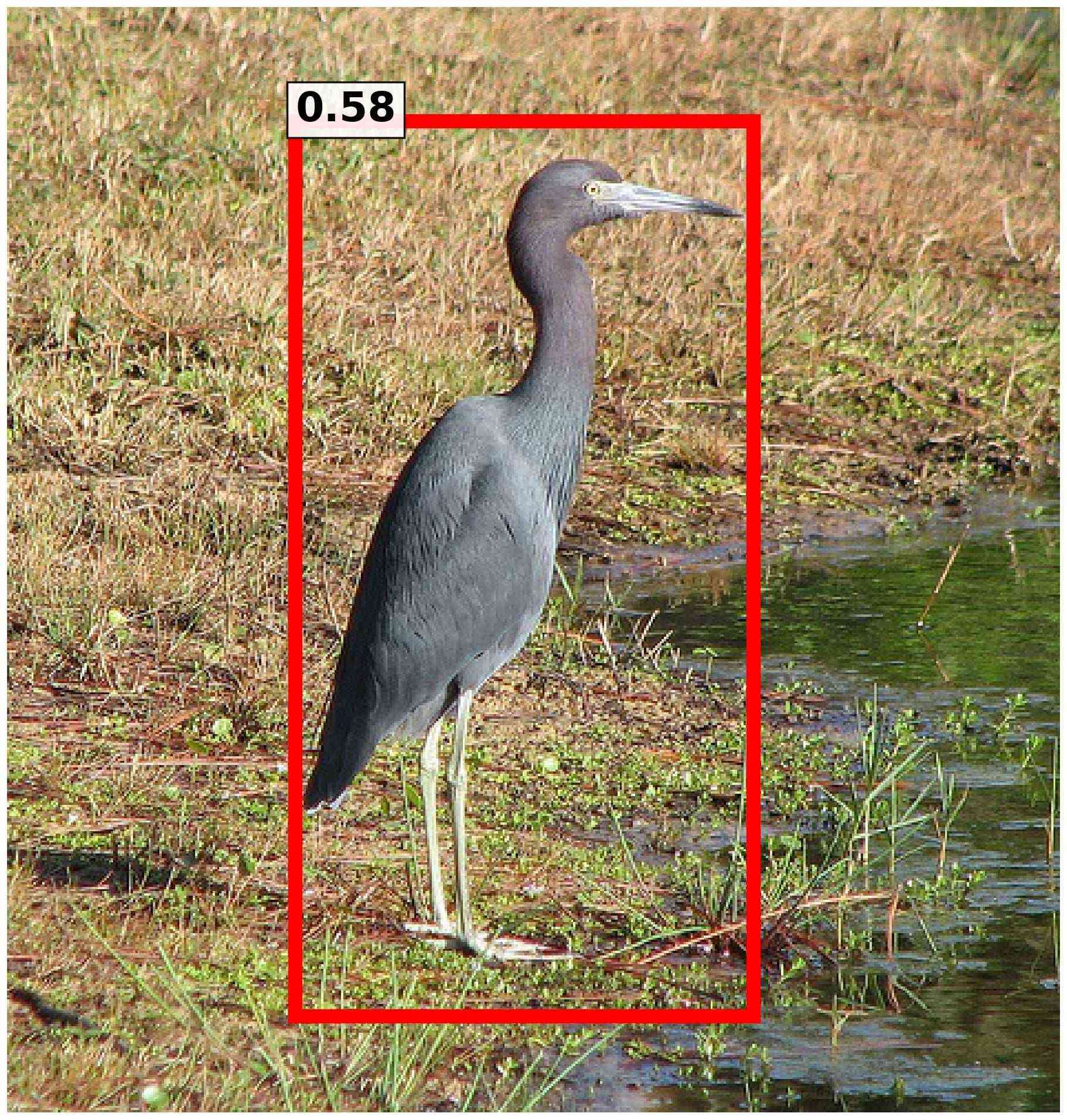}
  \end{subfigure}%
  \caption{Class-agnostic OD performance of DETReg \cite{detreg} trained using Selective Search \cite{uijlings2013selective} versus MAVL proposals. The images on the left side of each example correspond to DETReg trained with Selective search and the images on the right side correspond to the one trained with MAVL that results in better localized predictions}
  \label{figure:detreg}
\end{figure*}

We present examples of class-agnostic predictions of MDETR and MAVL across natural image dataset Pascal VOC \cite{voc}, COCO/LVIS \cite{coco,gupta2019lvis}, autonomous driving dataset KITTI \cite{kitti} and indoor Kitchen dataset \cite{kitchen} in Fig.~\ref{figure:class_agnostic1} and out-of-domain datasets that include sketches, painting, cartoons \cite{clipart-comic-water} and satellite images \cite{dota} in Fig.~\ref{figure:class_agnostic2}. The detections are generated using the natural language text query, ‘\txt{all objects}'. In Fig.~\ref{figure:detreg}, we present some qualitative examples of class-agnostic OD with DETReg \cite{detreg} trained using off-the-shelf proposals from Selective Search \cite{uijlings2013selective} in comparison with DETReg trained using MAVL proposals.

\begin{figure*}[!h]
  \centering
    \begin{subfigure}{.24\columnwidth}
    \centering
    \includegraphics[height=3cm,width=\linewidth]{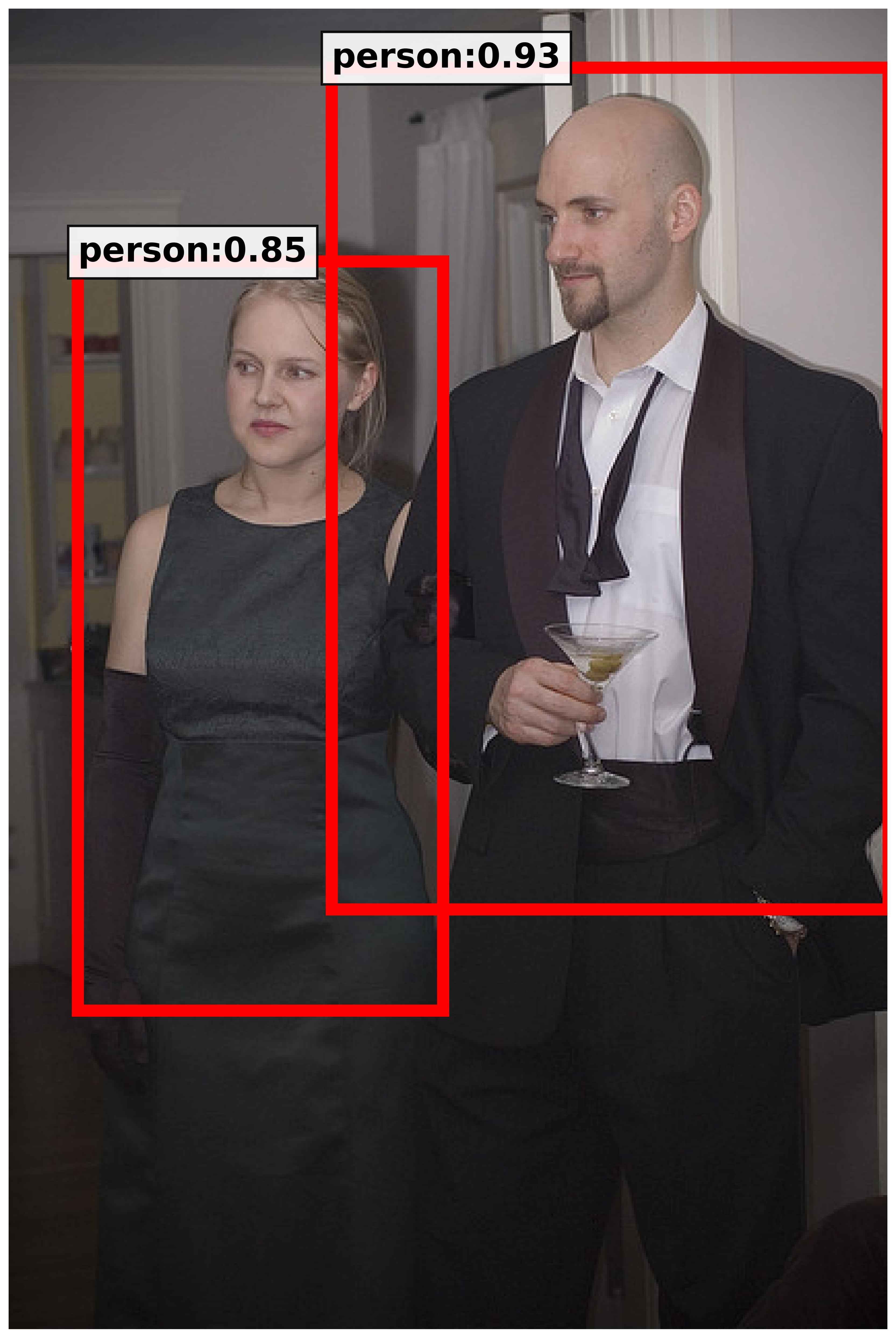}
  \end{subfigure}%
  \begin{subfigure}{.24\columnwidth}
    \centering
    \includegraphics[height=3cm,width=\linewidth]{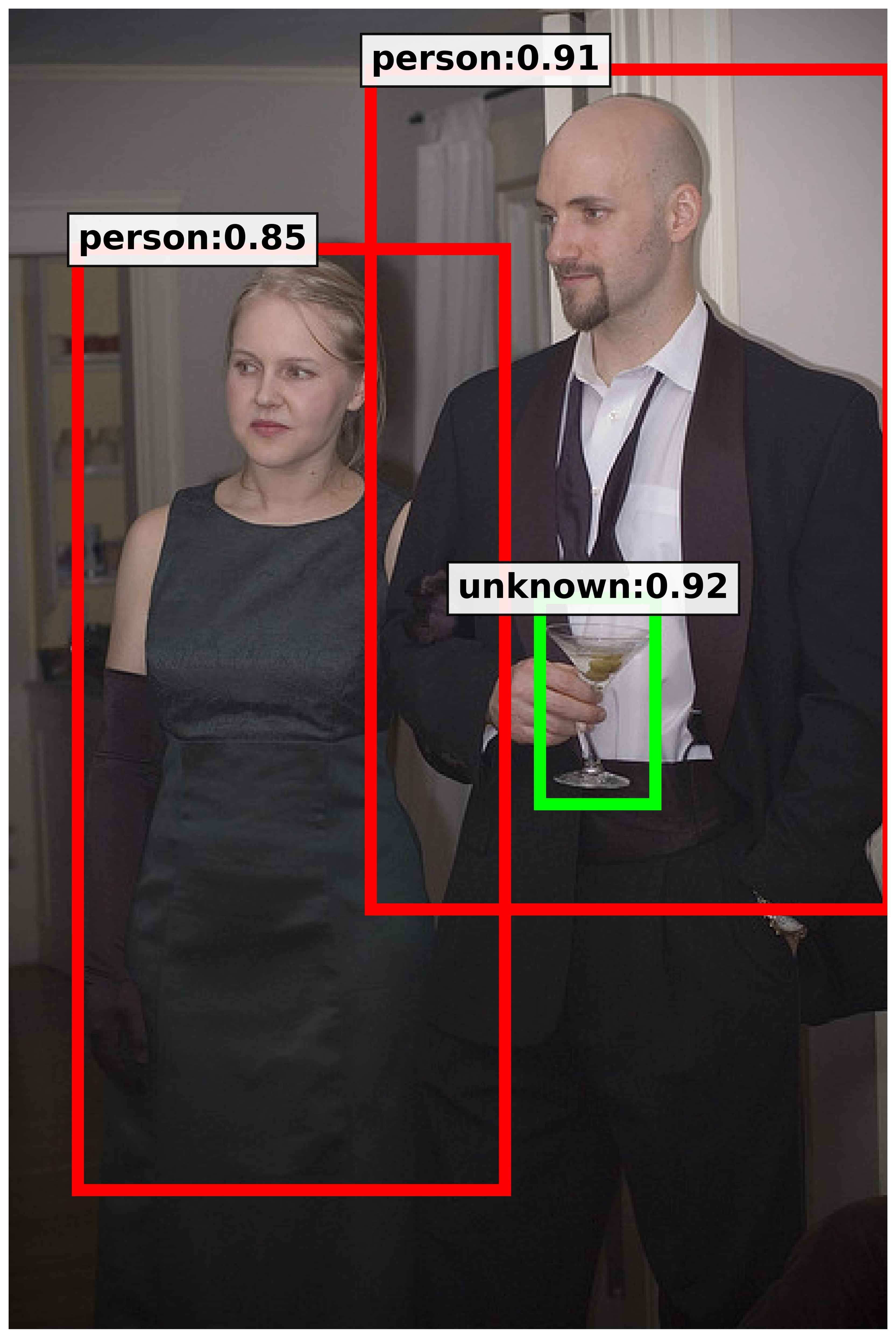}
  \end{subfigure}
  \begin{subfigure}{.24\columnwidth}
    \centering
    \includegraphics[height=3cm,width=\linewidth]{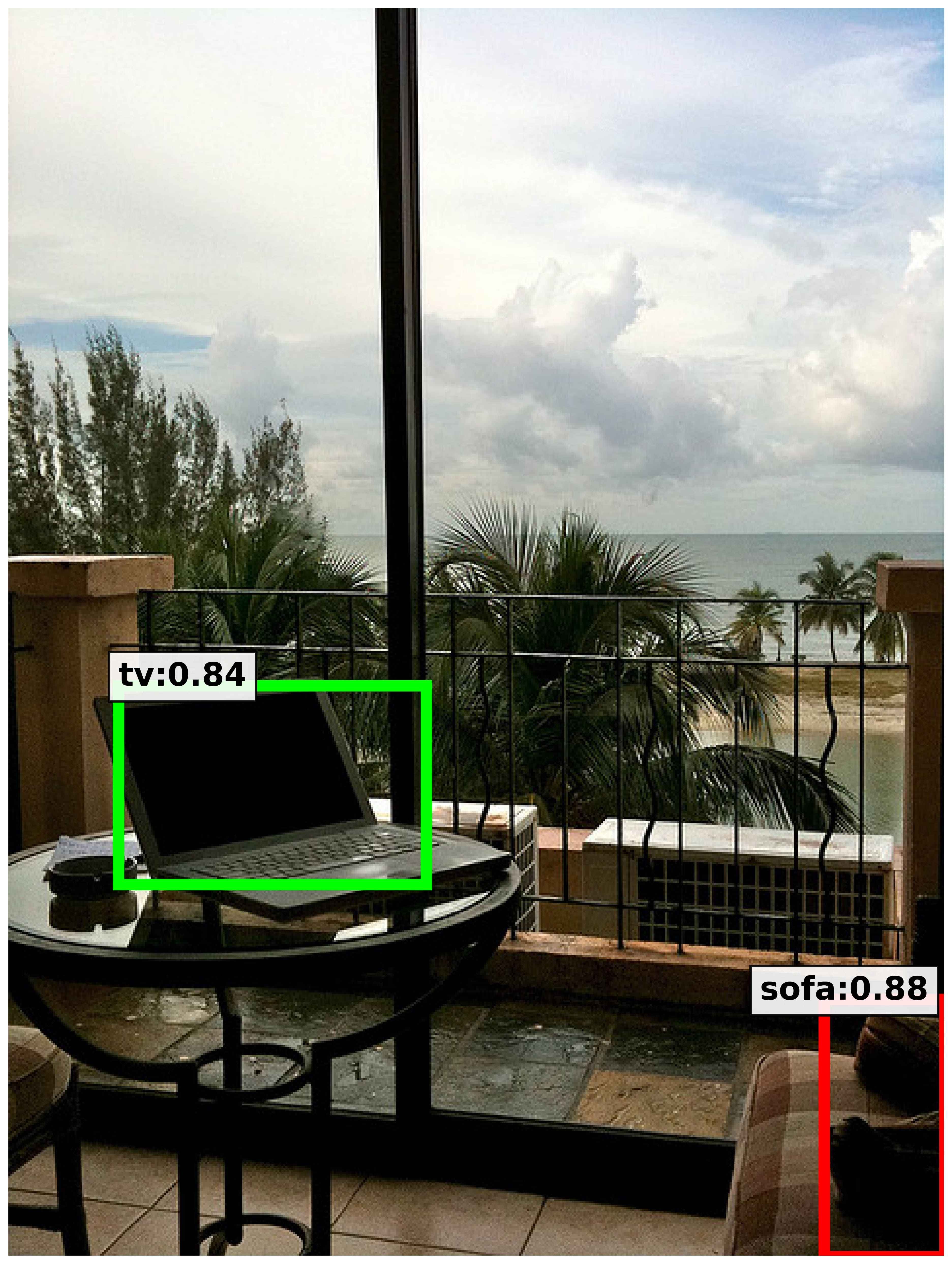}
  \end{subfigure}%
  \begin{subfigure}{.24\columnwidth}
    \centering
    \includegraphics[height=3cm,width=\linewidth]{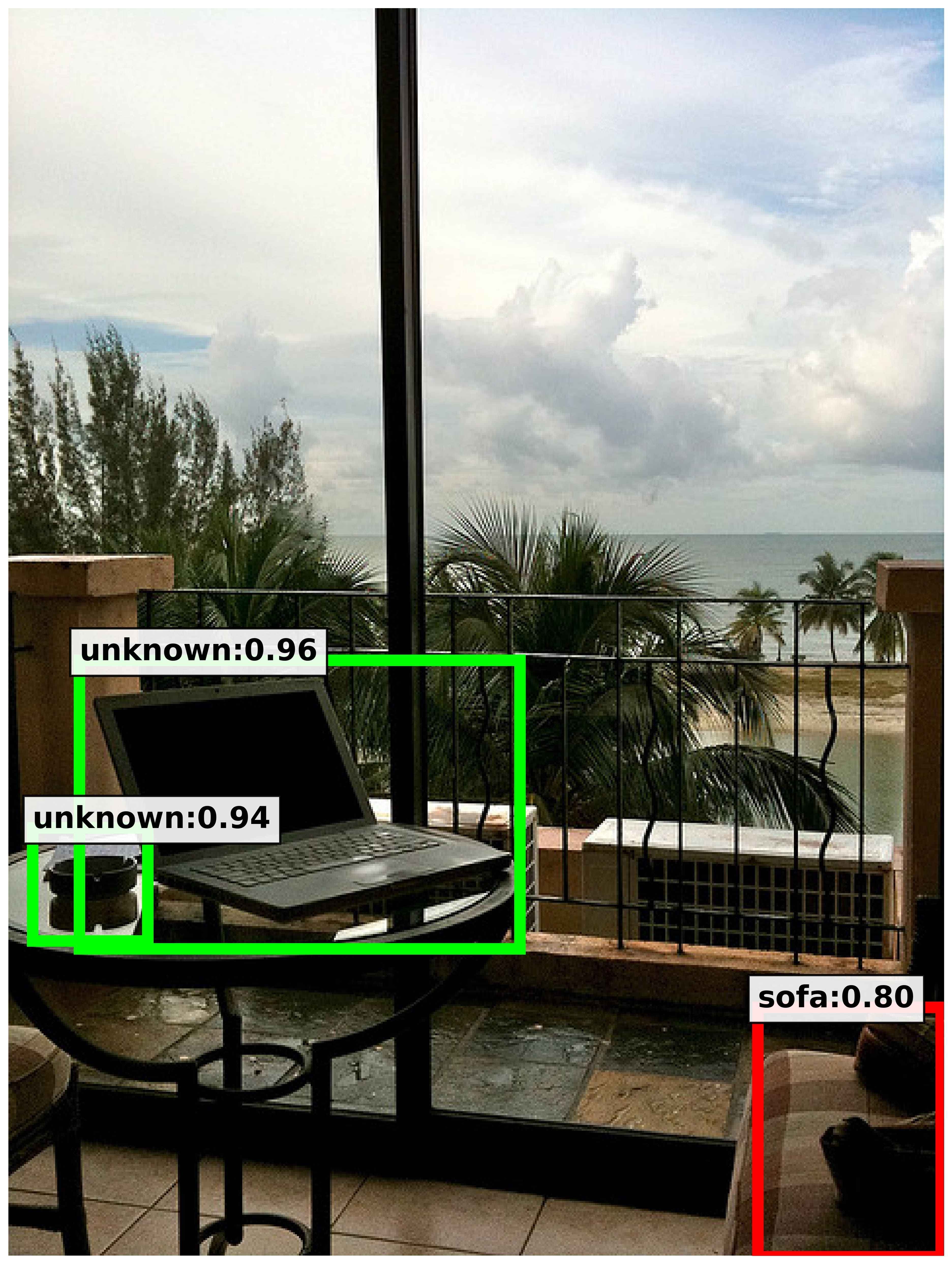}
  \end{subfigure}%
  \hfill
  \begin{subfigure}{.24\columnwidth}
    \centering
    \includegraphics[height=2.5cm,width=\linewidth]{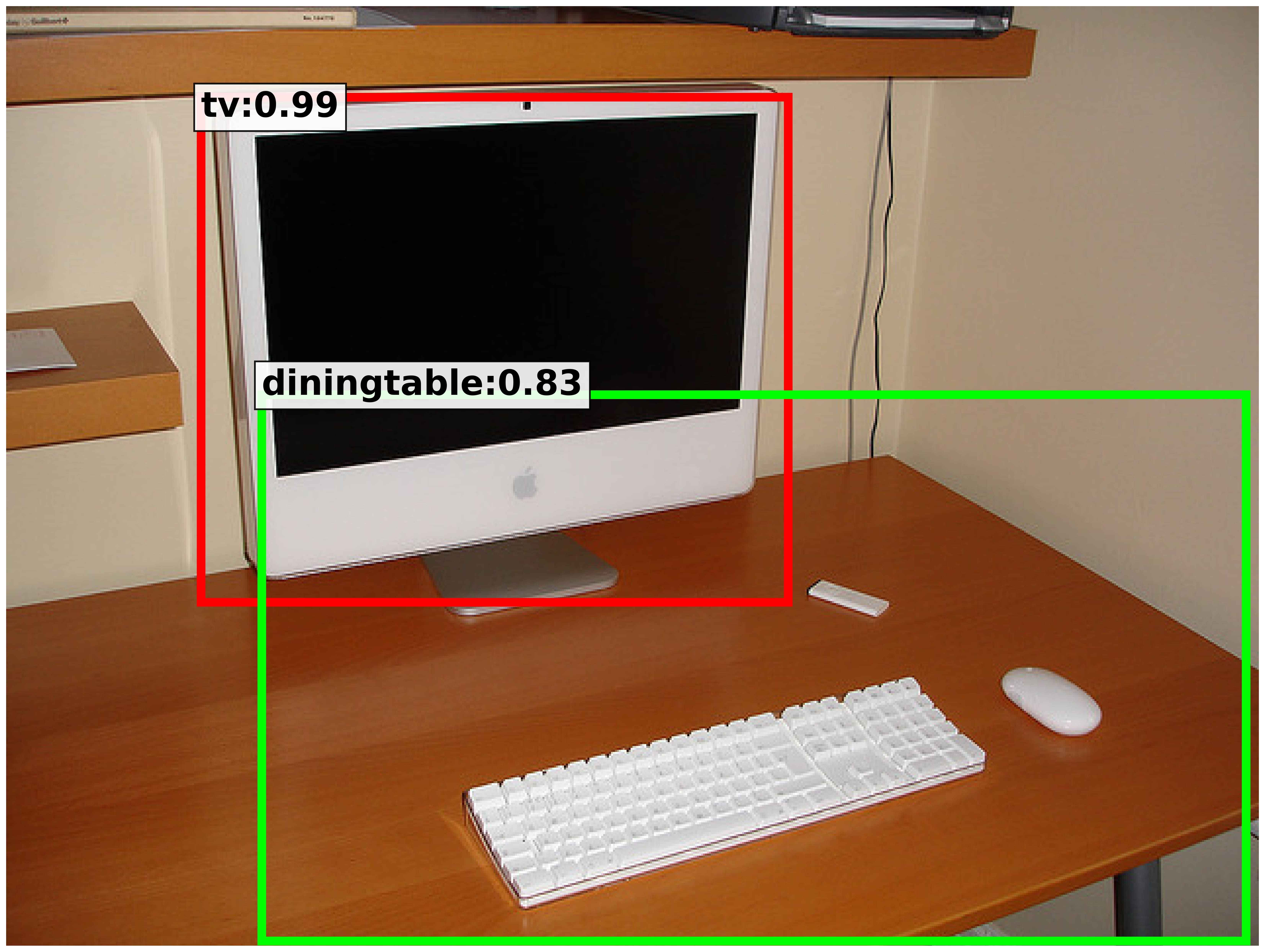}
  \end{subfigure}%
  \begin{subfigure}{.24\columnwidth}
    \centering
    \includegraphics[height=2.5cm,width=\linewidth]{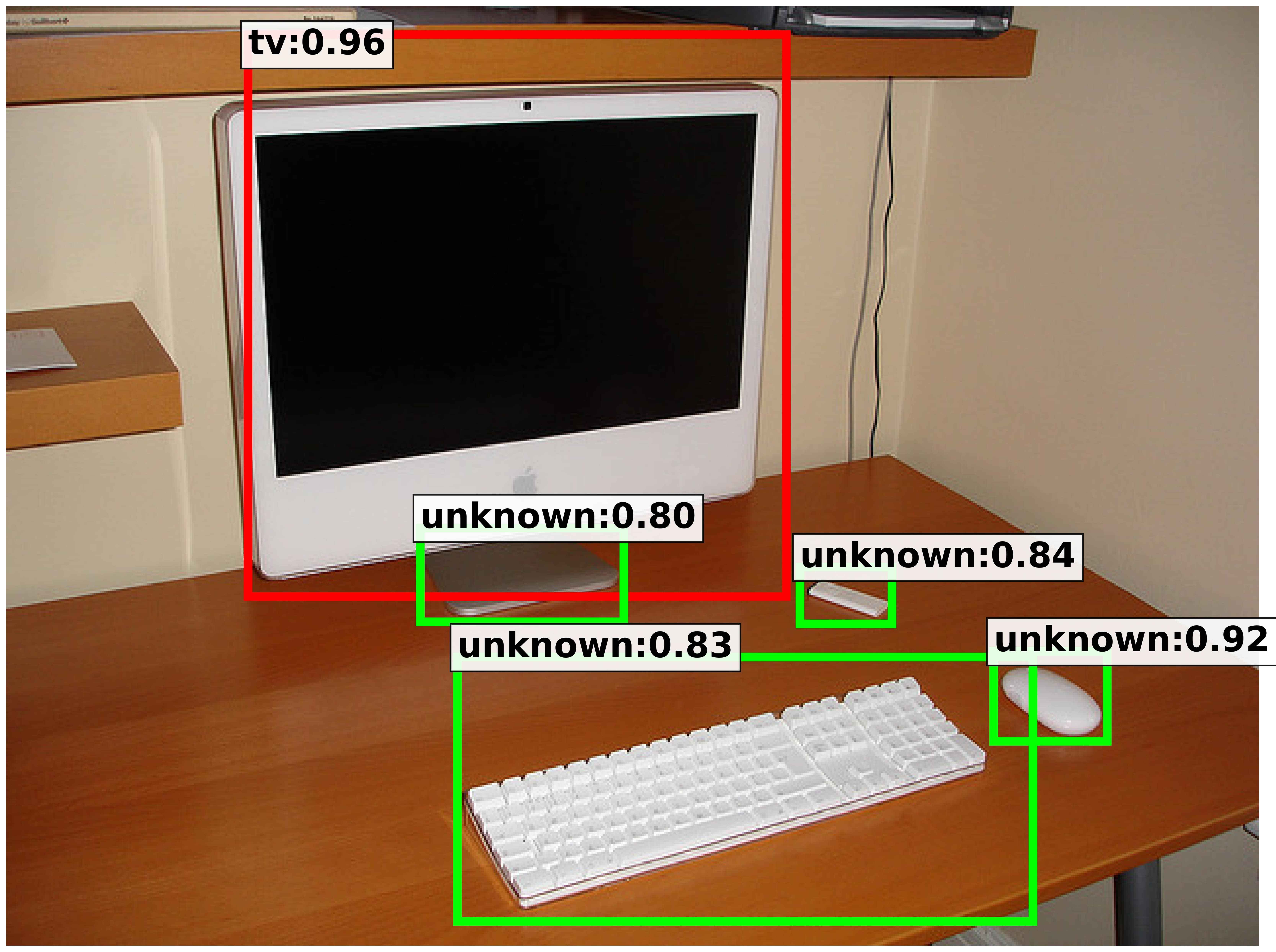}
  \end{subfigure}%
  \begin{subfigure}{.24\columnwidth}
    \centering
    \includegraphics[height=2.5cm,width=\linewidth]{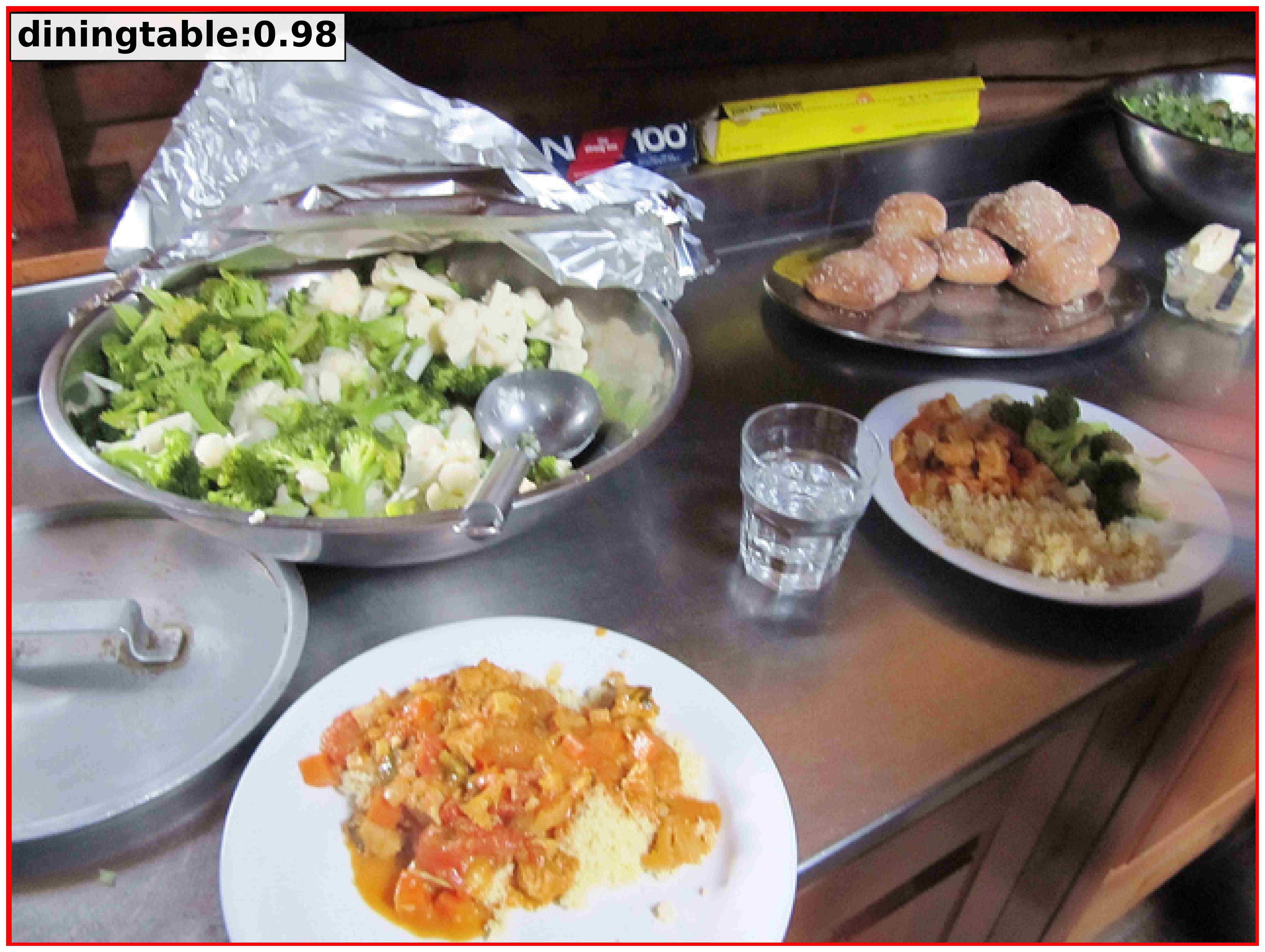}
  \end{subfigure}%
  \begin{subfigure}{.24\columnwidth}
    \centering
    \includegraphics[height=2.5cm,width=\linewidth]{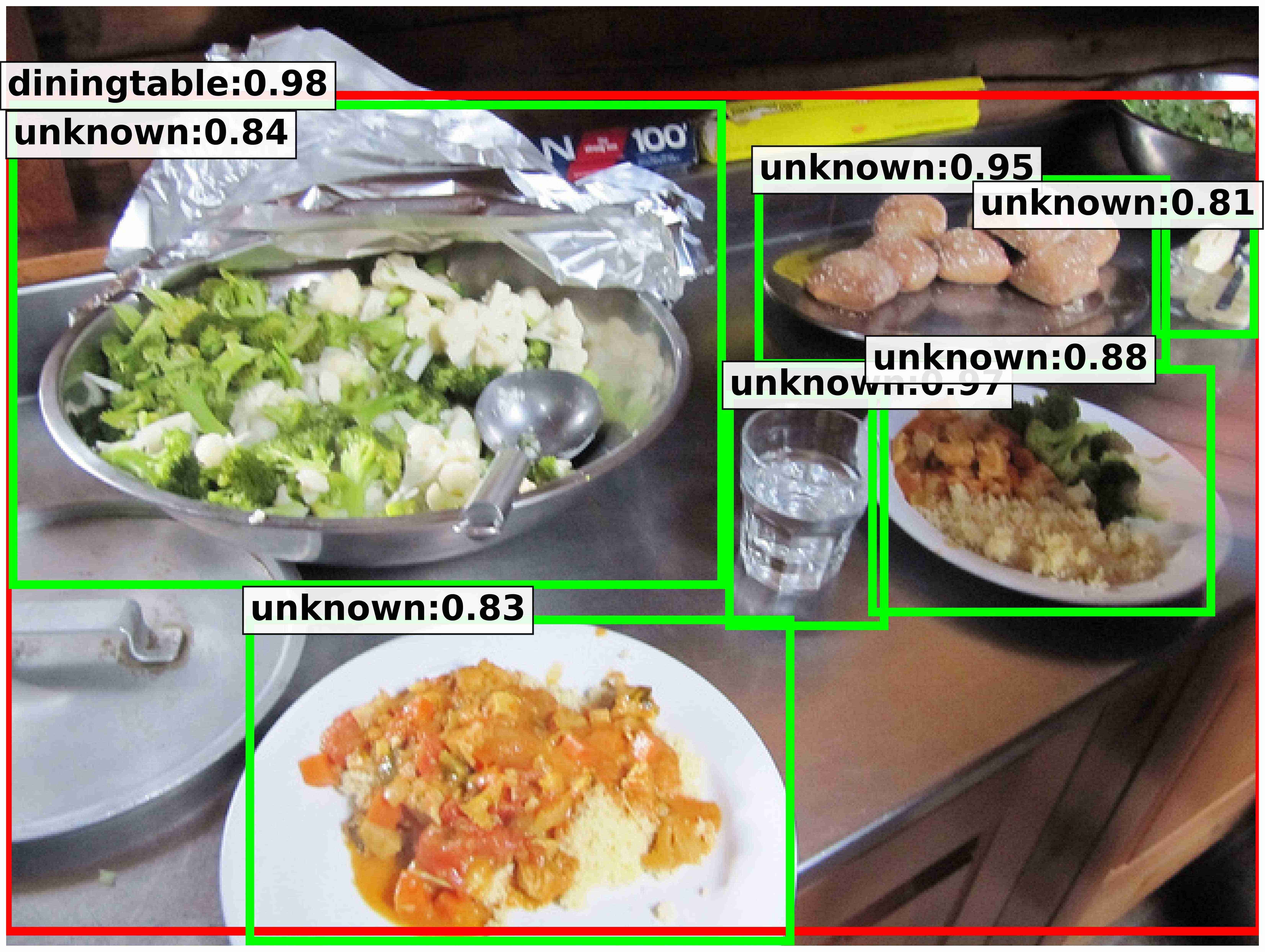}
  \end{subfigure}%
  \caption{Qualitative  results  of  unknown  detections in ORE \cite{joseph2021towards} when trained using RPN (left) versus MAVL (right) unknown pseudo-labels. Using proposals from MAVL as unknown pseudo-labels improves the prediction of unknowns. It reduces the misclassifications of unknown categories with other known categories. The second example (shown in top row - right side), corresponds to a sample in task 3 where ‘laptop' belongs to the unknown categories set, was misclassified as ‘TV', which is however correctly classified as an unknown with the improved model. This better aids in continual learning.}
  \label{figure:owod}
\end{figure*}
\begin{figure*}[!ht]
  \centering
  \begin{subfigure}{.24\columnwidth}
    \centering
    \includegraphics[height=2.5cm,width=\linewidth]{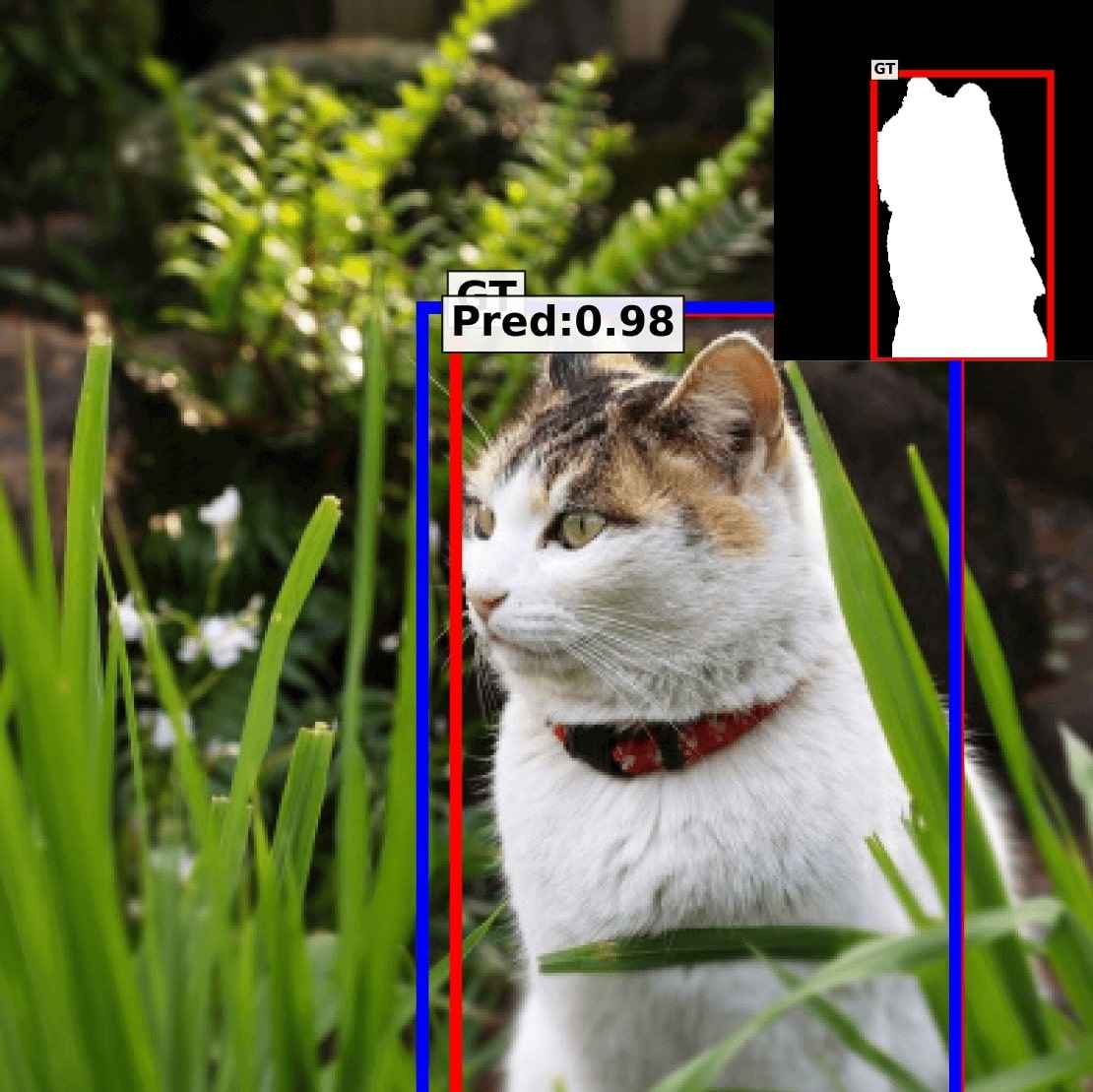}
  \end{subfigure}%
  \begin{subfigure}{.24\columnwidth}
    \centering
    \includegraphics[height=2.5cm,width=\linewidth]{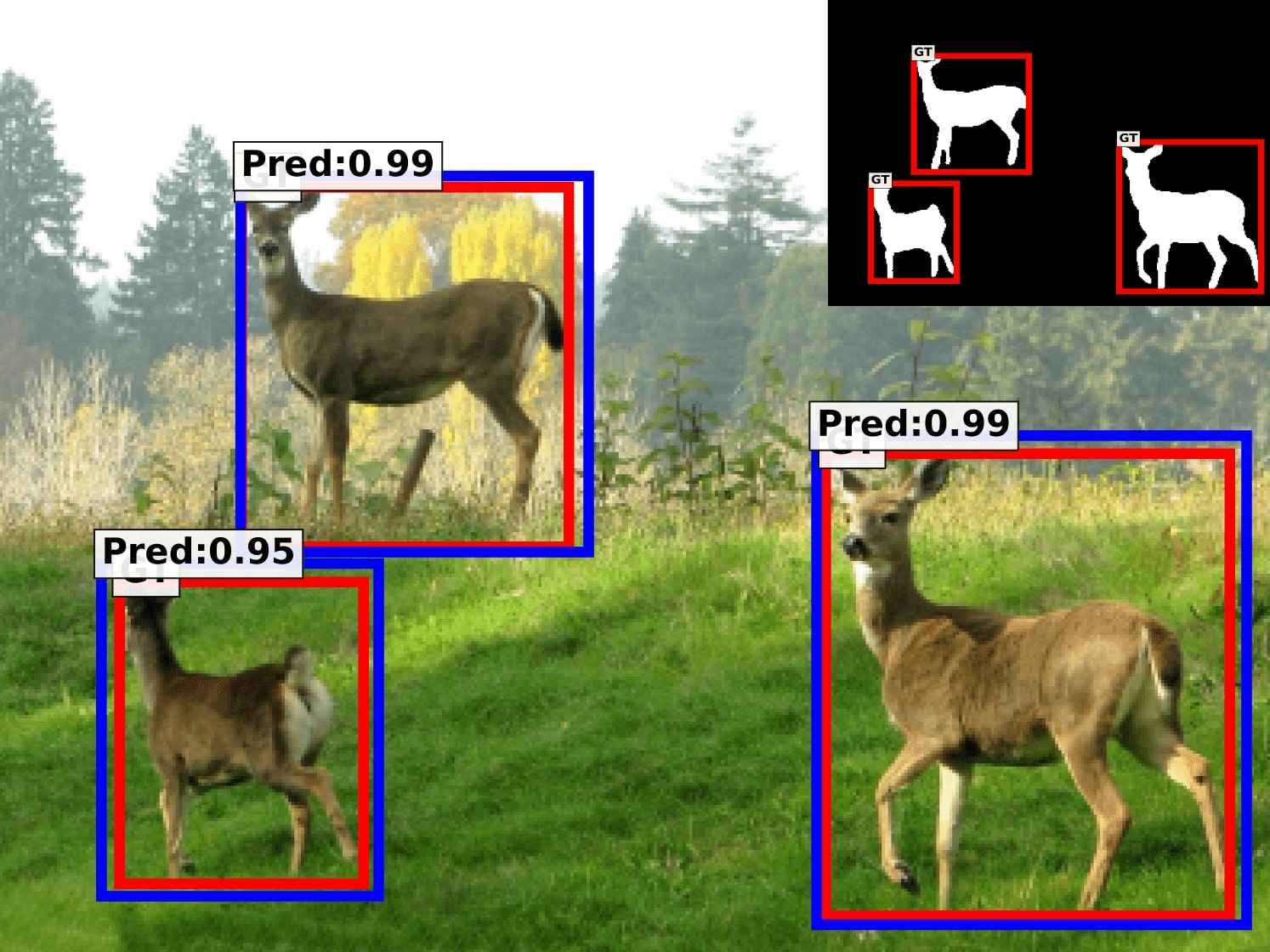}
  \end{subfigure}
  \begin{subfigure}{.24\columnwidth}
    \centering
    \includegraphics[height=2.5cm,width=\linewidth]{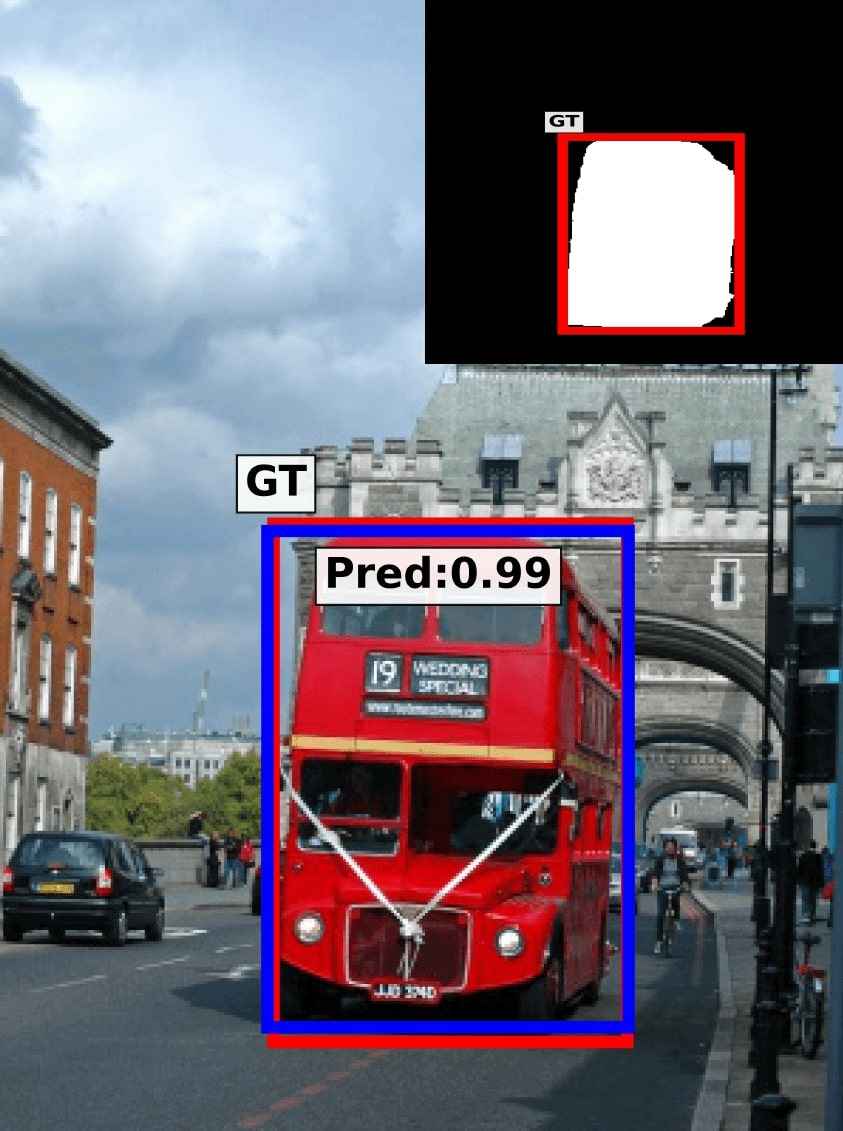}
  \end{subfigure}%
  \begin{subfigure}{.24\columnwidth}
    \centering
    \includegraphics[height=2.5cm,width=\linewidth]{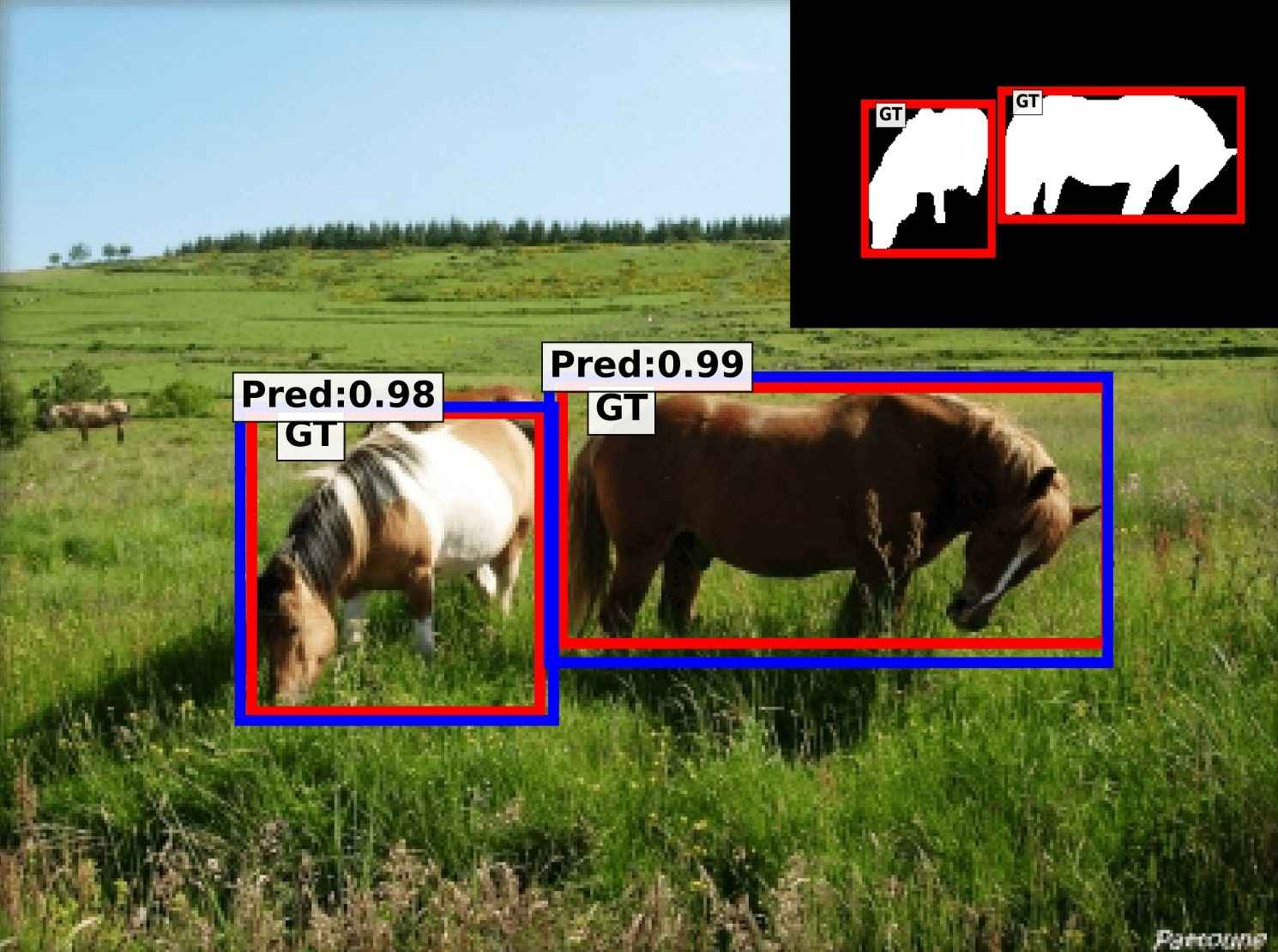}
  \end{subfigure}%
  \\
  \begin{subfigure}{.24\columnwidth}
    \centering
    \includegraphics[height=2.5cm,width=\linewidth]{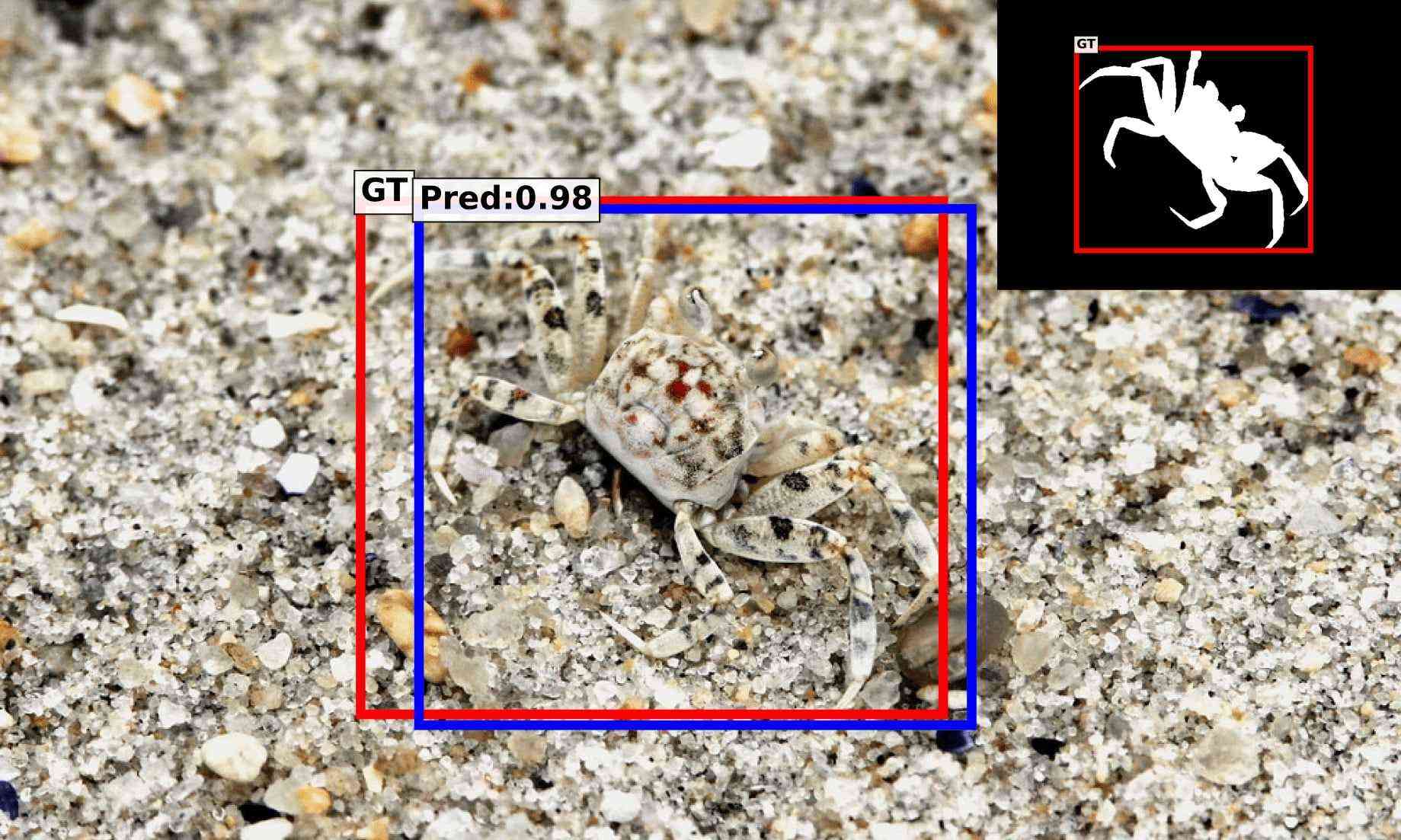}
  \end{subfigure}%
  \begin{subfigure}{.24\columnwidth}
    \centering
    \includegraphics[height=2.5cm,width=\linewidth]{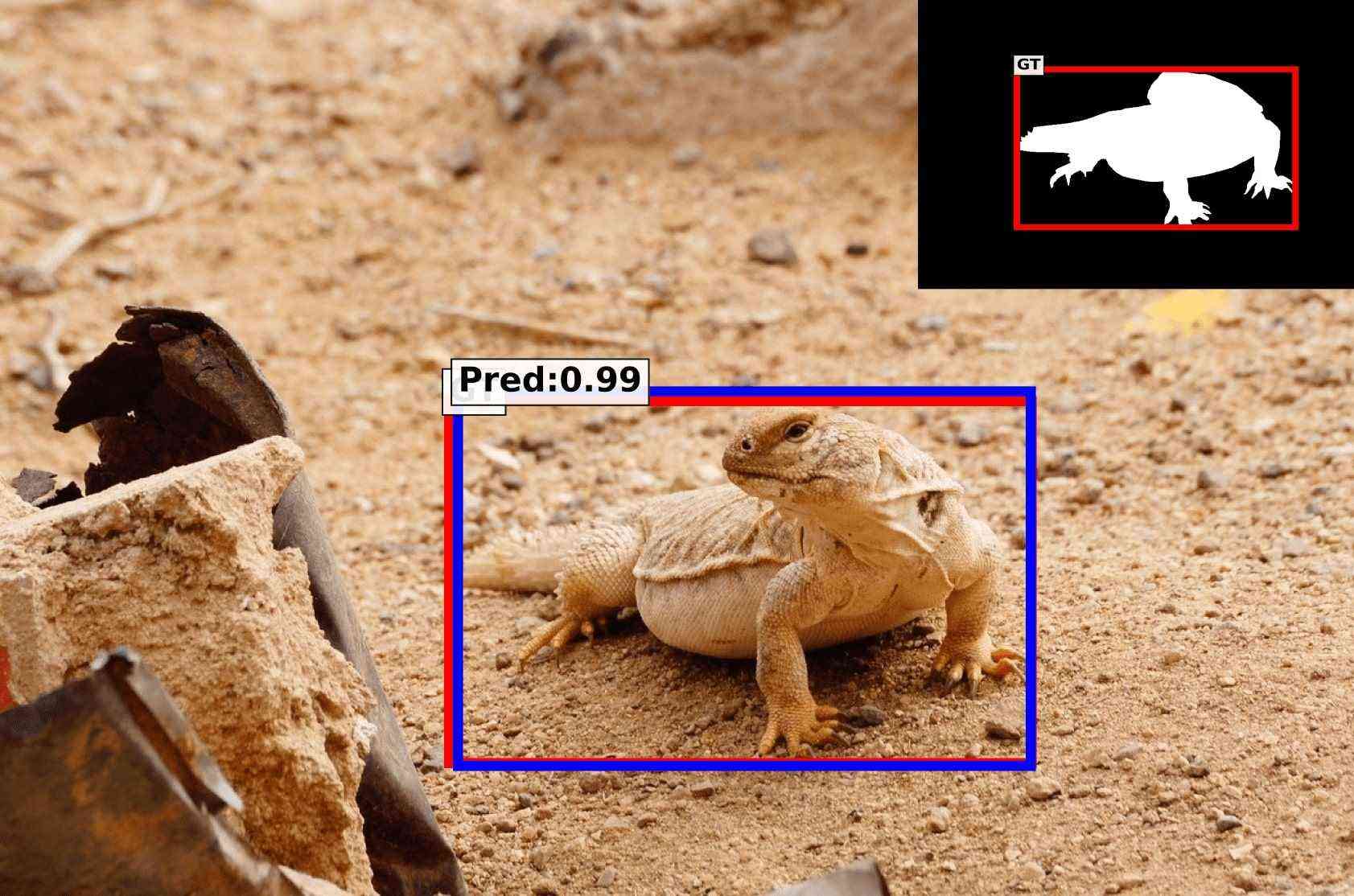}
  \end{subfigure}
  \begin{subfigure}{.24\columnwidth}
    \centering
    \includegraphics[height=2.5cm,width=\linewidth]{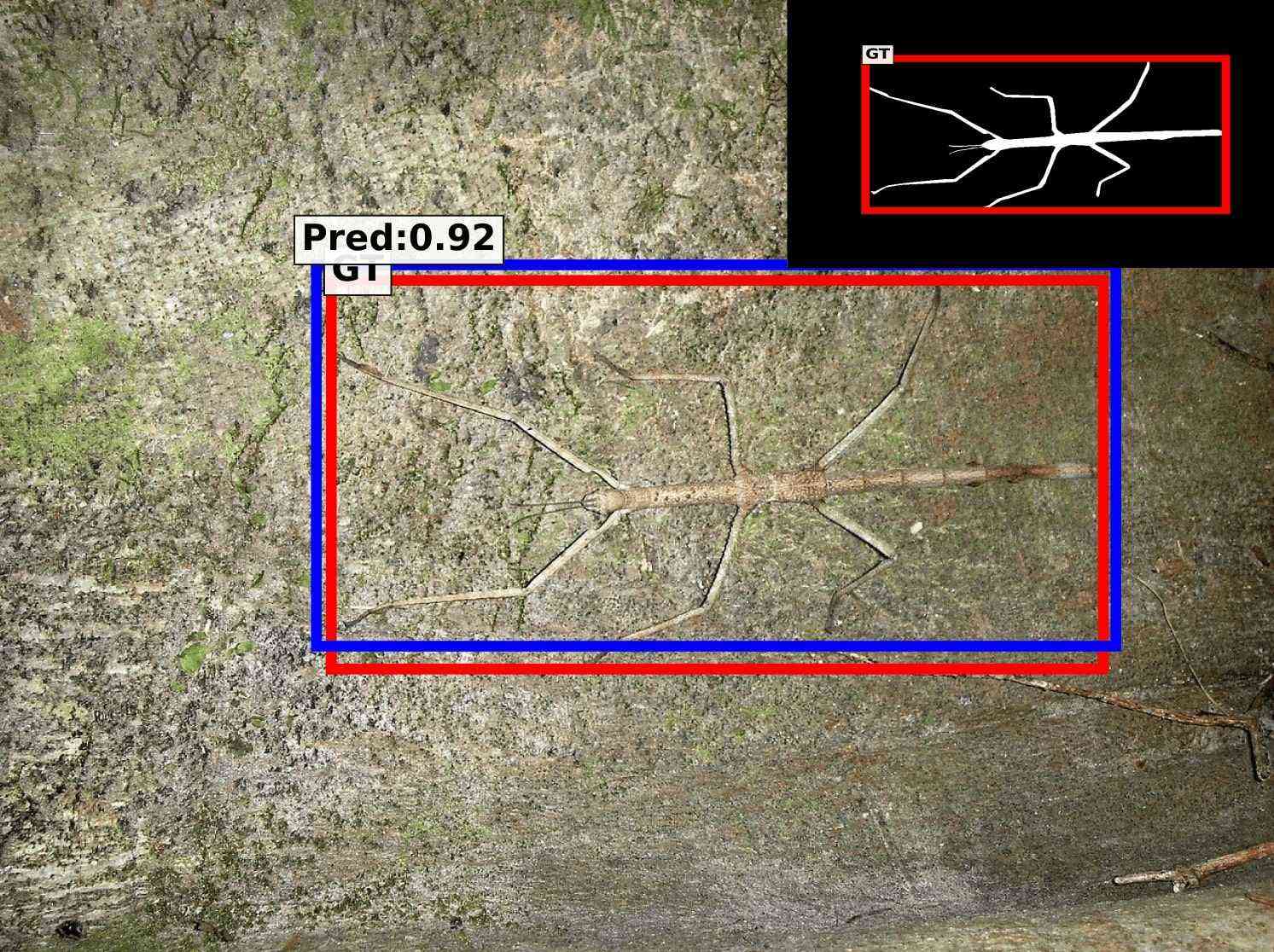}
  \end{subfigure}%
  \begin{subfigure}{.24\columnwidth}
    \centering
    \includegraphics[height=2.5cm,width=\linewidth]{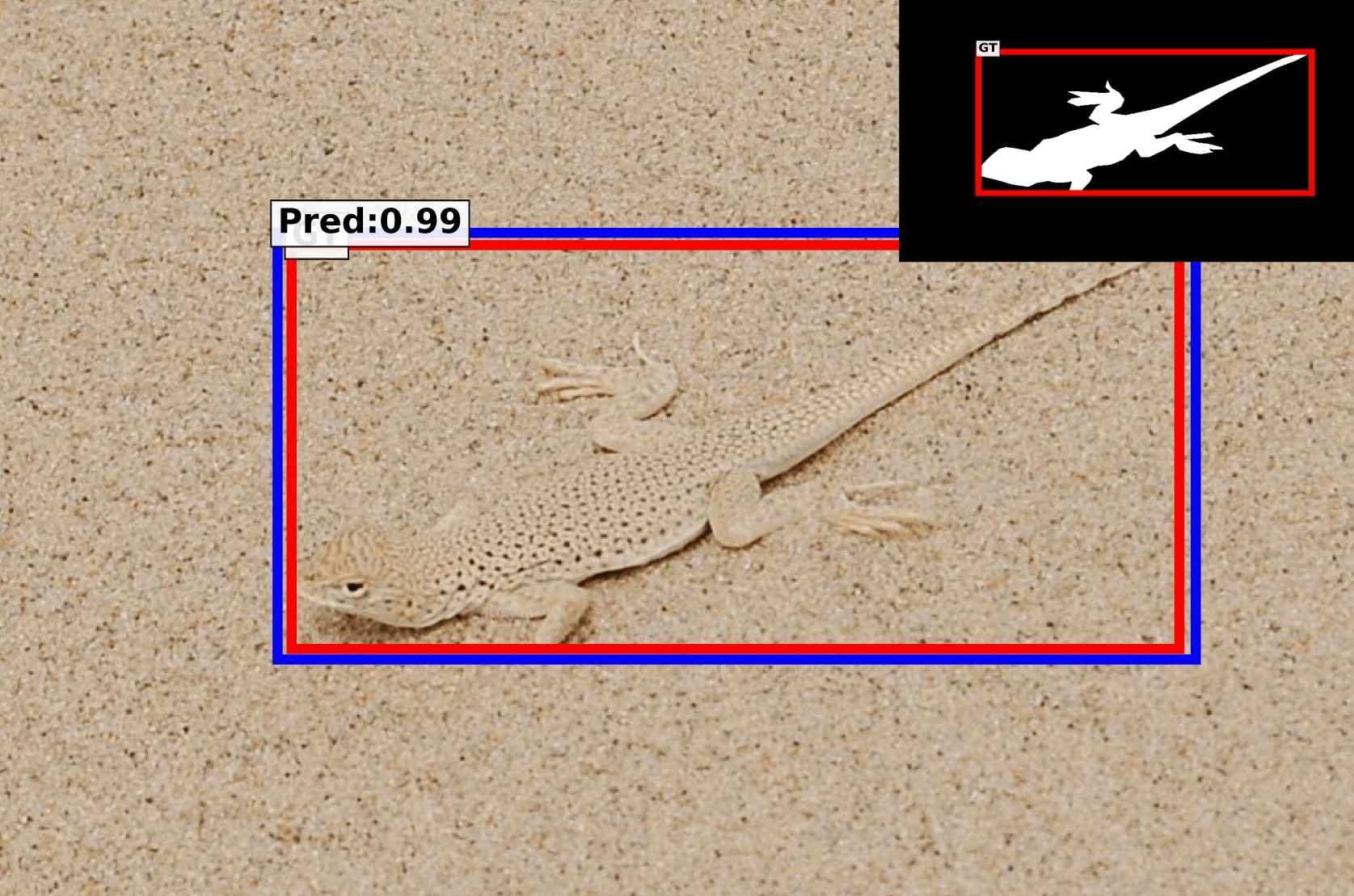}
  \end{subfigure}%
  \caption{\textbf{Top Rows}: Qualitative results of MAVL for Salient OD. \textbf{Bottom Rows}: Camouflaged OD (right) tasks. The ground-truth masks along with the generated  bounding boxes are shown on top right of the image}
  \label{figure:sod-cod}
\end{figure*}

Fig.~\ref{figure:owod} shows some examples of improved Open-world detector (ORE) trained with MAVL unknown pseudo-labels. The images on the left of each example correspond to the ORE trained with unknown pseudo-labels from RPN and on the right correspond to the ORE trained with unknown pseudo-labels from MAVL. The visualizations indicate that the improved model is better capable of detecting unknowns. Additionally, it reduces the misclassifications of unknown categories with other known categories. For example, the second sample in Fig.~\ref{figure:owod} (top row - right side), corresponds to a sample in task 3 where ‘laptop' belongs to the unknown categories set, was misclassified as ‘TV', which is however correctly classified as an unknown with the improved model. This is advantageous as it can better aid continual learning, \ie the model can learn about the unknown categories when additional information about the unknowns are obtained via supervision. In Fig.~\ref{figure:sod-cod}, we present examples of qualitative results obtained for salient OD and camouflaged OD with specific queries, ‘\txt{all salient objects}'  and ‘\txt{all camouflaged objects}' respectively, along with the bounding box annotations from the ground-truth masks.

\section{Additional Results}\label{app:add_results}

\subsection{Gains from MSDA in MAVL}
We ablate the contribution of MSDA in Table~\ref{rebuttal:msda_contribution} for our MAVL model. The class-agnostic OD results show the significance of MSDA.

\begin{table}[!h]
\centering
\begin{tabular}{lcccccc}
\toprule
\rowcolor{Gray} & \multicolumn{2}{c}{Pascal-VOC} & \multicolumn{2}{c}{MSCOCO} & \multicolumn{2}{c}{KITTI} \\
\rowcolor{Gray} Model & AP50 & R50 & AP50 & R50 & AP50 & R50 \\
\midrule
\midrule

MAVL w/o MSDA & 59.9 & 82.4 & 33.3 & 51.6 & 33.2 & 50.1 \\
MAVL & 65.0 & 89.1 & 39.3 & 62.0 & 39.0 & 61.0 \\
\bottomrule
\end{tabular}
\caption{Effect of removing MSDA from MAVL. It decreases the class-agnostic OD performance, indicating the importance of MSDA. The models are evaluated after 10 epochs for ablation.}
\label{rebuttal:msda_contribution}
\end{table}

\subsection{Impact of Late Fusion in MAVL}
The late fusion is crucial to our MAVL since it enables an efficient MViT design while keeping the multi-scale spatial information intact. Notably, early fusion (as in MDETR) ignores the spatial structure of images which makes it infeasible to operate with MSDA (that requires spatial information for deformable attention). Thus, MAVL effectively combines MSDA with late vision-text fusion and provides gain over MDETR in class-agnostic OD benchmarks. Unlike MDETR, our MAVL does not rely on contrastive alignment and thus removing MSDA significantly affects the results (Table~\ref{rebuttal:msda_contribution}).

\subsection{Generalization Ability onto Novel/Rare Classes}
Table~\ref{rebuttal:lvis_rare_recall} shows quantitative results on LVIS rare, common and frequent categories. (1) Similar to frequent and common, our MAVL provides good recall rates for rare LVIS categories, indicating its robust class-agnostic behavior. We note that most of the rare category instances in LVIS are of tiny size (area~$<$7${\times}$7 pixels) and have low recall ($\sim$19\%) as compared to the medium/large instances with much high recall ($\sim$86\%). (2) MAVL-ORE is trained by removing 60/80 common COCO categories from LMDet leaving only 0.76M/1.3M image-text pairs. This strict setting with much less training data also shows favorable rare class recall.

\begin{table}[!h]
\centering
\begin{tabular}{lccccc}
\toprule
\rowcolor{Gray} Model & Lang. & Rare & Common & Frequent & All \\
\midrule
\midrule
1:MAVL & $\times$ & 30.0 & 31.6 & 32.4 & 32.1   \\
\midrule
2:MAVL & \checkmark & 38.0 & 40.5 & 37.1 & 37.0   \\
3:MAVL-ORE & \checkmark & 33.4 & 36.7 & 33.2 & 33.1 \\
\bottomrule
\end{tabular}
\caption{Class-agnostic recall (R50) of MAVL on LVIS rare, common and frequent categories. MAVL-ORE is trained on a filtered dataset generated by removing all captions listing any of 60 unknown COCO categories evaluated in ORE~\cite{joseph2021towards}.}
\label{rebuttal:lvis_rare_recall}
\end{table}

\subsection{Querying All Class Names}
Table~\ref{rebuttal:category_wise_queries} shows the class-agnostic OD results of MAVL when queried using a general (e.g.,~combination of queries in Table 3) versus combining detections from all category specific queries. Specifically, we use query ‘{\small\txt{every $<$ category name $>$}}' for each category of a dataset and combine proposals using class-agnostic NMS. We note that MAVL generates better \emph{class-agnostic} detections with \emph{general} text queries.

\begin{table}[!h]
\centering
\begin{tabular}{lcccccc}
\toprule
\rowcolor{Gray} & \multicolumn{2}{c}{Pascal-VOC} & \multicolumn{2}{c}{MSCOCO} & \multicolumn{2}{c}{KITTI} \\
\rowcolor{Gray} Model & AP50 & R50 & AP50 & R50 & AP50 & R50 \\
\midrule
\midrule
MAVL (ours) & 68.6 & 91.3 & 43.6 & 65.0 & 48.2 & 63.5 \\
MAVL (cat-wise) & 61.7 & 91.2 & 36.7 & 64.6 & 47.7 & 59.8 \\
\bottomrule
\end{tabular}
\caption{Comparison of using general versus category-specific queries for class-agnostic OD on three datasets.}
\label{rebuttal:category_wise_queries}
\end{table}

\subsection{Salient Object Detection}\label{app:add_results_sod}
A common formulation of deep learning based Salient Object Detection (SOD) approaches is to predict a saliency map for each input image. We evaluate MAVL against state-of-the art SOD approaches by converting the bounding box predictions of the the MViT model to masks using a COCO \cite{coco} trained Mask-RCNN \cite{he2017mask} mask head. These converted masks are evaluated against the saliency predictions of PoolNet \cite{liu2019simple} and CPD \cite{wu2019cascaded} models on DUT-OMRON \cite{yang2013saliency} and ECSSD \cite{shi2015hierarchical} datasets (Table~\ref{table:sod_mask}). Following \cite{liu2019simple} and \cite{wu2019cascaded}, F-measure ($F_b$) and mean absolute Error (MAE) are reported. 

\begin{table}[!h]
\caption{Segmentation based evaluation of MAVL on salient and comouflaged object detection in comparison with the corresponding state-of-the art approaches. The MAVL proposals are converted to masks using COCO \cite{coco} trained mask head of Mask-RCNN \cite{he2017mask}.}
\setlength\tabcolsep{4pt}
    \begin{subtable}[t]{0.49\linewidth}
        \resizebox{\linewidth}{!}{
        \begin{tabular}{l *{4}{c}}
        \toprule
          Dataset $\rightarrow$ &\multicolumn{2}{c}{DUT-OMRON} & \multicolumn{2}{c}{ECSSD} \\
          Model & MAE 	$\downarrow$ & F-b $\uparrow$ & MAE 	$\downarrow$ & F-b $\uparrow$\\
          \midrule
          CPD \cite{wu2019cascaded} &0.06 & 0.79 & 0.04 & 0.94 \\
          PoolNet  \cite{liu2019simple} &0.05 & 0.87 & 0.04 & 0.95 \\
          MAVL(Ours) & 0.21 & 0.64 & 0.24 & 0.66\\
          \bottomrule                             
        \end{tabular}
        }
        \caption{\small 
        MAVL proposals from text query, \lq \txt{all salient objects} \rq are used.
        }
        \label{table:sod_mask}
    \end{subtable}
    \begin{subtable}[t]{0.49\linewidth}
        \resizebox{\linewidth}{!}{
        \begin{tabular}{l*{4}{c}}
          \toprule
          Model & $S_\alpha$ $\uparrow$ & $E_\phi$ $\uparrow$ & $F_\beta^w$ $\uparrow$ & MAE $\downarrow$\\
          \midrule
          SINET-V2 \cite{SINET-V2} & 0.78 & 0.87 & 0.66 & 0.04 \\
          MAVL(ours) & 0.49 & 0.53 & 0.28 & 0.27 \\
          \bottomrule                             
        \end{tabular}
        }
        \caption{\small 
        MAVL proposals generated using \lq \txt{all camouflaged objects}\rq~query are used.
        }
        \label{table:cod_mask}
    \end{subtable}
\end{table}

\subsection{Camouflaged Object Detection}\label{app:add_results_cod}
In this section, we compare camouflaged masks predictions of SINET-V2 \cite{SINET-V2} with MAVL. Similar to SOD task, the bounding box predictions from the MViT are converted to object masks using the mask head of COCO \cite{coco} trained Mask-RCNN \cite{he2017mask} model. Following \cite{fan2020camouflaged}, S-measure ($S_\alpha$), E-measure ($E_\phi$), weighted F-measure ($F_\beta^w$) and MAE of mask predictions are reported in Table \ref{table:cod_mask}.

\subsection{Effect of Various Backbones}
\noindent \textbf{ResNet vs. EfficientNet:}
We explore the class-agnostic OD performance of MViTs for different convolutional backbones. Following \cite{mdetr}, we compare the ResNet-101 \cite{he2016deep} taken from Torchvision with EfficientNet-E5 \cite{tan2019efficientnet} taken from Timm Library \cite{rw2019timm}. The ResNet model is trained on ImageNet \cite{russakovsky2015imagenet} and achieves $77.4$\% top-1 accuracy on ImageNet validation, while the EfficientNet model is trained using Noisy-Student \cite{noisy_student} on an additional 300M unlabelled images achieving 85.1\% top-1 accuracy on ImageNet validation.

Table \ref{table12:r101_vs_e5} indicates that using a stronger backbone improves the class-agnostic OD accuracy across different dataset domains. The performance boost is significant for out of domain datasets, Kitchen \cite{kitchen}, Clipart, Comic and Watercolor \cite{clipart-comic-water}, indicating better generalization of MViT when trained using a stronger backbone model.

\begin{table*}[!h]
\caption{Class-agnostic object detection performance of MDETR \cite{mdetr} for different convolutional backbones. The results indicate that the use of strong backbone improves the results especially on the out-of-domain (Kitchen \cite{kitchen}, Clipart, Comic, Watercolor \cite{clipart-comic-water}) datasets.}
\begin{center}
\resizebox{\linewidth}{!}{%
\begin{tabular}{l *{16}{c}}
  \toprule
  Dataset & \multicolumn{2}{c}{Pascal VOC} & \multicolumn{2}{c}{COCO} & \multicolumn{2}{c}{KITTI} & \multicolumn{2}{c}{Kitchen} & \multicolumn{2}{c}{Clipart} & \multicolumn{2}{c}{Comic} & \multicolumn{2}{c}{Watercolor} & \multicolumn{2}{c}{DOTA}
  \\ 
  Model & AP50 & R-50 & AP50 & R-50 & AP50 & R-50 & AP50 & R-50 & AP50 & R-50 & AP50 & R-50 & AP50 & R-50 & AP50 & R-50 \\
  \midrule
  \midrule
  MDETR-R101 & 66.0 & 90.1 & 40.7 & 62.2 & 46.7 & 67.2 & 38.4 & 91.4 & 44.9	& 90.7	& 55.8	& 89.5	& 63.6	& 94.3	& 1.94 & 21.8 \\ 
  MDETR-E5 & \textbf{69.6} & 90.0 & \textbf{42.3} & 61.3 & \textbf{48.1} & 65.2 & \textbf{53.3} & 91.5 & \textbf{62.3} & 92.7 & \textbf{69.9} & 90.5 & \textbf{74.4} & 95.0 & \textbf{3.71} & 24.9  \\
  \bottomrule                             
\end{tabular}}
\end{center}
\label{table12:r101_vs_e5}
\end{table*}

\section{Related work}\label{app:related_work}
\noindent \textbf{Class-Agnostic Detection:} 
The class-agnostic OD is relatively less studied compared to class-aware detection.However, many object proposal generation algorithms have been proposed, since it remains a critical step in many applications like recognition and detection. The proposal generation algorithms can be categorized into three categories: (a) bottom-up segmentation based, (b) edge information based and (c) data-driven approaches based on deep neural network (DNN) architectures. 
In the first category that uses segmentation to derive proposals, multiple pixel groupings (superpixels) are merged according to various heuristics. Alexe \etal proposed an objectness \cite{alexe2012measuring} scoring method that combines various low-level features such as edges, color and superpixels to score object proposals. Selective Search \cite{uijlings2013selective} uses multiple hierarchical segmentations based on superpixels for object proposals. Similarly, MCG \cite{pont2016multiscale} uses segment hierarchy to group regions. Among the second category approaches, EdgeBoxes \cite{zitnick2014edge} scores bounding box proposals based on contours that the boxes enclose. BING algorithm \cite{cheng2014bing,zhang2015bing++} generates binary features based on edge information for fast objectness estimation. 

DNNs have also been investigated for generating object proposals. DeepBox \cite{kuo2015deepbox} proposes a network that can be used to rerank any bottom-up proposals, \eg the ones generated by EdgeBox \cite{zitnick2014edge}. DeepMask \cite{pinheiro2015learning} generates rich object segmentations and an associated score of the likelihood of the patch to fully contain a centered object. A refinement of this method is proposed in SharpMask \cite{pinheiro2016learning}. Alternatively, Ren \etal proposed region proposal network (RPN) \cite{ren2015faster} for generating object proposals, that identifies a set of regions that potentially contain objects along with corresponding objectness score. These are then refined for classification and localization for class-aware object detection. These are widely used in many two-stage objects detectors \eg RCNN variants  \cite{ren2015faster,he2017mask,lin2017feature}. Jaiswal \etal proposed an adversarial framework \cite{jaiswal2021class} for class-agnostic object detection which replaces object type classification head with a binary classifier for class-agnostic detection. 
Another recent work proposes an Object Localization Network (OLN) \cite{kim2021learning} that replaces the classifier head in Faster-RCNN~\cite{ren2015faster}  with localization quality estimators such as centerness and IoU score for objectness estimation. 
Alternatively, Sim{\'e}oni \etal proposed a method \cite{simeoni2021localizing} that extracts features from a DINO \cite{caron2021emerging} self supervised pre-trained transformer that uses patch correlations in an image to propose object proposals.  

\noindent \textbf{Multi-modal Transformers:} Multi-modal Vision Transformers (MViT) typically involve learning task agnostic vision-language (V+L) representations using millions of image-text pairs and then transferring the knowledge to downstream tasks \cite{OSCAR,UNITER,mdetr}. Inspired from the success of BERT \cite{BERT} in natural language processing (NLP), VisualBERT \cite{VisualBERT}, ViLBERT \cite{ViLBERT} and LXMERT \cite{LXMERT} jointly learn V+L representations using image-caption pairs. They utilize a pretrained region proposal method \cite{ren2015faster} and learn the V+L correlation using self-supervised tasks such as mask language modeling and sentence image alignment. 
In a concurrent work, VL-BERT \cite{VL-BERT} performs pretraining on both text-only and visual-linguistic datasets and achieve an improved performance on multiple downstream visual comprehension tasks. UNITER \cite{UNITER} introduces Word-Region Alignment (WRA) pretraining task using Optimal Transport (OT) \cite{OT} which facilitates the alignment between text and image regions. It only masks one modality at a time while keeping the other modality intact which helps it to better capture the V+L relationships. 

All these methods utilize an off-the-shelf region proposal method \cite{ren2015faster} which usually produces noisy regions. OSCAR \cite{OSCAR} tries to mitigate this problem by using object detector tags for modeling V+L understanding. It relies on the fact that the salient objects in the image are easy to detect and are typically mentioned in the caption. Alternatively, MDETR \cite{mdetr} leverages explicit alignment between text and ground-truth bounding boxes to learn visual-language alignment. It builds on-top-of recently proposed DETR \cite{DETR} model, generalizes to unseen concepts and outperforms the previous methods on many V+L downstream tasks.
Going further, 12-in-1 \cite{12-in-1} utilizes the pretrained V+L representations and performs a joint training of a single model on 12 datasets. This learning paradigm improves the single task performance as compared to the typical task-wise training by achieving superior results on 11 out of 12 tasks. Gupta \etal proposed GPV-I \cite{gpv1}, a unified architecture for multi-task learning, where the task is inferred from the text prompt. It takes an image and a task description as input and outputs text with the corresponding bounding boxes. It is also based on DETR \cite{DETR}. We observe that these \cite{mdetr,gpv1} multi-modal transformers, which are trained using aligned image-text pairs, produce high quality object proposals by using simple text queries e.g., ‘\txt{all objects}'.

\noindent \textbf{Unsupervised Approaches:} Recently, many unsupervised pretraining methods are proposed for the object detection task. Xiao \etal introduced ReSim \cite{xiao2021region} to encode both the region and global representations during self-supervised pretraining. In addition to the standard contrastive learning objective \cite{MoCo,pmlr-v119-chen20j}, it slides a window in the overlapping region of the different views of an image and maximizes the feature similarity of the corresponding features across all convolutional layers. DetCo \cite{xie2021detco} approaches this problem by generating both the global views and local patches from an image and defines hierarchical global-to-global, local-to-local and global-to-local contrastive objectives. UP-DETR \cite{dai2021up} proposes ‘\txt{random query patch detection}' pretext task for pretraining of DETR \cite{DETR}. The random patches from the image are generated and the model is trained on a large-scale dataset to locate these patches. DETReg \cite{detreg} argues that it is necessary to pre-train both the backbone and the detection network for learning good representations for object detection downstream tasks. It utilizes an off-the-shelf selective search \cite{uijlings2013selective} proposal generation algorithm for acquiring pseudo-labels for localization and pretrained contrastive clustering based SwAV \cite{SwAV} model for separating categories in the feature space. All these methods can be used for generating class-agnostic object proposals after the unsupervised pretraining. However, as shown in our analysis, the unsupervised approaches do not perform as well as the proposed class-agnostic OD framework based on supervised MViTs.

\end{document}